\RequirePackage{fix-cm}
\documentclass[twocolumn]{svjour3}          % twocolumn
\smartqed  % flush right qed marks, e.g. at end of proof
\usepackage{graphicx}

\usepackage[toc,title]{appendix}
\usepackage{multirow}
\usepackage{ifthen}
\usepackage{latexsym}
\usepackage{color}
\usepackage{array}
\usepackage{algorithm2e}
\usepackage{url}
\usepackage{amsmath}
\usepackage{amssymb}
\usepackage{rotating}
\usepackage{bbold}
\usepackage{makeidx}
\usepackage{natbib}
\usepackage{pdflscape}
\usepackage{afterpage}
\usepackage{float}

\newcommand{\todo}[1]{{ \color{blue} #1 }}
\DeclareMathOperator*{\argmax}{arg\,max}

\newcommand{\side}[2]{\rotatebox{90}{\makebox[#1in][c]{#2}}}
\newcommand{\oi}{$^\dagger$}
\newcommand{\oii}{$^\ddag$}
\newcommand{\oiii}{$^\mp$}
\newcommand{\oiv}{$^\S$}
\newcommand{\ov}{$^\varsigma$}

\newcommand{\nv}{0.8in}
\newcommand{\iv}{2.5in}
\newcommand{\crv}{4.5in}

\newcommand{\nw}{0.8in}
\newcommand{\nwe}{0.8in}
\newcommand{\ciw}{2.75in}
\newcommand{\crw}{3.75in}

\newcommand{\nz}{0.6in}

\newcommand{\ciz}{1.6in}
\newcommand{\crz}{4.3in}
\newcommand{\cz}{0.2in}
\newcommand{\zl}{$\circ$}

\newcommand{\crf}{\tiny}
\newcommand{\nf}{\footnotesize}

\begin{document}

\title{ImageNet Large Scale Visual Recognition Challenge}

%\title{Insert your title here%\thanks{Grants or other notes

\titlerunning{ImageNet Large Scale Visual Recognition Challenge}        % if too long for running head

\author{Olga Russakovsky*         \and
        Jia Deng* \and
Hao Su \and 
Jonathan Krause \and \\
Sanjeev Satheesh \and
Sean Ma \and
Zhiheng Huang \and
Andrej Karpathy \and 
Aditya Khosla \and
Michael Bernstein \and 
Alexander C. Berg \and
Li Fei-Fei
}

% 2010: Alex, Jia, FF
% 2011: Alex, Jia, Sanjeev, Hao, FF
% 2012: Jia, Alex, Sanjeev, Hao, Aditya, FF
% 2013: Olga, Jia, Jon, Alex, FF
% 2014: Olga, Sean, Jon, Jia, Alex, FF


\institute{O. Russakovsky* \at
Stanford University, Stanford, CA, USA \\
\email {olga@cs.stanford.edu} 
\and
J. Deng* \at
University of Michigan, Ann Arbor, MI, USA \\
(* = authors contributed equally)
\and
H. Su \at
Stanford University, Stanford, CA, USA 
\and
J. Krause \at
Stanford University, Stanford, CA, USA 
\and
S. Satheesh \at
Stanford University, Stanford, CA, USA 
\and
S. Ma \at
Stanford University, Stanford, CA, USA 
\and
Z. Huang \at
Stanford University, Stanford, CA, USA 
\and
A. Karpathy \at
Stanford University, Stanford, CA, USA 
\and
A. Khosla \at
Massachusetts Institute of Technology, Cambridge, MA, USA
\and
M. Bernstein \at
Stanford University, Stanford, CA, USA 
\and
A. C. Berg \at
UNC Chapel Hill, Chapel Hill, NC, USA
\and
L. Fei-Fei \at
Stanford University, Stanford, CA, USA 
%F. Author \at
 %             first address \\
  %            Tel.: +123-45-678910\\
   %           Fax: +123-45-678910\\
    %          \email{fauthor@example.com}           %  \\
      %     \and
       %    S. Author \at
         %     second address
}

\date{Received: date / Accepted: date}

\maketitle

\begin{abstract}
The ImageNet Large Scale Visual Recognition Challenge is a
benchmark in object category classification and detection on hundreds of 
object categories and millions of images. The challenge has been run
annually from 2010 to present, attracting participation from
more than fifty institutions.

This paper describes the creation of this benchmark dataset and 
the advances in object recognition
that have been possible as a result. 
We discuss the challenges of collecting large-scale ground truth annotation, 
highlight key breakthroughs in categorical object recognition,
provide a detailed analysis of the current state of the field of large-scale image classification and object detection,
and compare the state-of-the-art computer vision accuracy with human accuracy.
We conclude with lessons learned in the five years of the challenge, and propose
future directions and improvements.

\end{abstract}

%%%%%%%%%%%%%%%%%%%%%%%%%%%%%%%%%%%%%%%%%%%%%%%%%%%%%%%%%%%%%%%%%%%%%%%%%%%%%%%%%%%%%%%%%%
%%%%%%%%%%%%%%%%%%%%%%%%%%%%%%%%%%%%%%%%%%%%%%%%%%%%%%%%%%%%%%%%%%%%%%%%%%%%%%%%%%%%%%%%%%
%%%%%%%%%%%                                           %%%%%%%%%%%%%%%%%%%%%%%%%%%%%%%%%%%%%%%%%%%%%%%%%%%%%%%%%%%
%%%%%%%%%%%             INTRODUCTION      %%%%%%%%%%%%%%%%%%%%%%%%%%%%%%%%%%%%%%%%%%%%%%%%%%%%%%%%%%%
%%%%%%%%%%%                                           %%%%%%%%%%%%%%%%%%%%%%%%%%%%%%%%%%%%%%%%%%%%%%%%%%%%%%%%%%%
%%%%%%%%%%%%%%%%%%%%%%%%%%%%%%%%%%%%%%%%%%%%%%%%%%%%%%%%%%%%%%%%%%%%%%%%%%%%%%%%%%%%%%%%%%
%%%%%%%%%%%%%%%%%%%%%%%%%%%%%%%%%%%%%%%%%%%%%%%%%%%%%%%%%%%%%%%%%%%%%%%%%%%%%%%%%%%%%%%%%%

\section{Introduction}
\label{intro}

\paragraph{Overview.} The ImageNet Large Scale Visual Recognition Challenge (ILSVRC) has been running annually for five years (since 2010) and has
become the standard benchmark for large-scale object recognition.\footnote{In this paper, we will be using the term \emph{object recognition}
broadly to encompass  both \emph{image classification} (a task requiring an algorithm to determine what object classes are present in the image) as well as \emph{object detection} (a task
requiring an algorithm to localize all objects present in the image).}
 ILSVRC follows in the footsteps of
the PASCAL VOC challenge~\citep{PASCALVOC}, established in 2005, which set the precedent for 
standardized evaluation of recognition algorithms in the form of yearly competitions. As in PASCAL VOC, ILSVRC consists of two components:
(1) a publically available \emph{dataset}, and (2) an annual \emph{competition} and corresponding workshop.
The dataset allows for the development and comparison of categorical object recognition algorithms, and the
competition and workshop provide a way to track the progress and discuss the lessons learned from the most successful and innovative
entries each year.

The publically released dataset contains a  set of manually annotated \emph{training} images. A set of \emph{test} images is also released, with the manual annotations withheld.\footnote{ In 2010, the test annotations were later released publicly;
since then the test annotation have been kept hidden.} Participants train their algorithms using the training images and then automatically annotate the test images. These predicted annotations are submitted to the  \emph{evaluation server}. Results of the evaluation are revealed at the end of the competition period and authors are invited to share insights at the workshop held at the International Conference on Computer Vision (ICCV) or European Conference on Computer Vision (ECCV) in alternate years. 

ILSVRC annotations fall into one of two categories: (1) \emph{image-level annotation} of a binary label for the presence or absence of an object class in the image, e.g., ``there are cars in this image'' but ``there are no tigers,'' and (2) \emph{object-level annotation} of a tight bounding box and class label around an object instance in the image, e.g., ``there is a screwdriver centered at position (20,25) with width of 50 pixels and height of 30 pixels''.  

\paragraph{Large-scale challenges and innovations.} In creating the dataset, several challenges had to be addressed. Scaling up from 19,737 images in PASCAL VOC 2010 to 1,461,406
in ILSVRC 2010 and  from 20 object classes to 1000 object classes brings with it several challenges. It is no longer feasible for a small group of annotators to annotate the data
as is done for other datasets~\citep{Caltech101,MSRC,PASCALVOC,SUN}.  Instead we turn to designing novel crowdsourcing approaches for collecting large-scale annotations~\citep{Su12,ImageNet,Deng14}.

Some of the 1000 object classes may not be as easy to annotate as the 20 categories of PASCAL VOC: e.g., bananas which appear in bunches may not be as easy to delineate as the basic-level categories of aeroplanes or cars. Having more than a million images makes it infeasible to annotate the locations of all objects (much less with object segmentations, human body parts, and other detailed annotations that subsets of PASCAL VOC contain). New evaluation criteria  have to be defined to take into account the facts that obtaining perfect manual annotations in this setting may be infeasible.

Once the challenge dataset was collected, its scale allowed for
unprecedented opportunities both in evaluation of object recognition
algorithms and in developing new techniques. Novel algorithmic innovations emerge with the availability of large-scale training data.  The broad spectrum of object categories motivated the 
need for algorithms that are even able to distinguish classes which are visually very similar. We highlight the most successful of these algorithms in this paper, and compare their performance with human-level accuracy.

Finally, the large variety of object classes in ILSVRC allows us to perform an analysis of statistical properties of objects and their impact on recognition algorithms. This type of analysis allows for a deeper understanding of object recognition, and for designing the next generation of general object recognition algorithms.

\paragraph{Goals.} This paper has three key goals:
\begin{enumerate}
\item To discuss the challenges of creating this large-scale object recognition benchmark dataset,
\item To highlight
the developments in  object classification and detection that have resulted from this effort, and 
\item To take a closer look at the current state of the field
of categorical object recognition.
\end{enumerate}
The paper may be of interest to researchers working on creating large-scale datasets, as well as to anybody interested in better understanding the history and the current state of large-scale object recognition.

The collected dataset and additional information about ILSVRC can be found at:
{\begin{center}
\url{http://image-net.org/challenges/LSVRC/}
\end{center}
}

\subsection{Related work}
\label{sec:rel}

We briefly discuss some prior work in constructing benchmark image datasets.

\paragraph{Image classification datasets.} Caltech 101~\citep{Caltech101} was among the first 
standardized datasets for multi-category image classification, with 101 object classes and commonly 15-30 training images per class. Caltech 256~\citep{Caltech256} increased the number of object classes to 256 and added images with greater scale and background variability.  The TinyImages dataset~\citep{TinyImages} contains 80 million 32x32 low resolution images collected from the internet using synsets in WordNet~\citep{WordNet} as queries. However, since this data has not been manually verified, there are many errors, making it less suitable for algorithm evaluation. Datasets such as 15 Scenes~\citep{Oliva01,BoW, Lazebnik06} or recent Places~\citep{Zhou14} provide a single scene category label (as opposed to an object category).

The ImageNet dataset~\citep{ImageNet} is the backbone of ILSVRC. 
ImageNet is an image dataset organized according to the WordNet hierarchy~\citep{WordNet}. Each  concept in WordNet, possibly described by multiple words or word phrases, is called a ``synonym set'' or ``synset''. ImageNet populates 21,841 synsets of WordNet with an average of 650 manually verified and full resolution images. As a result, ImageNet contains 14,197,122 annotated images organized by the semantic hierarchy of WordNet (as of August 2014). ImageNet is larger in scale and diversity than the other image classification datasets. ILSVRC uses a subset of ImageNet images for training the algorithms and some of ImageNet's image collection protocols for annotating additional images for testing the algorithms.

%MNIST, COIL, CIFAR

% Caltech Pedestrian Datasets

\paragraph{Image parsing datasets.} Many datasets aim to provide richer image annotations beyond image-category labels. LabelMe~\citep{LabelMe}  contains general photographs with multiple objects per image. It has bounding polygon annotations around objects, but the object names are not standardized: annotators are free to choose which objects to label and what to name each object. The SUN2012~\citep{SUN} dataset contains 16,873 manually cleaned up and fully annotated images more suitable for standard object detection training and evaluation. SIFT Flow~\citep{Liu11} contains 2,688 images labeled using the LabelMe system. The LotusHill dataset~\citep{LotusHill} contains very detailed annotations of objects in 636,748 images and video frames, but it is not available for free.  Several datasets provide pixel-level segmentations: for example, MSRC dataset~\citep{MSRC} with 591 images and 23 object classes, Stanford Background Dataset \citep{Gould09} with 715 images and 8 classes, and the Berkeley Segmentation dataset~\citep{Arbelaez11} with 500 images annotated with object boundaries. OpenSurfaces segments surfaces from consumer photographs and annotates them with surface properties, including material, texture, and contextual information~\citep{Bell13} . 

%It has been used in nonparametric, data-driven approaches~\citep{Tighe14, malisiewicz-cvpr08, 

The closest to ILSVRC is the PASCAL VOC dataset \citep{PASCALIJCV,PASCALIJCV2}, which provides a standardized test bed for object detection, image classification, object segmentation, person layout, and action classification. Much of the design choices in ILSVRC have been inspired by PASCAL VOC and the similarities and differences between the datasets are discussed at length throughout the paper. 
ILSVRC scales up  PASCAL VOC's goal of standardized training and evaluation of recognition algorithms by more than an order of magnitude in number
of object classes and images: PASCAL VOC 2012 has 20 object classes and 21,738 images compared to ILSVRC2012 with 1000 object classes and 1,431,167 annotated images.

% in ILSVRC2012, and 127,931 fully annotated images in ILSVRC2014. 

The recently released COCO dataset~\citep{COCO} contains more than 328,000 images with 2.5 million object instances manually segmented. It has fewer object categories than ILSVRC (91 in COCO versus 200 in ILSVRC object detection) but more instances per category (27K on average compared to about 1K in ILSVRC object detection). Further, it contains object segmentation annotations which are not currently available in ILSVRC. COCO is likely to become another important large-scale benchmark.

\paragraph{Large-scale annotation.} ILSVRC makes extensive use of Amazon Mechanical Turk to obtain accurate
annotations~\citep{Sorokin08}. Works such as~\citep{Welinder10,GAL,Vittayakorn11} describe quality control mechanisms for this marketplace. \citep{Vondrick12} provides a detailed overview of crowdsourcing video annotation. A related line of work is to obtain annotations through well-designed games, e.g.~\citep{vonAhn05}. 
Our novel approaches to crowdsourcing accurate image annotations are in Sections~\ref{sec:AnnotClsAnnot}, \ref{sec:AnnotLocBbox} and \ref{sec:AnnotDetList}.

\paragraph{Standardized challenges.} There are several datasets with standardized online evaluation 
similar to ILSVRC: the aforementioned PASCAL VOC~\citep{PASCALVOC}, Labeled Faces in the Wild~\citep{LFW} for unconstrained face recognition, 
Reconstruction meets Recognition~\citep{RR} for 3D reconstruction and KITTI~\citep{KITTI} for computer vision in autonomous driving. These datasets along with ILSVRC help benchmark progress in different areas of computer vision.
Works such as~\citep{Torralba11} emphasize the importance of examining the bias inherent in any standardized dataset.

% harvesting image databases from the web?
 
%\iffalse
\subsection{Paper layout} 

%\todo{jia: I in general think description of paper layout is useless and can be removed.  }

We begin with a brief overview of ILSVRC challenge tasks in Section~\ref{sec:Tasks}. Dataset
collection and annotation are described at length in Section~\ref{sec:Annot}. 
Section~\ref{sec:Evaluation} discusses the evaluation criteria of algorithms in the large-scale recognition setting. 
Section~\ref{sec:Methods} provides an overview of the methods developed by ILSVRC participants.
%, with Section~\ref{sec:MethodsDeep} focusing
%on a notable paradigm shift in object recognition that has occurred over the years. 

Section~\ref{sec:Results} contains an in-depth  analysis of ILSVRC results:
 Section~\ref{sec:ResultsYears} documents the progress of large-scale recognition over the years, Section~\ref{sec:ResultsStats} concludes that ILSVRC results are statistically significant,
Section~\ref{sec:ResultsICCV}  thoroughly analyzes the current state of the field of object recognition, and Section~\ref{sec:ResultsHuman} compares state-of-the-art computer vision accuracy with human accuracy. We conclude and
discuss lessons learned from ILSVRC in Section~\ref{sec:Conclusion}.
%\fi

%%%%%%%%%%%%%%%%%%%%%%%%%%%%%%%%%%%%%%%%%%%%%%%%%%%%%%%%%%%%%%%%%%%%%%%%%%%%%%%%%%%%%%%%%%
%%%%%%%%%%%%%%%%%%%%%%%%%%%%%%%%%%%%%%%%%%%%%%%%%%%%%%%%%%%%%%%%%%%%%%%%%%%%%%%%%%%%%%%%%%
%%%%%%%%%%%                                           %%%%%%%%%%%%%%%%%%%%%%%%%%%%%%%%%%%%%%%%%%%%%%%%%%%%%%%%%%%
%%%%%%%%%%%             TASKS                    %%%%%%%%%%%%%%%%%%%%%%%%%%%%%%%%%%%%%%%%%%%%%%%%%%%%%%%%%%%
%%%%%%%%%%%                                           %%%%%%%%%%%%%%%%%%%%%%%%%%%%%%%%%%%%%%%%%%%%%%%%%%%%%%%%%%%
%%%%%%%%%%%%%%%%%%%%%%%%%%%%%%%%%%%%%%%%%%%%%%%%%%%%%%%%%%%%%%%%%%%%%%%%%%%%%%%%%%%%%%%%%%
%%%%%%%%%%%%%%%%%%%%%%%%%%%%%%%%%%%%%%%%%%%%%%%%%%%%%%%%%%%%%%%%%%%%%%%%%%%%%%%%%%%%%%%%%%

\section{Challenge tasks}
\label{sec:Tasks}

The goal of ILSVRC is to estimate the content of photographs for the purpose of retrieval and automatic annotation.
Test images are presented with no initial annotation, and algorithms  have to produce labelings specifying what objects are present in the images. New test images are collected and labeled especially for this competition and are not part of the previously published ImageNet dataset~\citep{ImageNet}. 

%%%%%%%%%%%%%%%%%%%%%%%%%%%%%%%%%%%%%%%%%%%%%%%%%%%%%%%%

\begin{table*}
\begin{tabular}{|p{1in}|m{1.5in} | >{\centering}m{1in} | >{\centering}m{1.4in} | >{\centering\arraybackslash}m{1in}  |}
\hline
\multicolumn{2}{|l|}{Task} & 
\parbox[t]{1in}{\centering {\bf Image \\ classification}} 
&
 \parbox[t]{1in}{\centering {\bf Single-object \\ localization}}
&
\parbox[t]{1in}{\centering {\bf Object \\ detection}} 
\\[3ex]
\hline
\multirow{2}{*}{\parbox[t]{1in}{Manual labeling \\ on training set}}
& Number of object classes annotated per image
&1
& 1
& 1 or more
\\
\cline{2-5}
&
\parbox[t]{1in}{Locations of \\ annotated classes}
& ---
& \parbox[t]{1in}{\centering all instances \\ on some images} 
& \parbox[t]{1in}{\centering all instances \\ on all images} 
\\
\hline
\multirow{2}{*}{\parbox[t]{1in}{Manual labeling \\ on validation \\ and test sets}} &
Number of object classes annotated per image
&1
& 1
& all target classes 
\\
\cline{2-5}
&
\parbox[t]{1in}{Locations of \\ annotated classes}
& ---
& \parbox[t]{1in}{\centering all instances \\ on all images} 
& \parbox[t]{1in}{\centering all instances \\ on all images} 
\\
\hline
\iffalse
\multirow{2}{*}{\parbox[t]{1in}{Labeling returned \\ by the ideal \\ algorithm}} &
\parbox[t]{1.5in}{Number of object class \\ labels per image}
& 1-5 
& 1-5
& all target classes \\
\cline{2-5}
&
\parbox[t]{1in}{Object locations  }
& ---
& \parbox[t]{1.5in}{\centering 1 instance per class label \\ on all images}
& \parbox[t]{1in}{\centering all instances \\ on all images} \\
\hline
\fi
\end{tabular}
\caption{Overview of the provided annotations for each of the tasks in ILSVRC.}
\label{table:tasks}
\end{table*}

%%%%%%%%%%%%%%%%%%%%%%%%%%%%%

ILSVRC  over the years has consisted of one or more of the following tasks (years in parentheses):\footnote{In addition, ILSVRC in 2012 also included a taster fine-grained classification task, where algorithms would classify dog photographs into one of 120 dog breeds~\citep{Khosla11}. Fine-grained classification has evolved into its own Fine-Grained classification challenge in 2013~\citep{FGcomp}, which is outside the scope of this paper.}
\begin{enumerate}
\item {\bf Image classification} (2010-2014): Algorithms produce a list of object categories present in the image. 
\item {\bf Single-object localization} (2011-2014): Algorithms produce a list of object categories present in the image, along with an axis-aligned bounding box indicating the position and scale of \emph{one} instance of each object category. 
\item {\bf Object detection} (2013-2014): Algorithms produce a list of object categories present in the image along with an axis-aligned bounding box indicating the position and scale of \emph{every} instance of each object category.
\end{enumerate}
This section provides an overview and history of each of the three tasks. Table~\ref{table:tasks} shows summary statistics.

%The image classification task is arguably the most basic recognition tasks, with a rich history including Caltech 101~\cite{Caltech101} and PASCAL VOC~\cite{PASCALVOC}. 

%\todo{there should be some discussion about why these tasks are used and about why it evolved in this way. Maybe contrast it with PASCAL. }

%The later Sections~\ref{sec:Annot} and \ref{sec:Evaluation} provide details about the annotation and evaluation procedures respectively.

\subsection{Image classification task}
\label{sec:TasksCls}

Data for the image classification task consists of photographs collected from Flickr\footnote{\url{www.flickr.com}} and other search engines, manually labeled with the presence of one of 1000 object categories. Each image contains one ground truth label. 

For each image, algorithms  produce a list of object categories present in the image. The quality of a labeling is evaluated based on the label that best matches the ground truth label for the image (see Section~\ref{sec:EvaluationCls}).

Constructing ImageNet was an effort to scale up an image classification dataset to cover most nouns in English using tens of millions of manually verified photographs~\citep{ImageNet}. The image classification task of ILSVRC came as a direct extension of this effort. A subset of categories and images was chosen and fixed to provide a standardized benchmark while the rest of ImageNet continued to grow. 

\subsection{Single-object localization task}
\label{sec:TasksLoc}

The single-object localization task, introduced in 2011, built off of the image classification task to evaluate the ability of algorithms to learn the appearance of the target object itself rather than its image context.

Data for the single-object localization task consists of the same photographs collected for the image classification task, hand
labeled with the presence of one of 1000 object categories.  Each image contains one ground truth label. Additionally, 
every instance of this category is annotated with an axis-aligned bounding box.

For each image, algorithms produce a list of object categories present in the image, along with a bounding box  
 indicating the position and scale of one instance of each object category. The quality of a labeling is evaluated based on
the object category label that best matches the ground truth label, with the additional requirement that the location of the predicted instance is also accurate (see Section~\ref{sec:EvaluationLoc}).

\subsection{Object detection task}
\label{sec:TasksDet}

The object detection task went a step beyond single-object localization and tackled the problem of localizing multiple object categories in the image. This task has been a part of the PASCAL VOC for many years on the scale of 20 object categories and tens of thousands of images, but scaling it up by an order of magnitude in object categories and in images proved to be very challenging from a dataset collection and annotation point of view (see Section~\ref{sec:AnnotDet}).

Data for the detection tasks consists of new photographs collected from Flickr using scene-level queries. The images
are annotated with axis-aligned bounding boxes indicating the position and scale of every instance of each target object category. 
The training set is additionally supplemented with (a) data from the single-object localization task, which contains annotations
for all instances of  just one object category, and (b) negative images known not to contain any instance of some object categories.

For each image, algorithms produce bounding boxes indicating the position and scale of all instances of all target
object categories. The quality of labeling is evaluated by \emph{recall}, or number of target object instances detected, and
\emph{precision}, or the number of spurious detections produced by the algorithm (see Section~\ref{sec:EvaluationDet}).

%%%%%%%%%%%%%%%%%%%%%%%%%%%%%%%%%%%%%%%%%%%%%%%%%%%%%%%%%%%%%%%%%%%%%%%%%%%%%%%%%%%%%%%%%%
%%%%%%%%%%%%%%%%%%%%%%%%%%%%%%%%%%%%%%%%%%%%%%%%%%%%%%%%%%%%%%%%%%%%%%%%%%%%%%%%%%%%%%%%%%
%%%%%%%%%%%                                           %%%%%%%%%%%%%%%%%%%%%%%%%%%%%%%%%%%%%%%%%%%%%%%%%%%%%%%%%%%
%%%%%%%%%%%        ANNOTATION              %%%%%%%%%%%%%%%%%%%%%%%%%%%%%%%%%%%%%%%%%%%%%%%%%%%%%%%%%%%
%%%%%%%%%%%                                           %%%%%%%%%%%%%%%%%%%%%%%%%%%%%%%%%%%%%%%%%%%%%%%%%%%%%%%%%%%
%%%%%%%%%%%%%%%%%%%%%%%%%%%%%%%%%%%%%%%%%%%%%%%%%%%%%%%%%%%%%%%%%%%%%%%%%%%%%%%%%%%%%%%%%%
%%%%%%%%%%%%%%%%%%%%%%%%%%%%%%%%%%%%%%%%%%%%%%%%%%%%%%%%%%%%%%%%%%%%%%%%%%%%%%%%%%%%%%%%%%

\section{Dataset construction at large scale}
\label{sec:Annot}

Our process of constructing large-scale object recognition image datasets consists of three key steps.

The first step is defining the set of target object categories.  To do this, we select from among the existing ImageNet~\citep{ImageNet} categories. By using WordNet as a backbone~\citep{WordNet}, ImageNet already takes care of disambiguating word meanings and of combining
together synonyms into the same object category. Since the selection of object categories needs to be done only
once per challenge task, we use a combination of automatic heuristics and manual post-processing to create the list of target
categories appropriate for each task. For example, for image classification we may include broader scene categories such as a type of beach,
but for single-object localization and object detection we want to focus only on object categories which can be unambiguously localized in images (Sections~\ref{sec:AnnotClsObjects} and~\ref{sec:AnnotDetObjects}).

The second step is collecting a diverse set of candidate images to represent the selected categories. We use both automatic and manual strategies on multiple search engines to do the image collection. The process is modified for the different ILSVRC tasks.
For example, for object detection we focus our efforts on collecting scene-like images using generic queries such as ``African safari'' to find pictures likely to contain multiple animals in one scene (Section~\ref{sec:AnnotDetImages}). 

The third (and most challenging) step is annotating the millions of collected images to obtain a clean dataset.
We carefully design crowdsourcing strategies targeted to each individual ILSVRC task. For example, the bounding box annotation system used for localization and detection tasks consists  of three distinct parts in order to include automatic crowdsourced quality control (Section~\ref{sec:AnnotLocBbox}).  Annotating images fully with all target object categories (on a reasonable budget) for object detection  requires an additional hierarchical image labeling system (Section~\ref{sec:AnnotDetList}).

We  describe the data collection and annotation procedure for each of the ILSVRC tasks in order: image classification (Section~\ref{sec:AnnotCls}), single-object localization (Section~\ref{sec:AnnotLoc}), and object detection (Section~\ref{sec:AnnotDet}), focusing on the three key steps for each dataset.

\begin{figure*}
\centering
\begin{tabular}{r}
\includegraphics[width=5.5in]{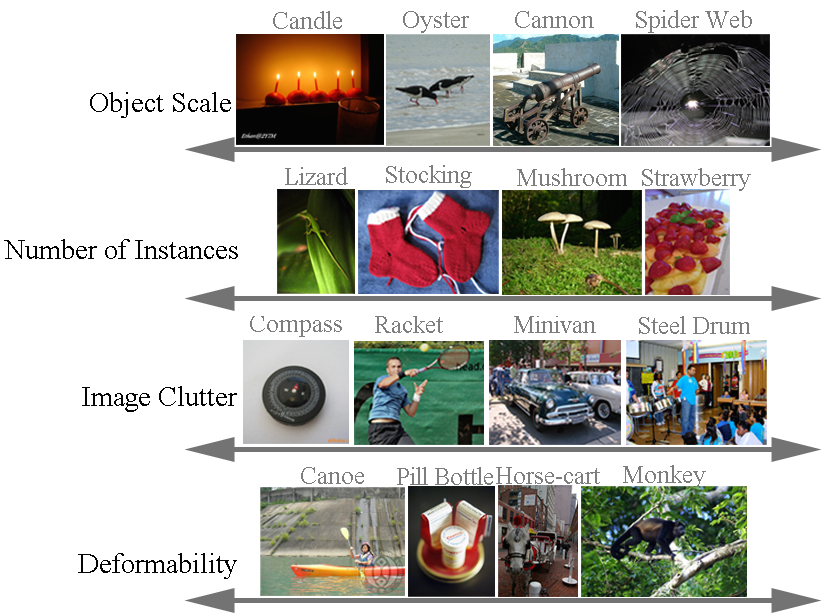}\\
\includegraphics[width=5.7in]{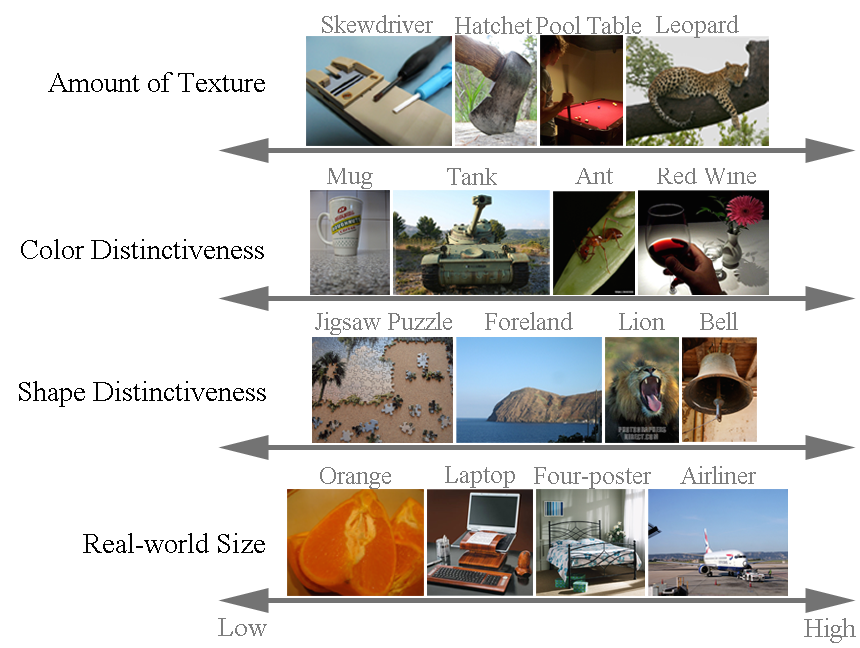}
\end{tabular}
\caption{The diversity of data in the ILSVRC image classification and single-object localization tasks. 
For each of the eight dimensions, we show example object categories along the range of that property.
Object scale, number of instances and image clutter for each object category are computed using the metrics defined in Section~\ref{sec:LocStats} and in Appendix~\ref{app:ICCV}. 
The other properties were computed by asking human subjects to annotate each of the 1000 object categories~\citep{Russakovsky13}.
} 
\label{fig:diversity}
\end{figure*}

%%%%%%%%%%%%%%%%%%%%%%%%%%%%%%%%%%%%%%%%%%%%%%%%%%%%%%%%%%%%%%%%%%%%%%%%%%%%%%%%%%%%%%%%%%
%%%%%%%%%%%                                                            %%%%%%%%%%%%%%%%%%%%%%%%%%%%%%%%%%%%%%%%%%%%%%%%%%%
%%%%%%%%%%%        Classification annotation              %%%%%%%%%%%%%%%%%%%%%%%%%%%%%%%%%%%%%%%%%%%%%%%%%%%
%%%%%%%%%%%                                          	             %%%%%%%%%%%%%%%%%%%%%%%%%%%%%%%%%%%%%%%%%%%%%%%%%%%
%%%%%%%%%%%%%%%%%%%%%%%%%%%%%%%%%%%%%%%%%%%%%%%%%%%%%%%%%%%%%%%%%%%%%%%%%%%%%%%%%%%%%%%%%%

\subsection{Image classification dataset construction}
\label{sec:AnnotCls}

The image classification task tests the ability of an algorithm to name the objects present in the image, without
necessarily localizing them. 

We describe the choices we made in constructing the ILSVRC image classification dataset: selecting the target
object categories from ImageNet (Section~\ref{sec:AnnotClsObjects}), collecting a diverse set of candidate images by using
multiple search engines and an expanded set of queries in multiple languages (Section~\ref{sec:AnnotClsImages}), and
finally filtering the millions of collected images using the carefully designed crowdsourcing strategy of ImageNet~\citep{ImageNet} (Section~\ref{sec:AnnotClsAnnot}). %We conclude with statistics of the collected dataset in Section~\ref{sec:ClsStats}.

\subsubsection{Defining object categories for the image classification dataset}
\label{sec:AnnotClsObjects}

The 1000 categories used for the image classification task were selected from the ImageNet~\citep{ImageNet} categories. The 1000 synsets are selected such that there is no overlap between synsets: for any synsets $i$ and $j$, $i$ is not an ancestor of $j$ in the ImageNet hierarchy. These synsets are part of the larger hierarchy and may have children in ImageNet; however, for ILSVRC we do not consider their child subcategories. The synset hierarchy of ILSVRC  can be thought of as a  ``trimmed'' version of the
complete ImageNet hierarchy. Figure~\ref{fig:diversity} visualizes the diversity of the ILSVRC2012 object categories. 

%Figure~\ref{fig:clsclasses} shows the synset hierarchy
%of ILSVRC2012. 

\begin{figure*}
\begin{center}
\begin{tabular}{cc|c}
&{ PASCAL} & { ILSVRC} \\
\side{1}{ birds} &
\includegraphics[height=1in]{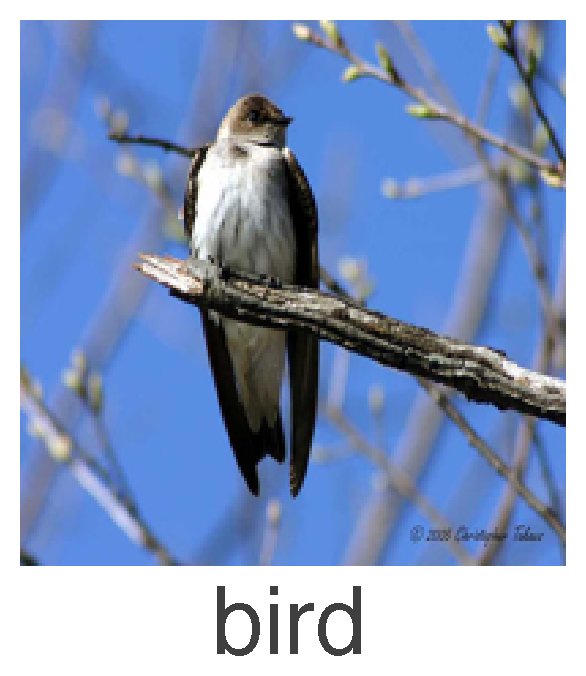} &
\includegraphics[height=1in]{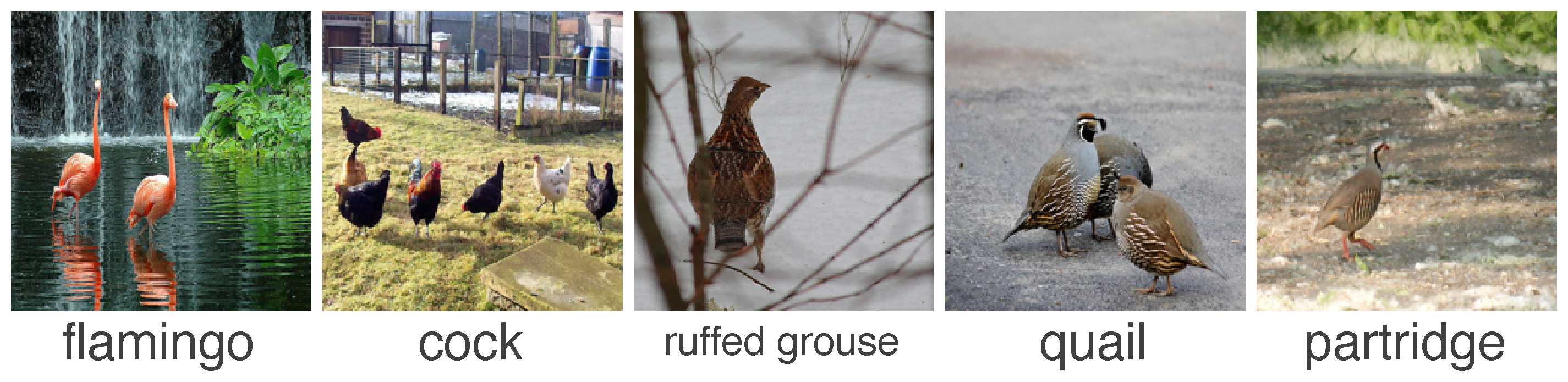} $\cdots$\\
\side{1}{ cats} &
\includegraphics[height=1in]{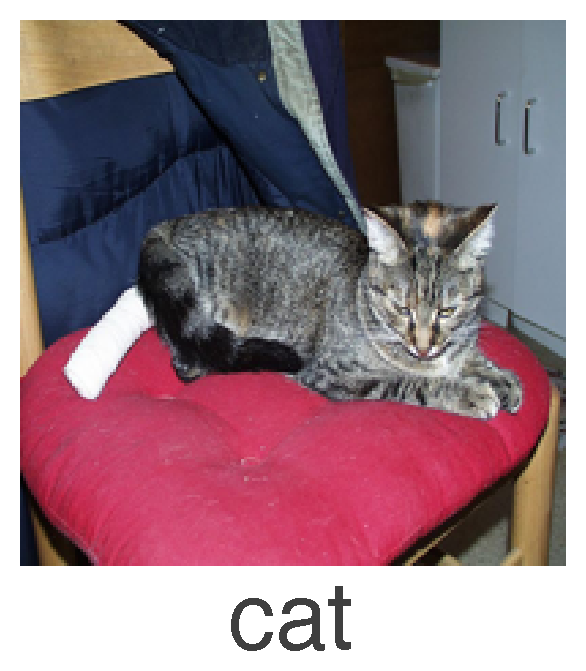} &
\includegraphics[height=1in]{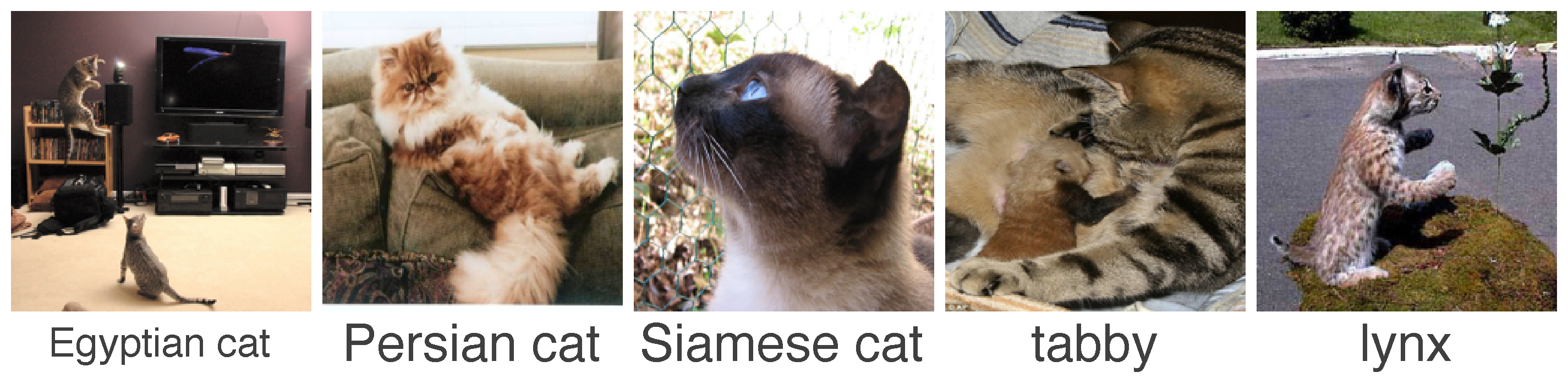} $\cdots$\\
\side{1}{ dogs} &
\includegraphics[height=1in]{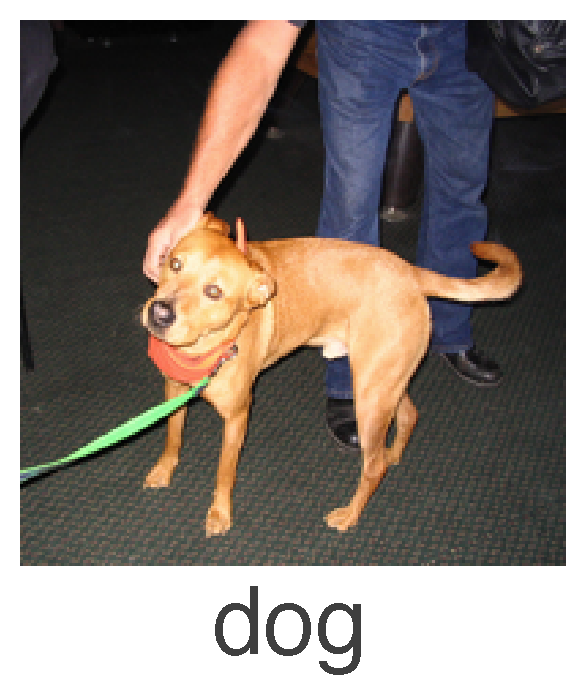} &
\includegraphics[height=1in]{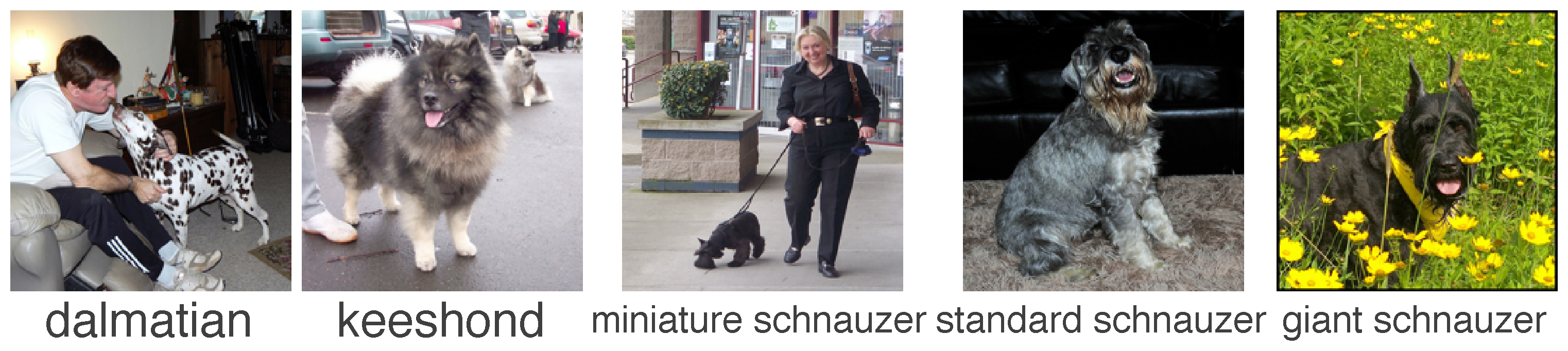} $\cdots$\\
\end{tabular}
\end{center}
\caption{The ILSVRC dataset contains many more fine-grained classes compared to the standard PASCAL VOC benchmark; for example, instead of the PASCAL ``dog'' category there are 120 different breeds of dogs in ILSVRC2012-2014 classification and single-object localization tasks.}
\label{fig:fg}
\end{figure*}

The exact 1000 synsets used for the image classification and single-object localization tasks have changed over the years. There are 639 synsets
which have been used in all five ILSVRC challenges so far. 
In the first year of the challenge synsets were selected randomly from the available ImageNet synsets at the time, followed by manual filtering to make sure the
object categories were not too obscure. With the introduction of the object localization challenge in 2011 there were 321 synsets that changed: categories such as ``New Zealand beach''
which were inherently difficult to localize were removed, and some new categories from ImageNet containing object localization annotations were added.
In ILSVRC2012,  90 synsets were replaced with categories corresponding to dog breeds to allow for evaluation of more fine-grained object classification, as shown in Figure~\ref{fig:fg}.
The synsets have remained consistent since year 2012. Appendix~\ref{app:classes} provides the complete list of object categories used in ILSVRC2012-2014.

\iffalse
\begin{table*}
{\tiny
\input{cls_hierarchy.tex}
}
\caption{Classification hierarchy}
\end{table*}

\begin{figure}
\includegraphics[width=\linewidth]{hierarchy2012_rename4_cropped.png} \\
\caption{Cropped ImageNet hierarchy of 1000 object classes used in the ILSVRC2012-2014 object classification and single-object localization tasks.
The leaf nodes (corresponding to classes used in the challenge) are black; the internal nodes are light blue. Best viewed under magnification. \todo{make readable}
}
\label{fig:clsclasses}
\end{figure}
\fi

%The ILSVRC
%dataset provides an unprecedented challenge for algorithms by having many ﬁne-grained classes, e.g., three
%schnauzer breeds in Figure~\ref{fig:fg}. 

\subsubsection{Collecting candidate images for the image classification dataset}
\label{sec:AnnotClsImages}

Image collection for ILSVRC classification task is the same as the strategy employed for constructing
 ImageNet~\citep{ImageNet}. Training images are taken directly from ImageNet. Additional
images are collected for the ILSVRC using this strategy and randomly partitioned into the validation and test sets.

We briefly summarize the process; \citep{ImageNet} contains further details. Candidate images are collected from the Internet by querying several image search engines. For each synset, the queries are the set of WordNet synonyms. Search engines typically limit the number of retrievable images (on the order of a few hundred to a thousand). To obtain as many images as possible, we expand the query set by appending the queries with the word from parent synsets, if the same word appears in the glossary of the target synset. For example, when querying ``whippet'', according to WordNet's glossary a ``small slender dog of greyhound type developed in England'', we also use ``whippet dog'' and ``whippet greyhound.''
To further enlarge and diversify the candidate pool, we translate the queries into other languages, including Chinese, Spanish, Dutch and Italian. We obtain accurate translations using WordNets in those languages.

\subsubsection{Image classification dataset annotation}
\label{sec:AnnotClsAnnot}

Annotating images with corresponding object classes follows the strategy employed by ImageNet~\citep{ImageNet}. We summarize it briefly here.

To collect a highly accurate dataset, we rely on humans to verify each candidate image collected in the previous step for a given synset. This is achieved by using Amazon Mechanical Turk (AMT), an online platform on which one can put up tasks for users for a monetary reward. With a global user base, AMT is particularly suitable for large scale labeling.
In each of our labeling tasks, we present the users with a set of candidate images and the definition of the target synset (including a link to Wikipedia). We then ask the users to verify whether each image contains objects of the synset. We encourage users to select images regardless of occlusions, number of objects and clutter in the scene to ensure diversity.

While users are instructed to make accurate judgment, we need to set up a quality control system to ensure this accuracy. There are two issues to consider. First, human users make mistakes and not all users follow the instructions. Second, users do not always agree with each other, especially for more subtle or confusing synsets, typically at the deeper levels of the tree. The solution to these issues is to have multiple users independently label the same image. An image is considered positive only if it gets a convincing majority of the votes. We observe, however, that different categories require different levels of consensus among users. For example, while five users might be necessary for obtaining a good consensus on “Burmese cat” images, a much smaller number is needed for “cat” images. We develop a simple algorithm to dynamically determine the number of agreements needed for different categories of images. For each synset, we first randomly sample an initial subset of images. At least 10 users are asked to vote on each of these images. We then obtain a confidence score table, indicating the probability of an image being a good image given the consensus among user votes. For each of the remaining candidate images in this synset, we proceed with the AMT user labeling until a pre-determined confidence score threshold is reached.

\paragraph{Empirical evaluation.}  Evaluation of the accuracy of the large-scale crowdsourced image annotation system was done on the entire ImageNet~\citep{ImageNet}. 
A total of 80 synsets were randomly sampled at every tree depth of
the mammal and vehicle subtrees. An independent group of subjects verified the correctness of each of the images. An average of $99.7\%$ precision is achieved across the synsets. We expect similar accuracy on ILSVRC image classification dataset since the image annotation pipeline has remained the same. To verify, we manually checked 1500 ILSVRC2012-2014 image classification test set images (the test set has remained unchanged in these three years). We found 5 annotation errors, corresponding as expected to $99.7\%$ precision.

\subsubsection{Image classification dataset statistics}
\label{sec:ClsStats}

Using the image collection and annotation procedure described in previous sections, we collected a large-scale dataset
used for ILSVRC classification task. There are 1000 object classes and approximately 1.2 million training images, 50 thousand validation images and 100 thousand test images. Table~\ref{table:clslocstats} (top) documents the size of the dataset over the years of the challenge. 

%%%%%%%%%%%%%%%%%%%%%%%%%%%%%%%%%%%%%%%%%%%%%%%%%%%%%
\begin{table*}

\begin{center} {\bf Image classification annotations (1000 object classes)} \end{center}

\begin{tabular}{|p{0.9in}|c|c|c|c|}
\hline
Year
 %&
%\parbox[t]{1in}{\centering {Object classes}}
&
\parbox[t]{2.3in}{\centering {Train images  (per class)}}
&
\parbox[t]{1.95in}{\centering {Val images  (per class)}}
&
\parbox[t]{1.25in}{\centering {Test images  (per class)}}
\\
\hline
ILSVRC2010 
&
1,261,406  (668-3047)
&
50,000  (50)
&
150,000  (150)
\\
\hline
ILSVRC2011
&
1,229,413 (384-1300)
&
50,000  (50)
&
100,000  (100)
\\
\hline
ILSVRC2012-14 
&
1,281,167  (732-1300)
&
 50,000  (50)
&
 100,000  (100)
\\
\hline
\end{tabular}

\begin{center}
{\bf Additional annotations for single-object localization (1000 object classes)}
\end{center}

\begin{tabular}{|p{0.9in}|c|c|c|c|c|c|}
\hline
Year
 %&
%\parbox[t]{0.4in}{\centering {Object \\ classes}}
&
\parbox[t]{1.1in}{\centering {Train images  with \\ bbox annotations \\ (per class)}}
&
\parbox[t]{1in}{\centering {Train bboxes\\ annotated \\ (per class)}}
&
\parbox[t]{0.95in}{\centering {Val images   with \\ bbox annotations \\ (per class)}}
&
\parbox[t]{0.8in}{\centering {Val bboxes \\annotated \\ (per class)}}
&
\parbox[t]{1.25in}{\centering {Test images  with \\ bbox annotations}}
\\
\hline
ILSVRC2011
& 315,525 (104-1256)
& 344,233 (114-1502)
& 50,000 (50)
& 55,388 (50-118)
& 100,000
\\
\hline
ILSVRC2012-14 
& 523,966 (91-1268)
& 593,173 (92-1418)
& 50,000 (50)
& 64,058 (50-189)
& 100,000
\\
\hline
\end{tabular}

\caption{Scale of ILSVRC image classification task (top) and single-object localization task (bottom). The numbers in parentheses correspond to (minimum per class - maximum per class). The 1000 classes change from year to year but are consistent between image classification and single-object localization tasks in the same year. All images from the image classification task may be used
for single-object localization.}
\label{table:clslocstats}
\end{table*}

%%%%%%%%%%

\iffalse
\begin{table}
\begin{tabular}{|p{0.8in}|c|c|c|}
\hline
Year & ILSVRC2010 & ILSVRC2011 & ILSVRC2012-14 \\
\hline
Object classes & 1000 & 1000 & 1000  \\
\hline
\parbox[t]{0.8in}{Train images \\ (per class)} &
\parbox[t]{0.7in}{\centering 1,261,406 \\ (668-3047)} &
\parbox[t]{0.7in}{\centering 1,229,413 \\ (384-1300)} &
\parbox[t]{0.7in}{\centering 1,281,167 \\ (732-1300)} 
 \\
\hline
\parbox[t]{0.8in}{Val images \\ (per class)} &
 \parbox[t]{0.7in}{\centering 50,000 \\ (50)} &
 \parbox[t]{0.7in}{\centering 50,000 \\ (50)} & 
 \parbox[t]{0.7in}{\centering 50,000 \\ (50)} 
\\
\hline
\parbox[t]{0.8in}{Test images \\ (per class)} &
 \parbox[t]{0.7in}{\centering 150,000 \\ (150)} &
 \parbox[t]{0.7in}{\centering 100,000 \\ (100)} &
 \parbox[t]{0.7in}{\centering 100,000 \\ (100)} 
\\
\hline
\end{tabular}
\fi

%%%%%%%%%%%%%%%%%%%%%%%%%%%%%%%%%%%%%%%%%%%%%%%%%%%%%

%%%%%%%%%%%%%%%%%%%%%%%%%%%%%%%%%%%%%%%%%%%%%%%%%%%%%%%%%%%%%%%%%%%%%%%%%%%%%%%%%%%%%%%%%%
%%%%%%%%%%%                                                            %%%%%%%%%%%%%%%%%%%%%%%%%%%%%%%%%%%%%%%%%%%%%%%%%%%
%%%%%%%%%%%        Localization annotation                %%%%%%%%%%%%%%%%%%%%%%%%%%%%%%%%%%%%%%%%%%%%%%%%%%%
%%%%%%%%%%%                                          	             %%%%%%%%%%%%%%%%%%%%%%%%%%%%%%%%%%%%%%%%%%%%%%%%%%%
%%%%%%%%%%%%%%%%%%%%%%%%%%%%%%%%%%%%%%%%%%%%%%%%%%%%%%%%%%%%%%%%%%%%%%%%%%%%%%%%%%%%%%%%%%

\subsection{Single-object localization dataset construction}
\label{sec:AnnotLoc}

The single-object localization task evaluates the ability of an algorithm to localize one instance of an object category.
It was introduced as a  taster task in ILSVRC 2011, and became an official part of ILSVRC in 2012.

The key challenge
was developing a scalable crowdsourcing method for object bounding box annotation. Our three-step self-verifying
pipeline is described in Section~\ref{sec:AnnotLocBbox}. Having the dataset collected, we perform detailed analysis in Section~\ref{sec:LocStats}
to ensure that the dataset is sufficiently varied to be suitable for evaluation of object localization algorithms.

\paragraph{Object classes and candidate images.}
The object classes for single-object localization task are the same as the object classes for image classification task described above in Section~\ref{sec:AnnotCls}. The training
images for localization task are a subset of the training images used for image classification task, and the validation and test
images are the same between both tasks. 

%Details on image collection and on annotation of each image
%with one object class are in Section~\ref{sec:AnnotClsImages} and Section~\ref{sec:AnnotClsAnnot} respectively.

%In this section we discuss the bounding box annotation procedure, and provide statistics of the single-object localization dataset.

\paragraph{Bounding box annotation.} Recall that for the image classification task every image was annotated with one object class label, corresponding to
one object that is present in an image. For the single-object localization task, every validation and test image and a subset of the training images
are annotated with axis-aligned bounding boxes around every instance of this object.

Every bounding box is required to be as small as possible while including all visible parts of the object instance. An alternate annotation procedure could be to annotate the \emph{full (estimated) extent} of the object: e.g., if a person's legs are occluded and only the torso is visible, the bounding box could be drawn to include the likely location of the legs. However, this alternative procedure is inherently ambiguous and ill-defined, leading to disagreement among annotators and among researchers (what is the true ``most likely'' extent of this object?). We follow the standard protocol of only annotating visible object parts~\citep{LabelMe,PASCALIJCV}.\footnote{Some datasets such as PASCAL VOC~\citep{PASCALIJCV} and LabelMe~\citep{LabelMe} are able to provide more detailed annotations: for example, marking individual object instances as being \emph{truncated}. We chose not to provide this level of detail in favor of annotating more images and more object instances.}

\subsubsection{Bounding box object annotation system}
\label{sec:AnnotLocBbox}

We summarize the crowdsourced bounding box annotation system described in detail in~\citep{Su12}. The goal is to build a system that is fully automated, highly accurate, and cost-effective. Given a collection of images where the object of interest has been verified to exist, for each image the system collects a tight bounding box for every instance of the object. 

There are two requirements:

\begin{itemize}
\item {\bf Quality} Each bounding box needs to be tight, i.e. the smallest among all bounding boxes that contains all visible parts of the object. This facilitates the object detection learning algorithms by providing the precise location of each object instance;
\item {\bf Coverage} Every object instance needs to have a bounding box. This is important for training localization algorithms because it tells the learning algorithms with certainty what is not the object. 
\end{itemize}

The core challenge of building such a system is  effectively controlling the data quality with minimal cost. Our key observation is that drawing a bounding box is significantly more difficult and time consuming than giving answers to multiple choice questions. Thus quality control through additional verification tasks is more cost-effective than consensus-based algorithms. This leads to the following workflow with simple basic subtasks:

\begin{enumerate}
\item {\bf Drawing} A worker draws one bounding box around one instance of an object on the given image.
\item {\bf Quality verification} A second worker checks if the bounding box is correctly drawn.
\item {\bf Coverage verification} A third worker checks if all object instances have bounding boxes.
\end{enumerate}

 The sub-tasks are designed following two principles. First, the tasks are made as simple as possible. For example, instead of asking the worker to draw all bounding boxes on the same image, we ask the worker to draw only one. This reduces the complexity of the task. Second, each task has a fixed and predictable amount of work. For example, assuming that the input images are clean (object presence is correctly verified) and the coverage verification tasks give correct results, the amount of work of the drawing task is always that of providing exactly one bounding box.

Quality control on Tasks 2 and 3 is implemented by embedding ``gold standard'' images where the correct answer is known. Worker training for each of these subtasks is described in detail in~\citep{Su12}.

\paragraph{Empirical evaluation.} The system is evaluated on 10 categories with ImageNet~\citep{ImageNet}:  balloon, bear, bed, bench, beach, bird, bookshelf, basketball hoop, bottle, and people. A subset of 200 images are randomly sampled from each category. On the image level, our evaluation shows that $97.9\%$ images are completely covered with bounding boxes. For the remaining $2.1\%$, some bounding boxes are missing. However, these are all difficult cases: the size is too small, the boundary is blurry, or there is strong shadow.

On the bounding box level, $99.2\%$ of all bounding boxes are accurate (the bounding boxes are visibly tight). The remaining $0.8\%$ are somewhat off. No bounding boxes are found to have less than $50\%$ intersection over union overlap with ground truth.

Additional evaluation of the overall cost and an analysis of quality control can be found in~\citep{Su12}.

%\todo{ annotate all (up to N) instances in the image with bboxes; examples critical for more complicated objects; pictures %from Hao's paper}

\subsubsection{Single-object localization dataset statistics}
\label{sec:LocStats}

Using the annotation procedure described above, we collect a large set of bounding box annotations
for the ILSVRC single-object classification task. All 50 thousand images in the validation set and
100 thousand images in the test set are annotated with bounding boxes around all instances of the
ground truth object class (one object class per image). In addition, in ILSVRC2011 
$25\%$ of training images are annotated with bounding boxes the same way, yielding more than 310 thousand
annotated images with more than 340 thousand annotated object instances. In ILSVRC2012 $40\%$ of
training images are annotated, yielding more than 520 thousand 
annotated images with more than 590 thousand annotated object instances. 
Table~\ref{table:clslocstats} (bottom) documents the size of this dataset.

In addition to the size of the dataset, we also analyze the level of difficulty of object localization
in these images compared to the PASCAL VOC benchmark. We compute statistics on the ILSVRC2012 single-object localization
validation set images compared to PASCAL VOC 2012 validation images. 

Real-world scenes are likely to contain multiple instances of some objects, and nearby object instances are particularly difficult to delineate.  The average object category in ILSVRC has $1.61$ target object instances on average per positive image, with each instance having on average $0.47$ neighbors (adjacent instances of the same object category). This is comparable to $1.69$ instances per positive image and $0.52$ neighbors per instance for an average object class in PASCAL. 

As described in~\citep{Hoiem12}, smaller objects tend to be significantly more difficult to localize. In the average object category in PASCAL
the object occupies $24.1\%$ of the image area, and in ILSVRC $35.8\%$. However, PASCAL has only 20 object
categories while ILSVRC has 1000. The 537 object categories of ILSVRC with the smallest objects on average occupy the same
fraction of the image as PASCAL objects: $24.1\%$. Thus even though on average the object instances tend to be bigger in ILSVRC images,
there are more than 25 times more object categories than in PASCAL VOC with the same average object scale. 

Appendix~\ref{app:ICCV} and~\citep{Russakovsky13} have additional comparisons.

%%%%%%%%%%%%%%%%%%%%%%%%%%%%%%%%%%%%%%%%%%%%%%%%%%%%%%%%%%%%%%%%%%%%%%%%%%%%%%%%%%%%%%%%%%
%%%%%%%%%%%                                                            %%%%%%%%%%%%%%%%%%%%%%%%%%%%%%%%%%%%%%%%%%%%%%%%%%%
%%%%%%%%%%%        Detection annotation                   %%%%%%%%%%%%%%%%%%%%%%%%%%%%%%%%%%%%%%%%%%%%%%%%%%%
%%%%%%%%%%%                                          	             %%%%%%%%%%%%%%%%%%%%%%%%%%%%%%%%%%%%%%%%%%%%%%%%%%%
%%%%%%%%%%%%%%%%%%%%%%%%%%%%%%%%%%%%%%%%%%%%%%%%%%%%%%%%%%%%%%%%%%%%%%%%%%%%%%%%%%%%%%%%%%

\subsection{Object detection dataset construction}
\label{sec:AnnotDet}

The ILSVRC task of object detection evaluates the ability of an algorithm to name and localize \emph{all} instances
of \emph{all} target objects present in an image. It is much more challenging than object localization because
some object instances may be small/occluded/difficult to accurately localize, and the algorithm is expected
to locate them all, not just the one it finds easiest. 

%The image collection and annotation procedures for this task are different than for either image classification or
%single-object localization. The images are chosen to be more scene-like, and annotating in a straight-forward
%way by creating a task for every (image, object class) pair is no longer feasible.

There are three key challenges in collecting the object detection dataset.
The first challenge is selecting the set of common objects which tend to appear in cluttered photographs
and are well-suited for benchmarking object detection performance. Our approach relies 
on statistics of the object localization dataset and the tradition of the
PASCAL VOC challenge (Section~\ref{sec:AnnotDetObjects}). 

The second challenge is obtaining a much more varied set of scene images than those used for
the image classification and single-object localization datasets. Section~\ref{sec:AnnotDetImages} describes the
procedure for utilizing as much data from the single-object localization dataset as possible and supplementing it with
Flickr images queried using hundreds of manually designed high-level queries.

The third, and biggest, challenge is 
completely annotating this dataset with all the objects. 
This is done in two parts. Section~\ref{sec:AnnotDetList} describes the first part: 
our hierarchical strategy for obtaining the list of all target objects which occur within every image. This is necessary since annotating in a straight-forward way by creating a task for every (image, object class) pair is no longer feasible at this scale. 
Appendix~\ref{sec:AppDetBbox} describes the second part: annotating the bounding boxes around these objects, using the single-object localization bounding box annotation pipeline of Section~\ref{sec:AnnotLocBbox} along with extra verification to ensure that \emph{every} instance of the object is annotated with exactly \emph{one} bounding box.

\subsubsection{Defining object categories for the object detection dataset}
\label{sec:AnnotDetObjects}

\begin{table}[t]
\centering
\begin{tabular}{|l|l| c| c|}
\hline
Class name in &  Closest class in & \multicolumn{2}{c|}{Avg object scale (\%)} \\
\cline{3-4}
PASCAL VOC & ILSVRC-DET & PASCAL & ILSVRC- \\
{\bf (20 classes)} & {\bf (200 classes)} & VOC & DET\\

%{\bf PASCAL VOC (20) } & {\bf Closest ILSVRC-DET (200)} \\
%PASCAL VOC class & Closest of ILSVRC-DET classes \\
\hline
aeroplane & airplane & 29.7 & 22.4\\
\hline
bicycle & bicycle & 29.3 & 14.3 \\
\hline
bird & bird & 15.9 & 20.1 \\
\hline
\emph{boat} & \emph{watercraft} & 15.2 & 16.5 \\
\hline
\emph{bottle} & \emph{wine bottle} & 7.3 & 10.4\\
\hline
bus & bus & 29.9 & 22.1 \\
\hline
car & car & 14.0 & 13.4\\
\hline
cat & domestic cat & 46.8 & 29.8 \\
\hline
chair & chair  & 12.8 & 10.1 \\
\hline
\emph{cow} & \emph{cattle} & 19.3 & 13.5 \\
\hline
\emph{dining table} & \emph{table} & 29.1 & 30.3 \\
\hline
dog & dog & 37.0 & 28.9 \\
\hline
horse & horse & 29.5 & 18.5 \\
\hline
motorbike & motorcyle& 32.0 & 20.7 \\
\hline
person & person & 17.5 & 19.3 \\
\hline
\emph{potted plant} & \emph{flower pot} & 12.3 & 8.1\\
\hline
sheep & sheep & 12.2 & 17.3 \\
\hline
sofa & sofa & 41.7 & 44.4 \\
\hline
train & train  & 35.4 & 35.1\\
\hline
tv/monitor & tv or monitor & 14.6 & 11.2\\
\hline
\end{tabular}
\caption{Correspondences between the object classes in the PASCAL VOC~\citep{PASCALIJCV} and the ILSVRC detection task.
Object scale is the fraction of image area (reported in percent) occupied by an object instance. It is computed on the validation sets of
PASCAL VOC 2012 and of ILSVRC-DET. The average object scale is $24.1\%$  across the 20 PASCAL VOC categories 
and $20.3\%$ across the 20 corresponding ILSVRC-DET categories. Section~\ref{sec:AnnotDetStats} reports additional dataset statistics.
%In addition, ILSVRC-DET has 10x more object classes than PASCAL VOC 
%and 3.5x more training instances on average per class (680 in PASCAL VOC 2012 versus 2394 in ILSVRC-DET 2014). 
}
\label{table:detclassespascal}
\end{table}

There are 200 object classes hand-selected for the detection task, eacg corresponding to a synset within ImageNet. 
These were chosen to be mostly basic-level object categories that would be easy for
people to identify and label. The rationale is that the object detection system developed for this task
can later be combined with a fine-grained classification model to further classify the objects if a finer
subdivision is desired.\footnote{Some of the training objects are actually annotated with more detailed classes:
for example, one of the 200 object classes is the category ``dog,'' and some training instances
are annotated with the specific dog breed.}
As with the 1000 classification classes, the synsets are selected such that there is no overlap:  for any synsets $i$ and $j$, $i$ is not an ancestor of $j$ in the ImageNet hierarchy. 

\iffalse 
\begin{figure}
\includegraphics[width=\linewidth]{detannotpipeline.png}
\caption{Selection of 200 object classes for the detection task.}
\label{fig:detannot}
\end{figure}
\fi

The selection of the 200 object detection classes in 2013 was guided by the ILSVRC 2012 classification and localization dataset.
 Starting with 1000 object classes and their bounding box annotations we
first eliminated all object classes which tended to be too ``big'' in the image (on average the object area was greater than $50\%$ of the
image area). These were classes such as T-shirt, spiderweb, or manhole cover. We then manually eliminated all classes
which we did not feel were well-suited for detection, such as hay, barbershop, or poncho. This left 494 object classes
which were merged into basic-level categories: for example, different species of birds were merged into just the ``bird'' class.
The classes remained the same in ILSVRC2014.
Appendix~\ref{app:hierarchy} contains the complete list of object categories used in ILSVRC2013-2014 (in the context of the hierarchy described in Section~\ref{sec:AnnotDetList}).

\iffalse

\begin{figure}
\caption{200 object classes of the object detection task. \todo{THIS FIGURE}}
\label{fig:detclasses}
\end{figure}
\fi
Staying mindful of the tradition of the PASCAL VOC dataset we also tried to ensure that the set of 200 classes contains
as many of the 20 PASCAL VOC classes as possible. Table~\ref{table:detclassespascal} shows the correspondences. 
The changes that were done were to ensure more accurate and consistent crowdsourced annotations.
The object class with the
weakest correspondence is ``potted plant'' in PASCAL VOC, corresponding to  ``flower pot''  in ILSVRC. ``Potted plant'' was  one
of the most challenging object classes to annotate consistently among the PASCAL VOC classes, and in order to obtain accurate  annotations using crowdsourcing we had to restrict the definition to a more concrete object.

\subsubsection{Collecting images for the object detection dataset}
\label{sec:AnnotDetImages}

\begin{figure}[t]
\includegraphics[width=\linewidth]{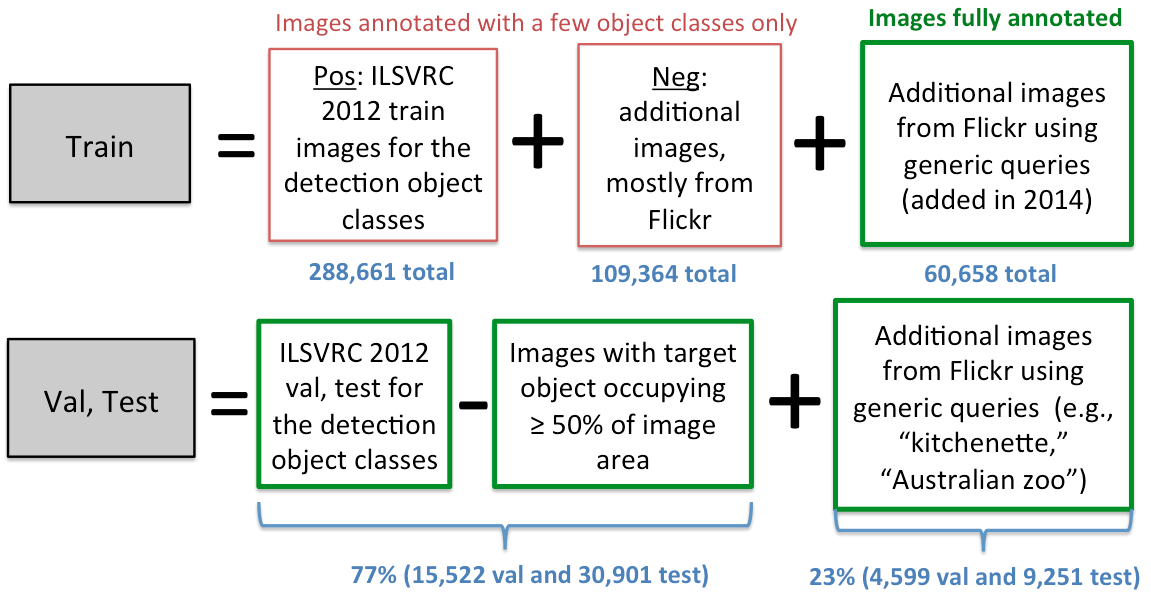}
\caption{Summary of images collected for the detection task. Images in green (bold) boxes have all instances of all 200 detection object classes fully annotated. Table~\ref{table:detstats} lists the complete statistics.}
\label{fig:detsummary}
\end{figure}

Many images for the detection task were collected differently than the images in ImageNet and the classification and single-object
localization tasks. Figure~\ref{fig:detsummary} summarizes the types of images that were collected. Ideally all of these images would be scene images fully annotated with all target categories. However, given budget constraints our goal was to provide as much suitable detection data as possible, even if the images were drawn from a few different sources and distributions.

The validation and test detection set images come from two sources (percent of images from each source in parentheses).
The first source $(77\%)$ is images from ILSVRC2012 single-object localization validation and test sets
corresponding to the 200 detection classes (or their children in the ImageNet hierarchy). 
Images where the target object occupied more than $50\%$ of the image area were discarded, since they were unlikely to contain
other objects of interest. The second source $(23\%)$ is images from Flickr collected specifically for detection task. We queried Flickr using a large set of manually defined queries, such as ``kitchenette'' or ``Australian zoo'' to retrieve images of scenes likely to contain
several objects of interest. Appendix~\ref{sec:AppQueries} contains the full list. We also added pairwise queries, or queries with two target object names such as ``tiger lion,'' which also often returned cluttered
scenes. 

\begin{figure*}
\includegraphics[width=\linewidth]{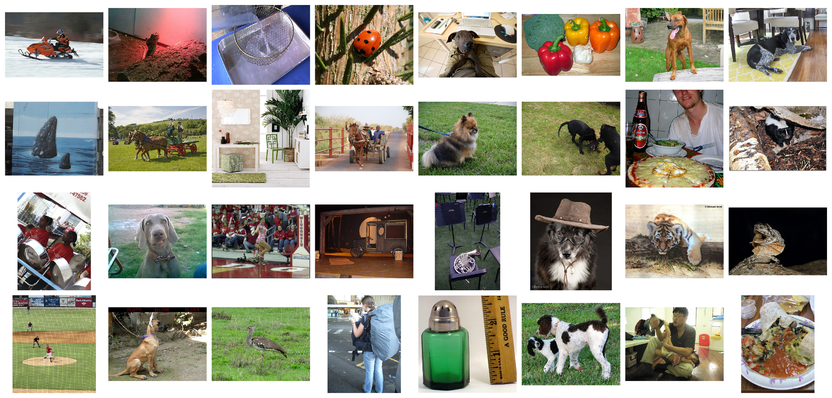}
\includegraphics[width=\linewidth]{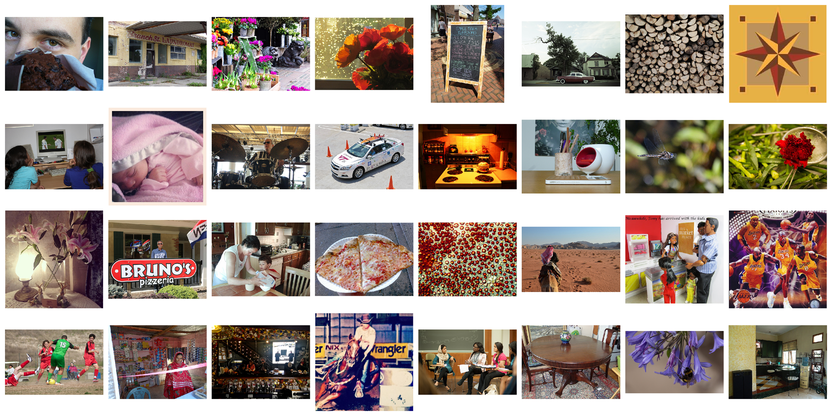}
\caption{Random selection of images in ILSVRC detection validation set. The images in the top 4 rows were taken from ILSVRC2012 single-object localization
validation set, and the images in the bottom 4 rows were collected from Flickr using scene-level queries.}
\label{fig:detimgs}
\end{figure*}

 Figure~\ref{fig:detimgs} shows a random set of both types of validation images. Images were randomly split, with $33\%$ going into the validation set and $67\%$ into the test set.\footnote{The validation/test split is consistent with ILSVRC2012:
validation images of ILSVRC2012 remained in the validation set of ILSVRC2013, and ILSVRC2012 test images remained in ILSVRC2013 test set.}
% In total there are $20,121$ validation and $40,152$ test images.

The training set for the detection task comes from three sources of images (percent of images from each source in parentheses). 
The first source $(63\%)$ is all training images from ILSVRC2012 single-object localization task corresponding
to the 200 detection classes (or their children in the ImageNet hierarchy). We did not filter by object size, allowing teams to take advantage of all the
positive examples available. The second source $(24\%)$ 
is negative images which were part of the original ImageNet collection process but voted as negative:
for example, some of the images were collected from Flickr and search engines for the ImageNet synset ``animals'' but during the manual
verification step did not collect enough votes to be considered as containing an ``animal.'' These images were manually re-verified for the detection task 
to ensure that they did not in fact contain the target objects.  The third source $(13\%)$ is images collected from Flickr specifically for the detection task. These images were added for ILSVRC2014 following the same protocol
as the second type of images in the validation and test set. This was done to bring the training and testing distributions closer together.

%In total there are 288,661 training images from the ILSVRC2012 single-object localization task (between 417 and 66,991 per class), 109,364 additional 
%annotated negative training images (185-10,073 per class) and 60,658 Flickr scene images added in ILSVRC2014.

%One of the big challenges in collecting images for the detection task was getting enough diversity in the data. For example, for small hand-held tools
%such as screwdrivers the photos tend to either be closeup ``product'' shots of the objects or scenes where the object is barely visible \todo{Figure?}.

%\todo{collection in ILSVRC2013 vs 2014: 2013 rely heavily on cluttered pictures from localization dataset + collect more from Flickr  using pairwise queries; training distribution different from test distribution which is bad; %2014 entirely collected from Flickr; reaching limits of images with small cluttered screwdrivers}

\subsubsection{Complete image-object annotation for the object detection dataset}
\label{sec:AnnotDetList}

The key challenge in annotating images for the object detection task is that all objects in all images need to be labeled. 
Suppose there are N inputs (images) which need to be annotated with the presence or absence of K labels (objects). A na\"ive approach would query humans for each combination of input and label, requiring $NK$ queries. However,  N and K can be very large and the cost of this exhaustive approach quickly becomes prohibitive. For example, annotating $60,000$ validation and test images with the presence or absence of $200$ object classes for the detection task na\"ively would take $80$ times more effort than
annotating $150,000$ validation and test images with $1$ object each for the classification task -- and this is not even counting the additional cost of collecting bounding box annotations around each object instance.  This quickly becomes infeasible. 

In~\citep{Deng14} we study strategies for scalable multilabel annotation, or for efficiently acquiring multiple labels from humans for a collection of items. 
We exploit three key observations for labels in real world applications (illustrated in Figure~\ref{fig:chipull}):

\begin{figure}
\flushright
\includegraphics[width=3.25in]{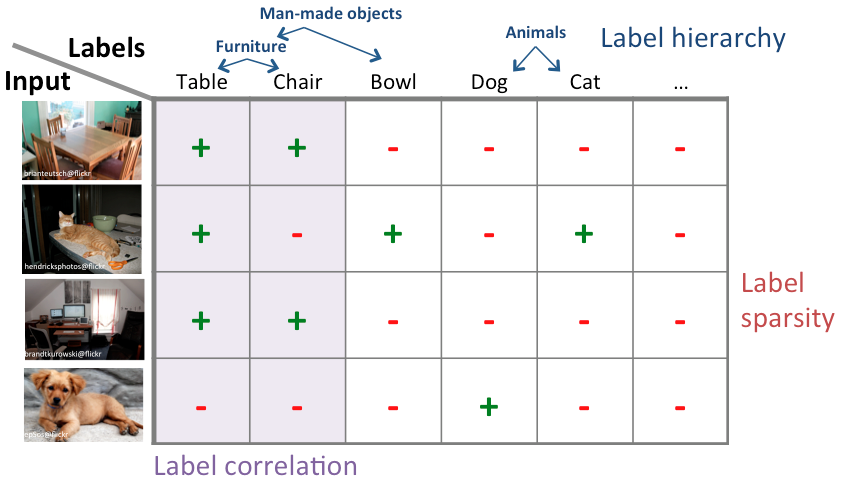}
\caption{Consider the problem of binary multi-label annotation. For each input (e.g., image) and each label (e.g., object), the goal is to determine the presence or absense (+ or -) of the label (e.g., decide if the object is present in the image). Multi-label annotation becomes much more efficient when considering real-world structure of data: correlation between labels, hierarchical organization of concepts, and sparsity of labels.}
\label{fig:chipull}
\end{figure}

\begin{enumerate}
\item {\bf Correlation.} Subsets of labels are often highly correlated. Objects such as a computer keyboard, mouse and monitor frequently co-occur  in images. Similarly, some labels tend to all be absent at the same time. For example, all objects that require electricity are usually absent in pictures taken outdoors. This suggests that we could potentially “fill in” the values of multiple labels by grouping them into only one query for humans. Instead of checking if dog, cat, rabbit etc. are present in the photo, we just check about the ``animal'' group If the answer is no, then this implies a no for all categories in the group.

\item {\bf Hierarchy.} The above example of grouping dog, cat, rabbit etc. into animal has implicitly assumed that labels can be
grouped together and humans can efficiently answer queries about the group as a whole. This brings up our second key observation: humans organize semantic concepts into hierarchies and are able to efficiently categorize at higher semantic levels~\citep{Thorpe96}, e.g. humans can determine the presence of an animal in an image as fast as every type of animal individually. This leads to substantial cost savings.

\item {\bf Sparsity.} The values of labels for each image tend to be sparse, i.e. an image is unlikely to contain more than a dozen types of objects, a small fraction of the hundreds of object categories. This enables rapid elimination of many objects by quickly filling in no. With a high degree of sparsity, an efficient algorithm can have a cost which grows logarithmically with the number of objects instead of linearly.
\end{enumerate}

We propose algorithmic strategies that exploit the above intuitions. The key is to select a sequence of queries for humans such that we achieve the same labeling results with only a fraction of the cost of the na\"ive approach. The main challenges include how to measure cost and utility of queries, how to construct good queries, and how to dynamically order them. A detailed description of the generic algorithm, along with theoretical analysis and empirical evaluation, is presented in~\citep{Deng14}.  

%Here we discuss the simplified version of this algorithm used for labeling images for ILSVRC detection task. 

\begin{figure}
\centering
\includegraphics[width=3.35in]{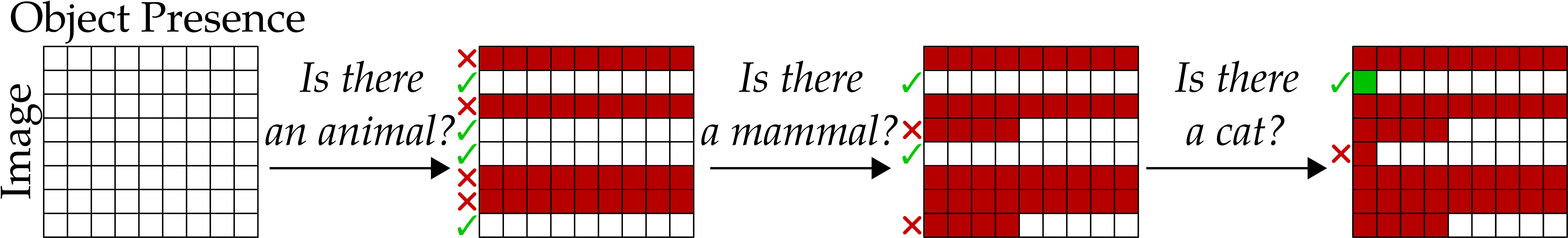}
\caption{Our algorithm dynamically selects the next query to efficiently determine the presence or absence of every object in every image. Green denotes a positive annotation and red denotes a negative annotation. This toy example illustrates a sample progression of the algorithm for one label (cat) on a set of images. }
\label{fig:chipipeline}
\end{figure}

\paragraph{Application of the generic multi-class labeling algorithm to our setting.}
The generic algorithm automatically selects the most informative queries to ask based on object label statistics learned from the training set.
In our case of 200 object classes, since obtaining the training set was by itself challenging we chose to design the queries by hand.
We created a hierarchy of queries of the type ``is there a... in the image?'' For example, one of the high-level questions was ``is there an animal in the image?'' We ask the crowd workers this question about every image we want to label. 
The children of the ``animal'' question would correspond to specific examples of animals: for example, ``is there a mammal in the image?'' or ``is there an animal with no legs?'' To annotate images efficiently, these questions are asked only on images determined to contain an animal.
The 200 leaf node questions correspond to the 200 target objects, e.g., ``is there a cat in the image?''. A few sample iterations of the algorithm are shown in Figure~\ref{fig:chipipeline}. 

Algorithm~\ref{algorithm:chi} is the formal algorithm for labeling an image with the presence or absence of each target object category.
With this algorithm in mind, the hierarchy of questions was constructed following the principle that false positives
 only add extra cost whereas false
negatives can significantly affect the quality of the labeling. Thus, it is always better to stick with 
more general but less ambiguous questions, such as ``is there a mammal in the image?'' as opposed
to asking
overly specific but potentially ambiguous questions, such as ``is there an animal that can climb trees?''
Constructing this hierarchy was a surprisingly time-consuming process, involving multiple iterations to ensure high accuracy of labeling and avoid question ambiguity. Appendix~\ref{app:hierarchy} shows the constructed hierarchy.

%Figure~\ref{fig:qhierarchy} shows the constructed hierarchy. 
%This was a 

\paragraph{Bounding box annotation.} 
Once all images are labeled with the presence or absence of all object categories we use the bounding box 
system described in Section~\ref{sec:AnnotLocBbox}
along with some additional modifications of Appendix~\ref{sec:AppDetBbox}
 to annotate the location of every instance of every present object category.

\begin{algorithm}[t]
 \SetAlgoLined
 \KwIn{Image $i$, queries $\mathcal{Q}$, directed graph $\mathcal{G}$ over $\mathcal{Q}$}
 \KwOut{Labels $L: \mathcal{Q} \rightarrow \{$``yes'', ``no''$\}$}
Initialize labels $L(q) = \emptyset \; \forall q \in \mathcal{Q}$\;
Initialize candidates $C = \{q \mbox{: } q \in Root(\mathcal{G})\}$\;
\While{$C$ not empty}{
Obtain answer $A$ to query $q* \in C$\;
$L(q*) = A$; $C  = C \backslash \{q*\}$\;
\uIf{A is ``yes''}{
$Chldr = \{q \in Children(q*,\mathcal{G}) \mbox{: } L(q) = \emptyset \}$\;
$C = C \cup Chldr$\;
}
\Else{
$Des = \{q \in Descendants(q*,\mathcal{G}) \mbox{: } L(q) = \emptyset \}$\;
$L(q) =``no'' \; \forall q \in Des$\;
$C = C \backslash Des$\;
}
 }

 \caption{The algorithm for complete multi-class annotation. This is a special case of the algorithm described in~\citep{Deng14}.
A hierarchy of questions  $\mathcal{G}$ is manually constructed.  All root questions are asked on every image. If the answer to query $q*$ on image $i$ is ``no'' then the answer is assumed to be ``no''  for all queries $q$ such that $q$ is a descendant of $q*$ in the hierarchy. We continue asking the queries until all queries are answered. For images
taken from the single-object localization task we used the known object label to initialize $L$.
}
 \label{algorithm:chi}
\end{algorithm}

% described in Appendix~\ref{sec:AppDetLocExt}. 

%\subsubsection{Bounding box annotation}
%\label{sec:DetBbox}

\subsubsection{Object detection dataset statistics}
\label{sec:AnnotDetStats}

Using the procedure described above, we collect a large-scale dataset
for ILSVRC object detection task. 
 There are 200 object classes and approximately 450K training images, 20K validation images and 40K test images. Table~\ref{table:detstats} documents the size of the dataset over the years of the challenge. The major change between ILSVRC2013 and ILSVRC2014 was the addition of 60,658 
fully annotated training images.

Prior to ILSVRC, the object detection benchmark was the PASCAL VOC challenge~\citep{PASCALIJCV}. 
ILSVRC has $10$ times more object classes than PASCAL VOC (200 vs 20), $10.6$ times more fully annotated training images (60,658 vs 5,717), $35.2$ times more training objects (478,807 vs 13,609),
$3.5$ times more validation images (20,121 vs 5823) and $3.5$ times more validation objects (55,501 vs 
15,787). ILSVRC has $2.8$ annotated objects per image on the validation set, compared to $2.7$ in PASCAL VOC. The average object in ILSVRC takes up $17.0\%$ of the image area and in PASCAL VOC takes up $20.7\%$; Table~\ref{table:detclassespascal} contains per-class comparisons. Additionally, ILSVRC contains a wide variety of objects, including tiny objects such as sunglasses ($1.3\%$ of image area on average), ping-pong balls ($1.5\%$ of image area on average) and basketballs ($2.0\%$ of image area on average). 

%\footnote{In the average object category, the average object instance occupies $16.0\%$ of the image area in the ILSVRC detection dataset, $24.1\%$ of the image area in PASCAL VOC 2012, and $35.8\%$ of the image area in ILSVRC2012 localization validation dataset.}

\begin{table*}

\begin{center} {\bf Object detection annotations (200 object classes)} \end{center}

\begin{tabular}{|p{0.7in}|c|c|c|c|c|}
\hline
Year
 &
\parbox[t]{1.3in}{\centering {Train images   \\ (per class)}}
&
\parbox[t]{1.3in}{\centering {Train bboxes annotated \\ (per class)}}
&
\parbox[t]{0.7in}{\centering {Val images  \\ (per class) }}
&
\parbox[t]{1.2in}{\centering {Val bboxes annotated \\(per class )}}
&
\parbox[t]{0.4in}{\centering {Test \\ images  }}
\\
\hline
ILSVRC2013
& \parbox[t]{1.3in}{\centering 395909 \\(417-561-66911 pos, \\185-4130-10073 neg)}
& \parbox[t]{1in}{\centering 345854 \\ (438-660-73799)}
& \parbox[t]{0.9in}{\centering 21121 \\ (23-58-5791 pos, \\ rest neg)}
& \parbox[t]{1in}{\centering 55501 \\ (31-111-12824)}
& 40152
\\
\hline
ILSVRC2014 
& \parbox[t]{1.3in}{\centering 456567 \\(461-823-67513 pos, \\42945-64614-70626 neg)}
& \parbox[t]{1in}{\centering 478807 \\ (502-1008-74517)}
& \parbox[t]{0.9in}{\centering 21121 \\ (23-58-5791 pos, \\ rest neg)}
& \parbox[t]{1in}{\centering 55501 \\ (31-111-12824)}
& 40152
\\
\hline
\end{tabular}

\caption{Scale of ILSVRC object detection task. Numbers in parentheses correspond to (minimum per class - median per class - maximum per class).}
\label{table:detstats}
\end{table*}

%%%%%%%%%%%%%%%%%%%%%%%%%%%%%%%%%%%%%%%%%%%%%%%%%%%%%%%%%%%%%%%%%%%%%%%%%%%%%%%%%%%%%%%%%%
%%%%%%%%%%%%%%%%%%%%%%%%%%%%%%%%%%%%%%%%%%%%%%%%%%%%%%%%%%%%%%%%%%%%%%%%%%%%%%%%%%%%%%%%%%
%%%%%%%%%%%                                           %%%%%%%%%%%%%%%%%%%%%%%%%%%%%%%%%%%%%%%%%%%%%%%%%%%%%%%%%%%
%%%%%%%%%%%        EVALUATION               %%%%%%%%%%%%%%%%%%%%%%%%%%%%%%%%%%%%%%%%%%%%%%%%%%%%%%%%%%%
%%%%%%%%%%%                                           %%%%%%%%%%%%%%%%%%%%%%%%%%%%%%%%%%%%%%%%%%%%%%%%%%%%%%%%%%%
%%%%%%%%%%%%%%%%%%%%%%%%%%%%%%%%%%%%%%%%%%%%%%%%%%%%%%%%%%%%%%%%%%%%%%%%%%%%%%%%%%%%%%%%%%
%%%%%%%%%%%%%%%%%%%%%%%%%%%%%%%%%%%%%%%%%%%%%%%%%%%%%%%%%%%%%%%%%%%%%%%%%%%%%%%%%%%%%%%%%%

\section{Evaluation at large scale}
\label{sec:Evaluation}

Once the dataset has been collected, we need to define a standardized evaluation procedure for algorithms. 
Some measures have already been established by datasets such as the Caltech 101~\citep{Caltech101} for image
classification and PASCAL VOC~\citep{PASCALVOC} for both image classification and object detection. 
To adapt these procedures to the large-scale setting we had to address three key challenges.
First, for the image classification and single-object localization tasks only one object category could be labeled in each image due to the scale of the dataset. This created potential ambiguity during evaluation (addressed in Section~\ref{sec:EvaluationCls}). 
Second, evaluating localization of object instances is inherently difficult in some images which contain a cluster of objects
(addressed in Section~\ref{sec:EvaluationLoc}). Third, evaluating localization of object instances which occupy  few pixels in the image is challenging (addressed in Section~\ref{sec:EvaluationDet}).

%. To address this we relaxed the object detection evaluation criteria pioneered by the PASCAL VOC (Section~\ref{sec:EvaluationDet}). 

In this section we describe the standardized evaluation criteria for each of the three ILSVRC tasks. We elaborate further  on these and other more minor challenges with large-scale evaluation. Appendix~\ref{sec:AppCompetition} describes the submission protocol and other details of running the competition itself.

%The second difference is in how algorithms are evaluated in inherently ambiguous cases. We adapt  the same approach as in the single-object localization %setting: we label entire (image, class) pairs as ``difficult'' and ignore them during evaluation.

% and manually mark individual instances (as opposed to entire object classes) as "difficult." 

\begin{figure*}
\includegraphics[width=\linewidth]{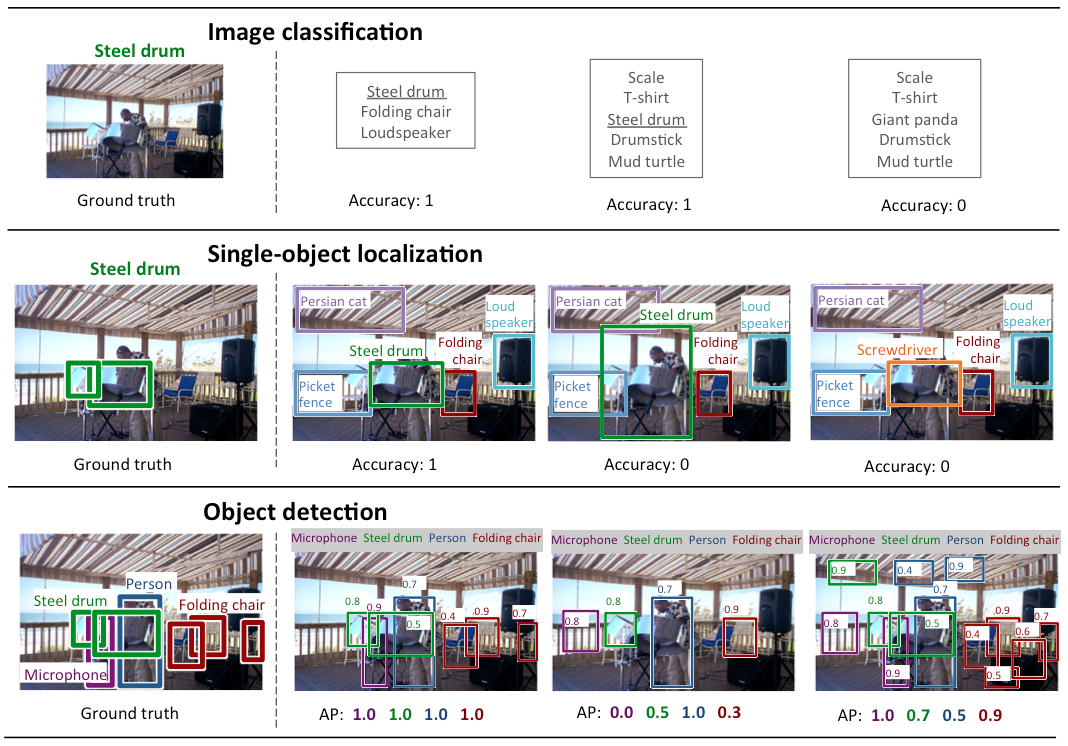}
\caption{Tasks in ILSVRC. The first column shows the ground truth labeling on an example image, and the next three show three sample outputs with the corresponding
evaluation score.}
\label{fig:tasks}
\end{figure*}

\subsection{Image classification}
\label{sec:EvaluationCls}

The scale of ILSVRC classification task (1000 categories and more than a million of images) makes it very expensive to label every instance of every object in every image.  Therefore, on this dataset only one object category is labeled in each image. This creates ambiguity in evaluation. For example, an image might be labeled as a ``strawberry'' but contain both a strawberry and an apple. Then an algorithm would not know which one of the two objects to name. For the image classification task we allowed an algorithm to identify multiple (up to 5) objects in an image and not be penalized as long as one of the objects indeed corresponded to the ground truth label. Figure~\ref{fig:tasks}(top row) shows some examples.

Concretely, each image $i$ has a single class label $C_i$. An algorithm is allowed to return 5 labels $c_{i1},\dots c_{i5}$, and is considered correct if $c_{ij} = C_i$ for some $j$. 
%The idea is to allow an algorithm to identify multiple objects in an image and not be penalized if one of the objects identified was in fact present, but %not included in the ground truth. 
%Figure~\ref{fig:tasks}(top) shows some examples.

Let the error of a prediction $d_{ij} = d(c_{ij},C_i)$ be $1$ if $c_{ij} \neq C_i$ and $0$ otherwise. The error of an algorithm
is the fraction of test images on which the algorithm makes a mistake:
\begin{align}
\label{eq:clserror}
\mbox{error } &= \frac{1}{N} \sum_{i=1}^N \min_j d_{ij} %(c_{ij}, C_i)
\end{align}

We used two additional measures of error. First, we evaluated top-1 error. In this case algorithms were penalized if their highest-confidence output label $c_{i1}$ did not match ground truth class $C_i$. Second, we evaluated hierarchical error. The intuition is that confusing two nearby classes (such as two different breeds of dogs) is not as harmful as confusing a dog for a container ship. For the hierarchical criteria, the cost of one misclassification, $d(c_{ij},C_i)$, is defined as the height of the lowest common ancestor of $c_{ij}$
and $C_i$ in the ImageNet hierarchy.  The height of a node is the
length of the longest path to a leaf node (leaf nodes have height
zero). 

However, in practice we found that all three measures of error (top-5, top-1, and hierarchical) produced the same ordering of results. Thus, since ILSVRC2012 we have been exclusively using the top-5 metric which is the simplest and most suitable to the dataset.

\subsection{Single-object localization}
\label{sec:EvaluationLoc}

The evaluation for single-object localization is similar to object classification, again using a top-5 criteria to allow the algorithm to return unannotated object classes without penalty. However, now the algorithm is considered correct only if it both correctly identifies the target class $C_i$ and accurately localizes one of its instances. Figure~\ref{fig:tasks}(middle row) shows some examples.

Concretely, an image is associated with object class $C_i$, with all instances of this object class annotated with bounding boxes $B_{ik}$. An algorithm returns $\{(c_{ij},b_{ij})\}_{j=1}^5$ of class labels $c_{ij}$ and associated locations $b_{ij}$. The error of a prediction $j$ is:
\begin{align}
%d(c_{ij},b_{ij},C_i,\{B_{ik}\}) = 
d_{ij} &= \max(d(c_{ij},C_i),\min_{k}d(b_{ij},B_{ik}))
\end{align}
Here $d(b_{ij},B_{ik})$ is the error of localization, defined as $0$ if the area of intersection of boxes $b_{ij}$ and $B_{ik}$ divided by the areas of their union is greater than $0.5$, and $1$ otherwise.~\citep{PASCALIJCV} The error of an algorithm is computed as in  Eq.~\ref{eq:clserror}.

\begin{figure*}
\includegraphics[width=\linewidth]{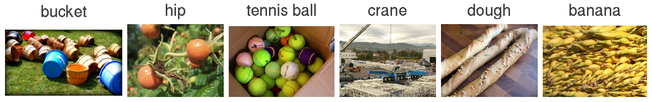}
\caption{Images marked as ``difficult'' in the ILSVRC2012 single-object localization validation set. Please refer to Section~\ref{sec:EvaluationLoc} for details.}
\label{fig:diff}
\end{figure*}

Evaluating localization is inherently difficult in some images. Consider a picture of a bunch of bananas or a carton of apples. It is easy to classify these images as containing bananas or apples, and even possible to localize a few instances of each fruit. However, in order for evaluation to be accurate \emph{every} instance of banana or apple needs to be annotated, and that may be impossible.
To handle the images where localizing individual object instances is inherently ambiguous we manually discarded $3.5\%$ of images  since ILSVRC2012.  Some examples of discarded images are shown in Figure~\ref{fig:diff}.

% In PASCAL VOC individual object instances which are considered too challenging to detect are manually
%marked as ``difficult'' and excluded from evaluation. Due to the scale of ILSVRC obtaining these (subjective) labels in a crowdsourced %setting quickly becomes infeasible. 

%During data collection we crowdsourced annotation of up to 10 instances of each object class on each image, and then manually verified all images with 10 instances to determine which ones are ``difficult''. This resulted in  images being discarded from the single-object localization validation and test set. 

\subsection{Object detection}
\label{sec:EvaluationDet}

The criteria for object detection was adopted from PASCAL VOC~\citep{PASCALIJCV}. It is designed to penalize the algorithm for
missing object instances, for duplicate detections of one instance, and for false positive detections. Figure~\ref{fig:tasks}(bottom row) shows examples.

For each object class and each image $I_i$, an algorithm returns predicted detections $(b_{ij},s_{ij})$ of predicted locations $b_{ij}$ with confidence scores $s_{ij}$. These detections are greedily matched to the ground truth boxes $\{B_{ik}\}$ using Algorithm~\ref{algo:deteval}. For every detection $j$ on image $i$ the algorithm returns $z_{ij} = 1$ if the detection is matched to a ground truth box according to the threshold criteria, and $0$ otherwise. For a given object class, let $N$ be the total number of ground truth instances across all images. Given a threshold $t$, 
define \emph{ recall} as the fraction of the $N$ objects  detected by the algorithm, and \emph{precision} as
the fraction of correct detections out of the total detections returned by the algorithm. Concretely,
\begin{align}
Recall(t) &= \frac{\sum_{ij}  1[s_{ij} \geq t] z_{ij} }{N} \\
Precision(t) &= \frac{\sum_{ij}  1[s_{ij} \geq t] z_{ij} }{\sum_{ij}  1[s_{ij} \geq t]} 
\end{align}

\begin{algorithm}[t]
 \SetAlgoLined
 \KwIn{Bounding box predictions with confidence scores $\{(b_j,s_j)\}_{j=1}^M$ and ground truth boxes $\mathcal{B}$ on image $I$ for a given object class.}
 \KwOut{Binary results $\{z_j\}_{j=1}^M$ of whether or not prediction $j$ is a true positive detection}
Let $\mathcal{U} = \mathcal{B}$ be the set of unmatched objects\;
%Let $\mbox{thr}(B_k) = \min(0.5, \frac{w_k h_k}{(w_k+10)(h_k+10)}$\;
Order $\{(b_j,s_j)\}_{j=1}^M$ in descending order of $s_j$\;
\For{j=1 \dots M}{
	Let $\mathcal{C} = \{B_k \in \mathcal{U} \; : \; \mbox{IOU}(B_k,b_j) \geq \mbox{thr}(B_k)\}$\;
	\uIf{$\mathcal{C} \neq \emptyset$}{
		Let $k^* = \argmax_{\{k \; : \; B_k \in \mathcal{C}\}}  \mbox{IOU}(B_k,b_j)$\;
		Set $\mathcal{U} = \mathcal{U} \backslash B_{k*}$\;
Set $z_j = 1$ since true positive detection\;	
	} \Else {
	Set $z_j = 0$ since false positive detection\;
}
}
 \caption{The algorithm for greedily matching object detection outputs to ground truth labels. The standard $\mbox{thr}(B_k) = 0.5$~\citep{PASCALIJCV}. ILSVRC computes  $\mbox{thr}(B_k)$  using Eq.~\ref{eq:thr} to better handle low-resolution objects.
}
 \label{algo:deteval} 
\end{algorithm}

The final metric for evaluating an algorithm on a given object class is \emph{average precision}
over the different levels of recall achieved by varying the threshold $t$. The winner of each object class 
is then the team with the highest average precision, and then winner of the challenge is the team that wins on the most
object classes.\footnote{In this paper we focus on the mean average precision across all categories as the measure of a team's performance. This is done for simplicity and is justified since the ordering of teams by mean average precision was always the same as the ordering by object categories won.}

\paragraph{Difference with PASCAL VOC.} 
Evaluating localization of object instances which occupy very few pixels in the image is challenging. The PASCAL VOC approach was to label such instances as ``difficult'' and ignore them during evaluation. However, since ILSVRC contains a more diverse set of object classes including, for example, ``nail'' and ``ping pong ball'' which have many very small instances, it is important to include even very small object instances in evaluation. 

%For the object detection evaluation, where every instance of every object had to be detected, we relaxed the criteria for a detection to be considered ``correct'' on ground truth object instances which occupy very few pixels

%To better evaluate how well small object instances are localized we had to relax the object detection evaluation criteria pioneered by %the PASCAL VOC. 

In Algorithm~\ref{algo:deteval}, a predicted bounding box $b$ is considered to have properly localized by a
ground truth bounding box $B$ if $IOU(b,B) \geq \mbox{thr}(B)$. The PASCAL VOC metric uses the threshold $\mbox{thr}(B) = 0.5$.
However,  for small objects even deviations of a few pixels would be unacceptable
  according to this threshold. For example, consider an object $B$ of size
  $10 \times 10$ pixels, with a detection window of $20 \times 20$ pixels which fully
  contains that object.  This would be an error of approximately $5$ pixels on each
  dimension, which is average human annotation error. However, the IOU
  in this case would be $100/400 = 0.25$, far below the threshold of $0.5$. Thus for
  smaller objects we loosen the threshold in ILSVRC to allow for the annotation
  to extend up to 5 pixels on average in each direction around the
object. Concretely, if the ground truth box $B$ is of dimensions $w \times h$ then
\begin{equation}
\label{eq:thr}
\mbox{thr}(B) = \min \left (0.5, \frac{w h}{(w+10)(h+10)} \right )
\end{equation}
 In practice, this changes the threshold only on objects which are
  smaller than approximately $25\times 25$ pixels, and affects $5.5\%$ of objects in the
 detection validation set.

\paragraph{Practical consideration.}
One additional practical consideration for ILSVRC detection evaluation is subtle and comes directly as a result of the 
scale of ILSVRC. In PASCAL, algorithms would often return many detections per class 
on the test set, including ones with low confidence scores. This allowed the algorithms to reach the level of 
high recall at least in the realm of very low precision. On ILSVRC detection test set if an algorithm returns 10 bounding boxes per object
per image this would result in  $10 \times 200 \times 40K = 80$M detections. Each detection contains an image index, a class index, 4 bounding box coordinates, and the confidence score, so it takes on the order of 28 bytes. The full set of detections would then require $2.24$Gb to store and submit to the evaluation server, which is impractical. This means that  algorithms are implicitly required to limit their predictions to only the most confident locations.

%%%%%%%%%%%%%%%%%%%%%%%%%%%%%%%%%%%%%%%%%%%%%%%%%%%%%%%%%%%%%%%%%%%%%%%%%%%%%%%%%%%%%%%%%%
%%%%%%%%%%%%%%%%%%%%%%%%%%%%%%%%%%%%%%%%%%%%%%%%%%%%%%%%%%%%%%%%%%%%%%%%%%%%%%%%%%%%%%%%%%
%%%%%%%%%%%                                           %%%%%%%%%%%%%%%%%%%%%%%%%%%%%%%%%%%%%%%%%%%%%%%%%%%%%%%%%%%
%%%%%%%%%%%        METHODS                    %%%%%%%%%%%%%%%%%%%%%%%%%%%%%%%%%%%%%%%%%%%%%%%%%%%%%%%%%%%
%%%%%%%%%%%                                           %%%%%%%%%%%%%%%%%%%%%%%%%%%%%%%%%%%%%%%%%%%%%%%%%%%%%%%%%%%
%%%%%%%%%%%%%%%%%%%%%%%%%%%%%%%%%%%%%%%%%%%%%%%%%%%%%%%%%%%%%%%%%%%%%%%%%%%%%%%%%%%%%%%%%%
%%%%%%%%%%%%%%%%%%%%%%%%%%%%%%%%%%%%%%%%%%%%%%%%%%%%%%%%%%%%%%%%%%%%%%%%%%%%%%%%%%%%%%%%%%

\section{Methods}
\label{sec:Methods}

The ILSVRC dataset and the competition has allowed significant algorithmic advances in large-scale image recognition and retrieval.

%%%%%%%%%%%%%%%%%%%%%%%%%%%%%%%%%%%%%%%%%%%%%%%%%%%%%%%%%%%%%%%%%%%%%%%%%%%%%%%%%%%%%%%%%%
%%%%%%%%%%%                                                            %%%%%%%%%%%%%%%%%%%%%%%%%%%%%%%%%%%%%%%%%%%%%%%%%%%
%%%%%%%%%%%        Challenge participation                 %%%%%%%%%%%%%%%%%%%%%%%%%%%%%%%%%%%%%%%%%%%%%%%%%%%
%%%%%%%%%%%                                          	             %%%%%%%%%%%%%%%%%%%%%%%%%%%%%%%%%%%%%%%%%%%%%%%%%%%
%%%%%%%%%%%%%%%%%%%%%%%%%%%%%%%%%%%%%%%%%%%%%%%%%%%%%%%%%%%%%%%%%%%%%%%%%%%%%%%%%%%%%%%%%%

\subsection{Challenge entries}
\label{sec:MethodsEntries}

This section is organized chronologically, highlighting the particularly innovative and successful methods which participated in the ILSVRC each year. Tables~\ref{table:sub10-12}, \ref{table:sub13} and \ref{table:sub14} list all the participating teams. We see a turning point in 2012
with the development of large-scale convolutional neural networks.

\paragraph{ILSVRC2010.} The first year the challenge consisted of just the classification task. The winning entry from NEC team~\citep{Lin11} used SIFT~\citep{Lowe04} and LBP~\citep{LBP} features with two non-linear coding representations~\citep{Zhou10,LLC} and a stochastic SVM. The honorable mention XRCE team~\citep{Perronnin10} used an improved Fisher vector representation~\citep{Perronnin07} along with PCA dimensionality reduction and data compression followed by a linear SVM. Fisher vector-based methods have evolved over five years of the challenge and continued performing strongly in every ILSVRC from 2010 to 2014.

\paragraph{ILSVRC2011.} The winning classification entry in 2011 was the 2010 runner-up team XRCE, applying high-dimensional image signatures~\citep{Perronnin10} with compression using product quantization~\citep{Sanchez11} and one-vs-all linear SVMs. The single-object localization competition was held for the first time, with two brave entries. The winner was the UvA team using a selective search approach to generate class-independent object hypothesis regions~\citep{vanDeSande11}, followed by dense sampling and vector quantization of several color SIFT features~\citep{vanDeSande10}, pooling with spatial pyramid matching~\citep{Lazebnik06}, and classifying with a histogram intersection kernel SVM~\citep{Maji09} trained on a GPU~\citep{vanDeSande11b}.

\paragraph{ILSVRC2012.} This was a turning point for large-scale object recognition, when large-scale deep neural networks entered the scene. The undisputed winner of both the classification and localization tasks in 2012 was the SuperVision team. They trained a large, deep convolutional neural network on RGB values, with 60 million parameters using an efficient GPU implementation and a novel hidden-unit dropout trick~\citep{Krizhevsky12,Hinton12}. The second place in image classification went to the ISI team, which used Fisher vectors~\citep{Sanchez11} and a streamlined version of Graphical Gaussian Vectors~\citep{Harada12}, along with linear classifiers using Passive-Aggressive (PA) algorithm~\citep{Crammer06}.  The second place in single-object localization went to the VGG, with an image classification system including dense SIFT features and color statistics~\citep{Lowe04}, a Fisher vector representation~\citep{Sanchez11}, and a linear SVM classifier, plus additional insights from~\citep{Arandjelovic12,Sanchez12}. Both ISI and VGG used \citep{Felzenszwalb10} for object localization; SuperVision used a regression model trained to predict bounding box locations. Despite the weaker detection model, SuperVision handily won the object localization task. 
A detailed analysis and comparison of the SuperVision and VGG submissions on the single-object localization task can be found in~\citep{Russakovsky13}.
The influence of the success of the SuperVision model can be clearly seen in ILSVRC2013 and ILSVRC2014.

\paragraph{ILSVRC2013.} There were 24 teams participating in the ILSVRC2013 competition, compared to 21 in the previous three years \emph{combined}. Following the success of the deep learning-based method in 2012, the vast majority of entries in 2013 used deep convolutional neural networks in their submission. The winner of the classification task was Clarifai, with several large deep convolutional networks averaged together. The network architectures were chosen using the visualization technique of~\citep{Zeiler13}, and they were trained on the GPU following~\citep{Zeiler11} using the dropout technique~\citep{Krizhevsky12}.

The winning single-object localization OverFeat submission was based on an integrated framework for using convolutional networks for classification, localization and detection with a multiscale sliding window approach~\citep{Sermanet13}. They were the only team tackling all three tasks. 

The winner of object detection task was UvA team, which utilized a new way of efficient encoding~\citep{vanDeSande14} densely sampled color descriptors~\citep{vanDeSande10} pooled using a multi-level spatial pyramid in a selective search framework \citep{Uijlings13}. The detection results were rescored using a full-image convolutional network classifier.

\paragraph{ILSVRC2014.} 2014 attracted the most submissions, with 36 teams submitting 123 entries compared to just 24 teams in 2013 -- a 1.5x increase in participation.\footnote{Table~\ref{table:sub14} omits 4 teams which submitted results but chose not to officially participate in the challenge.} As in 2013 almost all teams used convolutional neural networks as the basis for their submission. Significant progress has been made in just one year: image classification error was almost halved since ILSVRC2013 and object detection mean average precision almost doubled compared to ILSVRC2013. Please refer to Section~\ref{sec:ResultsYears} for details.

In 2014 teams were allowed to use outside data for training their models in the competition, so there were six tracks: provided and outside data tracks in each of image classification, single-object localization, and object detection tasks.  

\afterpage{
\begin{landscape}
\clearpage

\begin{table}
{%\footnotesize

\begin{center} {\bf ILSVRC 2010} \end{center}
\begin{tabular}{p{\nv}cc>{\crf}p{\iv}>{\crf}p{\crv}}
\hline
{\bf Codename} & CLS & {\color{white} LOC } & {\bf Insitutions} & {\bf Contributors and references} \\
\hline
Hminmax & 54.4 && Massachusetts Institute of Technology & Jim Mutch, Sharat Chikkerur, Hristo Paskov, Ruslan Salakhutdinov, Stan Bileschi, Hueihan Jhuang \\
\hline
IBM & 70.1 && IBM research\oi, Georgia Tech\oii & Lexing Xie\oi, Hua Ouyang\oii, Apostol Natsev\oi \\
\hline
ISIL & 44.6 && Intelligent Systems and Informatics Lab., The University of Tokyo & Tatsuya Harada, Hideki Nakayama, Yoshitaka Ushiku, Yuya Yamashita, Jun Imura, Yasuo Kuniyoshi  \\
\hline
ITNLP & 78.7 && Harbin Institute of Technology &Deyuan Zhang, Wenfeng Xuan, Xiaolong Wang, Bingquan Liu, Chengjie Sun \\
\hline
LIG & 60.7 && Laboratoire d'Informatique de Grenoble & Georges Qu\'enot  \\
\hline
NEC & 
{\bf 28.2} && 
NEC Labs America\oi, University of Illinois at Urbana-Champaign\oii, Rutgers\oiii &
\parbox[t]{\crv}{ Yuanqing Lin\oi, Fengjun Lv\oi, Shenghuo Zhu\oi, Ming Yang\oi, Timothee Cour\oi, Kai Yu\oi, LiangLiang Cao\oii, Zhen Li\oii, Min-Hsuan Tsai\oii, Xi Zhou\oii, Thomas Huang\oii, Tong Zhang\oiii \\ \citep{Lin11}}
 \\
\hline
NII & 74.2	&& National Institute of Informatics, Tokyo,Japan\oi, Hefei Normal Univ. Heifei, China\oii & Cai-Zhi Zhu\oi, Xiao Zhou\oii, Shin\'ichi Satoh\oi \\
\hline
NTU & 58.3 && CeMNet, SCE, NTU, Singapore & Zhengxiang Wang,  Liang-Tien Chia \\
\hline
Regularities & 75.1 && SRI International & Omid Madani, Brian Burns \\
\hline
UCI & 46.6 &&University of California Irvine & Hamed Pirsiavash, Deva Ramanan, Charless Fowlkes \\
\hline
XRCE
& 33.6 && Xerox Research Centre Europe & \parbox[t]{\crv}{Jorge Sanchez, Florent Perronnin, Thomas Mensink \\ \citep{Perronnin10} }\\
\hline
\end{tabular}

\begin{center} {\bf ILSVRC 2011} \end{center}

\begin{tabular}{p{\nv}cc>{\crf}p{\iv}>{\crf}p{\crv}}
\hline
{\bf Codename} & CLS & LOC & {\bf Institutions} & {\bf Contributors and references} \\
\hline
ISI & 36.0 & - & Intelligent Systems and Informatics lab, University of Tokyo & Tatsuya Harada,
Asako Kanezaki, Yoshitaka Ushiku, Yuya Yamashita, 
Sho Inaba, 
Hiroshi Muraoka, 
Yasuo Kuniyoshi \\
\hline
 NII & 50.5 & - & National Institute of Informatics, Japan	 & Duy-Dinh Le, Shin\'ichi Satoh \\
\hline
UvA
 & 31.0 & {\bf 42.5} & University of Amsterdam\oi, University of Trento\oii & 
 \parbox[t]{\crv}{Koen E. A. van de Sande\oi, Jasper R. R. Uijlings\oii, 
Arnold W. M. Smeulders\oi,
Theo Gevers\oi, 
Nicu Sebe\oii,
Cees Snoek\oi \\ \citep{vanDeSande11} } \\
\hline
XRCE 
 & {\bf 25.8} & 56.5 & Xerox Research Centre Europe\oi, CIII\oii & \parbox[t]{\crv}{Florent Perronnin\oi, Jorge Sanchez\oi\oii \\ \citep{Sanchez11} } \\
\hline
\end{tabular}

\begin{center} {\bf ILSVRC 2012} \end{center}

\begin{tabular}{p{\nv}cc>{\crf}p{\iv}>{\crf}p{\crv}}
\hline
{\bf Codename} & CLS & LOC & {\bf Institutions} & {\bf Contributors and references} \\
\hline
ISI & 26.2 &  53.6 &  University of Tokyo\oi, JST PRESTO\oii & \parbox[t]{\crv}{Naoyuki Gunji\oi, Takayuki Higuchi\oi, Koki Yasumoto\oi, Hiroshi Muraoka\oi, Yoshitaka Ushiku\oi, Tatsuya Harada\oi\oii, Yasuo Kuniyoshi\oi  \\ \citep{Harada12} }
\\
\hline
LEAR & 34.5 & - &  LEAR  INRIA Grenoble\oi, TVPA Xerox Research Centre Europe\oii & 
\parbox[t]{\crv}{Thomas Mensink\oi\oii, 
Jakob Verbeek\oi, Florent Perronnin\oii, Gabriela Csurka\oii \\ \citep{Mensink12} } \\
\hline
VGG & 27.0 & 50.0 & University of Oxford & \parbox[t]{\crv}{Karen Simonyan, Yusuf Aytar, Andrea Vedaldi, Andrew Zisserman \\
\citep{Arandjelovic12,Sanchez12} }
\\
\hline
SuperVision & {\bf 16.4} & {\bf 34.2} & University of Toronto & \parbox[t]{\crv}{Alex Krizhevsky, Ilya Sutskever, Geoffrey Hinton \\ \citep{Krizhevsky12} }\\
\hline
UvA  & 29.6 & - & University of Amsterdam  &  \parbox[t]{\crv}{Koen E. A. van de Sande,  Amir Habibian, Cees G. M. Snoek \\ 
\citep{Sanchez11,Scheirer12} }
\\
\hline
XRCE & 27.1 & - & Xerox Research Centre Europe\oi, LEAR INRIA \oii & \parbox[t]{\crv}{Florent Perronnin\oi,
Zeynep Akata\oi\oii, 
Zaid Harchaoui\oii,
Cordelia Schmid\oii \\ \citep{Perronnin12} } \\
\hline

\end{tabular}

}
\caption{Teams participating in ILSVRC2010-2012, ordered alphabetically. Each method is identified with a codename used in the text. 
We report flat top-5 classification and single-object localization error, in percents (lower is better).  For teams which submitted multiple entries we report the best score. In 2012, SuperVision also submitted entries trained with the extra data from the ImageNet Fall 2011 release, and obtained $15.3\%$ classification error and $33.5\%$ localization error.
Key references are provided where available. More details about the winning entries can be found in Section~\ref{sec:MethodsEntries}. }
\label{table:sub10-12}
\end{table}

\end{landscape}
}

\afterpage{
\begin{landscape}
\clearpage
\begin{table}

\begin{center}{\centering \bf ILSVRC 2013}\end{center}

\begin{tabular}{p{\nw}ccc>{\crf}p{\ciw}>{\crf}p{\crw}}
\hline
{\bf Codename} & CLS & LOC & DET & { \bf Insitutions} & { \bf Contributors and references} \\
\hline
% &  $\bullet$ &&&  & \parbox[t]{3in}{  \\ \citep{}} \\
% &  $\bullet$ &&&  & \\

Adobe &  15.2 &-&-& Adobe\oi, University of Illinois at Urbana-Champaign\oii & \parbox[t]{\crw}{Hailin Jin\oi, Zhe Lin\oi,  Jianchao Yang\oi, Tom Paine\oii \\ \citep{Krizhevsky12}}  \\
\hline

AHoward &  13.6 &- & - & Andrew Howard Consulting & Andrew Howard\\
\hline

BUPT &  25.2 & - & - &  Beijing University of Posts and Telecommunications\oi, Orange Labs International Center Beijing\oii & Chong Huang\oi, 
Yunlong Bian\oi, Hongliang Bai\oii, 
Bo Liu\oi, Yanchao Feng\oi, Yuan Dong\oi \\
\hline

Clarifai &  {\bf 11.7 } & - & - & Clarifai & \parbox[t]{\crw}{ Matthew Zeiler \\ \citep{Zeiler13,Zeiler11}}  \\
\hline

\parbox[t]{\nwe}{CogVision} &  16.1 & - & - &  Microsoft Research\oi,
Harbin Institute of Technology\oii  & Kuiyuan Yang\oi, Yalong Bai\oi, Yong Rui\oii \\
\hline

decaf &  19.2 & - & - & University of California Berkeley & \parbox[t]{\crw}{ Yangqing Jia, Jeff Donahue, Trevor Darrell \\  \citep{Donahue13}}  \\
\hline

Deep Punx &  20.9 & - & - &  Saint Petersburg State University  & \parbox[t]{\crw}{ Evgeny Smirnov, Denis Timoshenko, Alexey Korolev \\ \citep{Krizhevsky12,Wan13,Tang13}}
 \\
\hline

Delta &  - & - & 6.1 & National Tsing Hua University & Che-Rung Lee, Hwann-Tzong Chen, Hao-Ping Kang,  Tzu-Wei Huang,  Ci-Hong Deng, Hao-Che Kao \\
\hline

\parbox[t]{\nwe}{IBM } &  20.7 & - & - & University of Illinois at Urbana-Champaign\oi, IBM Watson Research Center\oii, IBM Haifa Research Center\oiii & Zhicheng Yan\oi,  Liangliang Cao\oii,  John R Smith\oii, Noel Codella\oii,Michele Merler\oii, Sharath Pankanti\oii, Sharon Alpert\oiii, Yochay Tzur\oiii, \\
\hline

MIL &  24.4 & - & - &  University of Tokyo & Masatoshi Hidaka, Chie Kamada, Yusuke Mukuta, Naoyuki Gunji, Yoshitaka Ushiku, Tatsuya Harada  \\
\hline

Minerva & 21.7 &&& Peking University\oi, Microsoft Research\oii, Shanghai Jiao Tong University\oiii, XiDian University\oiv, Harbin Institute of Technology\ov & 
\parbox[t]{\crw}{ 
Tianjun Xiao\oi\oii, Minjie Wang\oiii\oii, Jianpeng Li\oiv\oii,   Yalong Bai\ov\oii,
Jiaxing Zhang\oii, Kuiyuan Yang\oii,  Chuntao Hong\oii, Zheng Zhang\oii
\\
\citep{Wang14} }
%Srivastava13,Goodfellow13,Krizhevsky12
\\
\hline

NEC & - & - & 19.6 & NEC Labs America\oi, University of Missouri \oii & \parbox[t]{\crw}{Xiaoyu Wang\oi, Miao Sun\oii, 
Tianbao Yang\oi, Yuanqing Lin\oi, Tony X. Han\oii, Shenghuo Zhu\oi \\
\citep{Wang13} }
\\
\hline

NUS &  13.0 &&& National University of Singapore &  \parbox[t]{\crw}{Min Lin*, Qiang Chen*, Jian Dong, Junshi Huang, Wei Xia, Shuicheng Yan (* = equal contribution) \\ \citep{Krizhevsky12} } \\
\hline

Orange &  25.2 &&& Orange Labs International Center Beijing\oi, Beijing University of Posts and Telecommunications\oii & Hongliang BAI\oi, 
Lezi Wang\oii, Shusheng Cen\oii, YiNan Liu\oii, Kun Tao\oi, Wei Liu\oi, Peng Li\oi, Yuan Dong\oi  \\
\hline

OverFeat & 14.2  & {\bf 30.0} & (19.4) & New York University  & \parbox[t]{\crw}{Pierre Sermanet, David Eigen, Michael Mathieu, Xiang Zhang, Rob Fergus, Yann LeCun \\
\citep{Sermanet13} } \\
\hline

Quantum &  82.0 & - & - & Self-employed\oi, Student in Troy High School, Fullerton, CA\oii  & \parbox[t]{\crw}{Henry Shu\oi, Jerry Shu\oii 
\\ \citep{cloudcv} } \\
\hline

SYSU & - & - & 10.5 & Sun Yat-Sen University, China. & \parbox[t]{\crw}{Xiaolong Wang \\ \citep{Felzenszwalb10} }\\
\hline

Toronto & - & - & 11.5 & University of Toronto & Yichuan Tang*, Nitish Srivastava*, Ruslan Salakhutdinov (* = equal contribution)  \\
\hline

Trimps &  26.2 & - & - & The Third Research Institute of the Ministry of Public Security, P.R. China & Jie Shao, Xiaoteng Zhang, Yanfeng Shang, Wenfei Wang, Lin Mei, Chuanping Hu \\
\hline

UCLA &  - &- & 9.8 & University of California Los Angeles  & Yukun Zhu, Jun Zhu, Alan Yuille\\
\hline

UIUC & - & - & 1.0 & University of Illinois at Urbana-Champaign & \parbox[t]{\crw}{Thomas Paine, Kevin Shih, Thomas Huang \\ \citep{Krizhevsky12} } \\
\hline

UvA &  14.3 & - & {\bf 22.6} &  University of Amsterdam, Euvision Technologies &
\parbox[t]{\crw}{ Koen E. A. van de Sande, Daniel H. F. Fontijne, Cees G. M. Snoek,
Harro M. G. Stokman, Arnold W. M. Smeulders \\ \citep{vanDeSande14} } \\ %Uijlings13,vanDeSande10,Krizhevsky12
\hline

VGG &  15.2 & 46.4 & - & Visual Geometry Group, University of Oxford & \parbox[t]{\crw}{ Karen Simonyan,
Andrea Vedaldi, 
Andrew Zisserman \\ \citep{Simonyan13} } \\
\hline

ZF &  13.5 & - & - & New York University & \parbox[t]{\crw}{Matthew D Zeiler, Rob Fergus \\ \citep{Zeiler13,Zeiler11} } \\
\hline

\end{tabular}

\caption{Teams participating in ILSVRC2013, ordered alphabetically. Each method is identified with a codename used in the text. 
For classificaton and single-object localization we report flat top-5 error, in percents (lower is better).  For detection we report mean average precision, in percents (higher is better). Even though the winner of the challenge was determined by the number of object categories won, this correlated strongly with mAP. Parentheses indicate the team used outside training data and was not part of the official competition. Some competing
teams also submitted entries trained with outside data: Clarifai with $11.2\%$ classification error, NEC with $20.9\%$ detection mAP. Key references are provided where available. More details about the winning entries can be found in Section~\ref{sec:MethodsEntries}. }
\label{table:sub13}
\end{table}
\end{landscape}
}

\afterpage{
\begin{landscape}
\clearpage
\begin{table}[H]
\begin{center} {\bf ILSVRC 2014}  \end{center}
\vspace{-0.1in}
\begin{tabular}{p{\nz}>{\centering\nf}p{\cz}>{\centering\nf}p{\cz}>{\centering\nf}p{\cz}>{\centering\nf}p{\cz}>{\centering\nf}p{\cz}>{\centering\nf}p{\cz}>{\crf}p{\ciz}>{\crf}p{\crz}}
\hline
{\bf Codename} & CLS & CLSo & LOC & LOCo & DET & DETo & { \bf Insitutions} & { \bf Contributors and references} \\
\hline
Adobe& - & 11.6& - & {\bf 30.1}& - & - & Adobe\oi, UIUC\oii&\parbox[t]{\crz}{Hailin Jin\oi, Zhaowen Wang\oii, Jianchao Yang\oi, Zhe Lin\oi}\\ 
\hline 
AHoward& 8.1& - & \zl & - & - & - & Howard Vision Technologies &\parbox[t]{\crz}{Andrew Howard~\citep{Howard14} }\\ 
\hline 
BDC & 11.3& - & \zl & - & - & - & Institute for Infocomm Research\oi, Université Pierre et Marie Curie\oii &\parbox[t]{\crz}{Olivier Morère\oi\oii, Hanlin Goh\oi, Antoine Veillard\oii, Vijay Chandrasekhar\oi \citep{Krizhevsky12} }\\ 
\hline 
Berkeley& - & - & - & - & - & 34.5&UC Berkeley  &\parbox[t]{\crz}{Ross Girshick,  Jeff Donahue, Sergio Guadarrama, Trevor Darrell,  Jitendra Malik \citep{Girshick13,Girshick14}}\\ 
\hline 
BREIL& 16.0& - & \zl & - & - & - & KAIST department of EE &\parbox[t]{\crz}{  Jun-Cheol Park, Yunhun Jang, Hyungwon Choi, JaeYoung Jun~\citep{Chatfield14,caffe}}\\ 
\hline 
Brno& 17.6& - & 52.0& - & - & - & Brno University of Technology &\parbox[t]{\crz}{Martin Kol\'a\v{r}, Michal Hradi\v{s}, Pavel Svoboda~\citep{Krizhevsky12,Mikolov13,caffe}}\\ 
\hline 
CASIA-2& - & - & - & - & 28.6& - &Chinese Academy of Science\oi, Southeast University\oii  &\parbox[t]{\crz}{Peihao Huang\oi,  Yongzhen Huang\oi,  Feng Liu\oii,  Zifeng Wu\oi, Fang Zhao\oi, Liang Wang\oi, Tieniu Tan\oi\citep{Girshick14} }\\ 
\hline 
CASIAWS& - & {\bf 11.4}& - & \zl & - & - & CRIPAC, CASIA &\parbox[t]{\crz}{Weiqiang Ren,  Chong Wang,  Yanhua Chen, Kaiqi Huang, Tieniu Tan \citep{Arbelaez14}}\\ 
\hline 
Cldi& 13.9& - & 46.9& - & - & - & KAIST\oi, Cldi Inc.\oii &\parbox[t]{\crz}{Kyunghyun Paeng\oi, Donggeun Yoo\oi, Sunggyun Park\oi, Jungin Lee\oii, Anthony S. Paek\oii, In So Kweon\oi, Seong Dae Kim\oi \citep{Krizhevsky12, Perronnin10}}\\ 
\hline 
CUHK& - & - & - & - & - & 40.7&The Chinese University of Hong Kong  &\parbox[t]{\crz}{Wanli Ouyang, Ping Luo, Xingyu Zeng, Shi Qiu, Yonglong Tian, Hongsheng Li, Shuo Yang, Zhe Wang, Yuanjun Xiong, Chen Qian, Zhenyao Zhu, Ruohui Wang, Chen-Change Loy, Xiaogang Wang, Xiaoou Tang \citep{Ouyang14,Ouyang13}  }\\ 
\hline 
%CUHK-2& - & - & - & - & - & 40.4& The Chinese University of Hong Kong &\parbox[t]{\crz}{Wanli Ouyang, Xingyu Zeng, Shi Qiu, Ping Luo, Yonglong %Tian, Hongsheng Li, Shuo Yang, Zhe Wang, Yuanjun Xiong, Chen Qian, Zhenyao Zhu, Ruohui Wang, Chen-Change Loy, Xiaogang Wang, Xiaoou Tang  }\\ 
%\hline 
DeepCNet& 17.5& - & \zl & - & - & - & University of Warwick &\parbox[t]{\crw}{Ben Graham \citep{Graham13,Schmidhuber12}}\\ 
\hline 
DeepInsight& - & - & - & - & - & 40.5& NLPR\oi, HKUST\oii &\parbox[t]{\crw}{Junjie Yan\oi, Naiyan Wang\oii, Stan Z. Li\oi, Dit-Yan Yeung\oii \citep{Girshick14} }\\ 
\hline 
%DV& 9.5& - & \zl & - & - & - &  DeeperVision &\parbox[t]{\crw}{DeeperVision}\\ 
%\hline 
FengjunLv& 17.4& - & \zl& - & - & - &   Fengjun Lv Consulting &\parbox[t]{\crw}{Fengjun Lv \citep{Krizhevsky12, Harel07}}\\ 
\hline 
GoogLeNet& {\bf 6.7}& - & 26.4& - & - & {\bf 43.9}& Google &\parbox[t]{\crz}{Christian Szegedy, Wei Liu, Yangqing Jia, Pierre Sermanet, Scott Reed, Drago Anguelov, Dumitru Erhan, Andrew Rabinovich~\citep{Szegedy14}}\\ 
\hline 
HKUST& - & - & - & - & 28.9& - &Hong Kong U. of Science and Tech.\oi, Chinese U. of H. K.\oii, Stanford U.\oiii &\parbox[t]{\crz}{Cewu Lu\oi, Hei Law*\oi, Hao Chen*\oii, Qifeng Chen*\oiii,  Yao Xiao*\oi Chi Keung Tang\oi \citep{Uijlings13,Girshick13,Perronnin10,Felzenszwalb10} }\\ 
\hline 
%lffall& - & - & - & - & - & 30.3& Southeast U. &\parbox[t]{\crz}{Feng Liu}\\ 
%\hline 
libccv& 16.0& - & \zl & - & - & - & libccv.org &\parbox[t]{\crz}{Liu Liu \citep{Zeiler13}}\\ 
\hline 
MIL& 18.3& - & 33.7& - & - & 30.4& The University of Tokyo\oi,  IIT Guwahati\oii &\parbox[t]{\crz}{Senthil Purushwalkam\oi\oii, Yuichiro Tsuchiya\oi, Atsushi Kanehira\oi, Asako Kanezaki\oi, Tatsuya Harada\oi  \citep{Kanezaki14,Girshick13}}\\ 
\hline 
MPG\_UT& - & - & - & - & - & 26.4& The University of Tokyo &\parbox[t]{\crz}{Riku Togashi, Keita Iwamoto, Tomoaki Iwase, Hideki Nakayama \citep{Girshick14} }\\ 
\hline 
MSRA & 8.1& - & 35.5& - & 35.1& - & Microsoft Research\oi, Xi'an Jiaotong U.\oii, U. of Science and Tech. of China\oiii &\parbox[t]{\crw}{Kaiming He\oi, Xiangyu Zhang\oii, Shaoqing Ren\oiii, Jian Sun\oi \citep{He14}}\\ 
\hline 
NUS& - & - & - & - & {\bf 37.2}& - &National University of Singapore\oi, IBM Research Australia\oii  &\parbox[t]{\crz}{Jian Dong\oi, Yunchao Wei\oi, Min Lin\oi, Qiang Chen\oii, Wei Xia\oi, Shuicheng Yan\oi  \citep{Lin14,Chen14} }\\ 
\hline 
NUS-BST& 9.8& - & \zl & - & - & - & National Univ. of Singapore\oi, Beijing Samsung Telecom R\&D Center\oi &\parbox[t]{\crz}{Min Lin\oi, Jian Dong\oi, Hanjiang Lai\oi, Junjun Xiong\oii, Shuicheng Yan\oi \citep{Lin14,Howard14,Krizhevsky12} }\\ 
\hline 
Orange& 15.2& 14.8& 42.8& 42.7& - & 27.7& Orange Labs Beijing\oi, BUPT China\oii &\parbox[t]{\crz}{Hongliang Bai\oi,  Yinan Liu\oi, Bo Liu\oii,  Yanchao Feng\oii,  Kun Tao\oi, Yuan Dong\oi  \citep{Girshick14}}\\ 
\hline 
PassBy& 16.7& - & \zl & - & - & - & LENOVO\oi, HKUST\oii, U. of Macao\oiii &\parbox[t]{\crz}{Lin Sun\oi\oii, Zhanghui Kuang\oi, Cong Zhao\oi, Kui Jia\oiii, Oscar C.Au\oii \citep{caffe,Krizhevsky12}}\\ 
\hline 
SCUT& 18.8& - &\zl & - & - & - & South China Univ. of Technology &\parbox[t]{\crz}{Guo Lihua, Liao Qijun, Ma Qianli, Lin Junbin}\\ 
\hline 
Southeast& - & - & - & - & 30.5& - & Southeast U.\oi,  Chinese A. of Sciences\oii &\parbox[t]{\crz}{Feng Liu\oi,  Zifeng Wu\oii,  Yongzhen Huang\oii}\\ 
\hline 
SYSU& 14.4& - & 31.9& - & - & - & Sun Yat-Sen University &\parbox[t]{\crz}{Liliang Zhang, Tianshui Chen, Shuye Zhang, Wanglan He, Liang Lin, Dengguang Pang, Lingbo Liu}\\ 
\hline 
Trimps& - & 11.5& - & 42.2& - & 33.7& The Third Research Institute of the Ministry of Public Security &\parbox[t]{\crz}{Jie Shao, Xiaoteng Zhang, JianYing Zhou, Jian Wang, Jian Chen, Yanfeng Shang, Wenfei Wang, Lin Mei, Chuanping Hu \citep{Girshick14,Manen13,Howard14} }\\ 
\hline 
TTIC& 10.2& - & 48.3& - & - & - & Toyota Technological Institute at Chicago\oi, Ecole Centrale Paris\oii &\parbox[t]{\crz}{George Papandreou\oi, Iasonas Kokkinos\oii \citep{Papandreou14a,Papandreou14c,Jojic03,Krizhevsky12,Sermanet13,Dubout12,Iandola12} }\\ 
\hline 
UI& 99.5& - & \zl & - & - & - &  University of Isfahan &\parbox[t]{\crz}{Fatemeh Shafizadegan, Elham Shabaninia \citep{Yang09} }\\ 
\hline 
UvA& 12.1& - & \zl & - & 32.0& 35.4& U. of Amsterdam and Euvision Tech. &\parbox[t]{\crz}{Koen van de Sande, Daniel Fontijne, Cees Snoek, Harro Stokman, Arnold Smeulders~\citep{vanDeSande14}  }\\ 
\hline 
VGG& 7.3& - & {\bf 25.3}& - & - & - & University of Oxford &\parbox[t]{\crz}{Karen Simonyan,  Andrew Zisserman \citep{Simonyan14}}\\ 
\hline 
%Virginia Tech& - & - & - & - & - & 30.3&  &\parbox[t]{\crz}{Akrit Mohapatra, Neelima Chavali  Virginia Tech}\\ 
%\hline 
XYZ& 11.2& - & \zl & - & - & - &The University of Queensland  &\parbox[t]{\crz}{Zhongwen Xu and Yi Yang \citep{Krizhevsky12,caffe,Zeiler13,Lin14}}\\ 
\hline 

\end{tabular}
\caption{ Teams participating in ILSVRC2014, ordered alphabetically. Each method is identified with a codename used in the text. For classificaton and single-object localization we report flat top-5 error, in percents (lower is better).  For detection we report mean average precision, in percents (higher is better). CLSo,LOCo,DETo corresponds to entries using outside training data (officially allowed in ILSVRC2014).  \zl{ } means localization error greater than $60\%$ (localization submission was required with every classification submission).
 Key references are provided where available. More details about the winning entries can be found in Section~\ref{sec:MethodsEntries}. }
\label{table:sub14}
\end{table}
\end{landscape}
}

The winning image classification with provided data team was GoogLeNet, which explored an improved convolutional neural network architecture combining the multi-scale idea with intuitions gained from the Hebbian principle. Additional dimension reduction layers allowed them to increase both the depth and the width of the network significantly without incurring significant computational overhead. In the image classification with external data track, CASIAWS won by using weakly supervised object localization from only classification labels to improve image classification. MCG region proposals~\citep{Arbelaez14} pretrained on PASCAL VOC 2012 data are used to extract region proposals, regions are represented using convolutional networks, and a multiple instance learning strategy is used to learn weakly supervised object detectors to represent the image.

 In the single-object localization with provided data track, the winning team was VGG, which explored the effect of convolutional neural network depth on its accuracy by using three different architectures with up to 19 weight layers with rectified linear unit non-linearity, building off of the implementation of Caffe~\citep{caffe}. For localization they used per-class bounding box regression similar to OverFeat~\citep{Sermanet13}. In the single-object localization with external data track, Adobe used 2000 additional ImageNet classes to train the classifiers in an integrated convolutional neural network framework for both classification and localization, with bounding box regression. At test time they used k-means to find bounding box clusters and rank the clusters according to the classification scores.

In the object detection with provided data track, the winning team  NUS used the RCNN framework~\citep{Girshick13} with the network-in-network method \citep{Lin14}  and improvements of \citep{Howard14}. Global context information was incorporated following~\citep{Chen14}. In the object detection with external data track, the winning team was GoogLeNet (which also won image classification with provided data). It is truly remarkable that the same team was able to win at both image classification and object detection, indicating that their methods are able to not only classify the image based on scene information but also accurately localize multiple object instances. Just like most teams participating in this track, GoogLeNet used the image classification dataset  as extra training data.

%%%%%%%%%%%%%%%%%%%%%%%%%%%%%%%%%%%%%%%%%%%%%%%%%%%%%%%%%%%%%%%%%%%%%%%%%%%%%%%%%%%%%%%%%%
%%%%%%%%%%%                                                            %%%%%%%%%%%%%%%%%%%%%%%%%%%%%%%%%%%%%%%%%%%%%%%%%%%
%%%%%%%%%%%        Algorithmic innovation                 %%%%%%%%%%%%%%%%%%%%%%%%%%%%%%%%%%%%%%%%%%%%%%%%%%%
%%%%%%%%%%%                                          	             %%%%%%%%%%%%%%%%%%%%%%%%%%%%%%%%%%%%%%%%%%%%%%%%%%%
%%%%%%%%%%%%%%%%%%%%%%%%%%%%%%%%%%%%%%%%%%%%%%%%%%%%%%%%%%%%%%%%%%%%%%%%%%%%%%%%%%%%%%%%%%

\subsection{Large scale algorithmic innovations}
\label{sec:MethodsDeep}

ILSVRC over the past five years has paved the way for several breakthroughs in computer vision. 

The field of categorical object recognition has dramatically evolved in the large-scale setting. 
Section~\ref{sec:MethodsEntries} documents the progress, starting from coded SIFT features and evolving to large-scale
convolutional neural networks dominating at all three tasks of image classification, single-object localization, and object detection. With the availability of so much training data (along with an efficient algorithmic implementation and GPU computing resources) it became possible to learn neural networks directly from the image data, without needing to create  multi-stage hand-tuned pipelines of extracted features and discriminative classifiers. The major breakthrough came in 2012 with the win of the SuperVision team on image classification and single-object localization tasks~\citep{Krizhevsky12}, and by 2014 all of the top contestants were relying heavily on convolutional neural networks. 

%The success of convolutional neural networks inspired a variety of innovations in that field, such as %\todo{visualization methods, etc.}

Further, over the past few years there has been a lot of focus on large-scale recognition in the computer vision community . Best paper awards at top vision conferences in 2013 were awarded to large-scale recognition methods: at CVPR 2013 to "Fast, Accurate Detection of 100,000 Object Classes on a Single Machine" \citep{Dean13} and at ICCV 2013 to "From Large Scale Image Categorization to Entry-Level Categories" \citep{Ordonez13}. Additionally, several influential lines of research have emerged, such as large-scale weakly supervised localization work of~\citep{Kuettel12}
which was awarded the best paper award in ECCV 2012 and large-scale zero-shot learning, e.g.,~\citep{Frome13}.

%%%%%%%%%%%%%%%%%%%%%%%%%%%%%%%%%%%%%%%%%%%%%%%%%%%%%%%%%%%%%%%%%%%%%%%%%%%%%%%%%%%%%%%%%%
%%%%%%%%%%%%%%%%%%%%%%%%%%%%%%%%%%%%%%%%%%%%%%%%%%%%%%%%%%%%%%%%%%%%%%%%%%%%%%%%%%%%%%%%%%
%%%%%%%%%%%                                           %%%%%%%%%%%%%%%%%%%%%%%%%%%%%%%%%%%%%%%%%%%%%%%%%%%%%%%%%%%
%%%%%%%%%%%        RESULTS                     %%%%%%%%%%%%%%%%%%%%%%%%%%%%%%%%%%%%%%%%%%%%%%%%%%%%%%%%%%%
%%%%%%%%%%%                                           %%%%%%%%%%%%%%%%%%%%%%%%%%%%%%%%%%%%%%%%%%%%%%%%%%%%%%%%%%%
%%%%%%%%%%%%%%%%%%%%%%%%%%%%%%%%%%%%%%%%%%%%%%%%%%%%%%%%%%%%%%%%%%%%%%%%%%%%%%%%%%%%%%%%%%
%%%%%%%%%%%%%%%%%%%%%%%%%%%%%%%%%%%%%%%%%%%%%%%%%%%%%%%%%%%%%%%%%%%%%%%%%%%%%%%%%%%%%%%%%%

\section{Results and analysis}
\label{sec:Results}

%%%%%%%%%%%%%%%%%%%%%%%%%%%%%%%%%%%%%%%%%%%%%%%%%%%%%%%%%%%%%%%%%%%%%%%%%%%%%%%%%%%%%%%%%%
%%%%%%%%%%%                                                            %%%%%%%%%%%%%%%%%%%%%%%%%%%%%%%%%%%%%%%%%%%%%%%%%%%
%%%%%%%%%%%        Improvements over years              %%%%%%%%%%%%%%%%%%%%%%%%%%%%%%%%%%%%%%%%%%%%%%%%%%%
%%%%%%%%%%%                                          	             %%%%%%%%%%%%%%%%%%%%%%%%%%%%%%%%%%%%%%%%%%%%%%%%%%%
%%%%%%%%%%%%%%%%%%%%%%%%%%%%%%%%%%%%%%%%%%%%%%%%%%%%%%%%%%%%%%%%%%%%%%%%%%%%%%%%%%%%%%%%%%

\subsection{Improvements over the years}
\label{sec:ResultsYears}

\begin{figure}
\includegraphics[width=\linewidth]{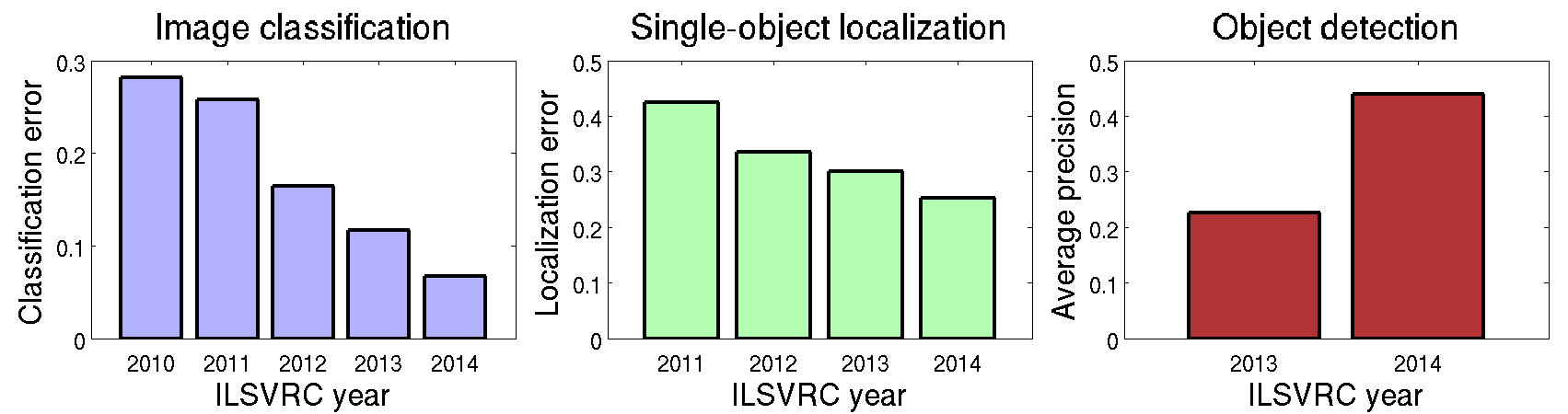}
\caption{Performance of winning entries in the ILSVRC2010-2014 competitions in each of the three tasks (details about the entries and numerical results are in Section~\ref{sec:MethodsEntries}). There is a steady reduction of error every year in object classification and single-object localization tasks, and a 1.9x improvement in mean average precision in object detection. There are two considerations in making these comparisons. (1) The object categories used in ISLVRC changed between years 2010 and 2011, and between 2011 and 2012. However,
the large scale of the data (1000 object categories, 1.2 million training images) has remained the same, making it possible to compare results. Image classification and single-object localization entries shown here use only provided
training data.  (2) The size of the object detection training data has increased significantly between years 2013 and 2014 (Section~\ref{sec:AnnotDet}).
Section~\ref{sec:ResultsYears} discusses the relative effects of training data increase versus algorithmic improvements.  
 }
\label{fig:years}
\end{figure}

State-of-the-art accuracy has improved significantly from ILSVRC2010 to ILSVRC2014, showcasing the massive progress that has been made in large-scale object recognition over the past five years. The performance of the winning ILSVRC entries for each task and each year are shown in Figure~\ref{fig:years}. The improvement over the years is clearly visible. In this section we quantify and analyze this improvement.

\subsubsection{Image classification and single-object localization improvement over the years}

There has been a 4.2x reduction in image classification error (from $28.2\%$ to $6.7\%$) and a 1.7x reduction
in single-object localization error (from $42.5\%$ to $25.3\%$) since the beginning of the challenge. 
For consistency, here we consider only teams that use the provided training data.
Even though the exact object categories have changed (Section~\ref{sec:AnnotClsObjects}), the large scale of the dataset has remained the same (Table~\ref{table:clslocstats}), making the results comparable across the years. The dataset has not changed since 2012, and there has been a 2.4x reduction in image classification error (from $16.4\%$ to $6.7\%$) and a 1.3x in single-object localization error (from $33.5\%$ to $25.3\%$) in the past three years.

\subsubsection{Object detection improvement over the years}

Object detection accuracy as measured by the mean average precision (mAP) has increased 1.9x since the introduction of this task, from $22.6\%$ mAP in ILSVRC2013 to $43.9\%$ mAP in ILSVRC2014. However, these results are not directly comparable for two reasons. First, the size of the object detection training data has increased significantly from 2013 to 2014 (Section~\ref{sec:AnnotDet}). 
Second, the $43.9\%$ mAP result was obtained with the addition of the image classification and single-object localization training data. Here we attempt to understand the relative effects of the training set size increase versus algorithmic improvements. All models are evaluated on the same ILSVRC2013-2014 object detection test set.

First, we quantify the effects of increasing detection training data between the two challenges by comparing the same model trained on ILSVRC2013 detection data versus ILSVRC2014 detection data.  The UvA team's framework from 2013 achieved $22.6\%$ with ILSVRC2013 data  (Table~\ref{table:sub13}) and $26.3\%$ with ILSVRC2014 data and no other modifications.\footnote{Personal communication with members of the UvA team.} The absolute increase in mAP was $3.7\%$.
The RCNN model  achieved $31.4\%$ mAP with ILSVRC2013 detection plus image classification data~\citep{Girshick13} and $34.5\%$ mAP with ILSVRC2014 detection plus image classification data (Berkeley team in Table~\ref{table:sub14}). The absolute increase in mAP by expanding ILSVRC2013 detection data to ILSVRC2014 was $3.1\%$. 

Second, we quantify the effects of adding in the external data for training object detection models. The NEC model in 2013 achieved $19.6\%$ mAP trained on ILSVRC2013 detection data alone and $20.9\%$ mAP trained on ILSVRC2013 detection plus classification data (Table~\ref{table:sub13}). The absolute increase in mAP was $1.3\%$. The UvA team's best entry in 2014 achieved $32.0\%$ mAP trained on ILSVRC2014 detection data and  $35.4\%$ mAP trained on ILSVRC2014 detection plus classification data. The absolute increase in mAP was $3.4\%$. 

Thus, we conclude based on the evidence so far that expanding the ILSVRC2013 detection set to the ILSVRC2014 set, as well as adding in additional training data from the classification task, all account for approximately $1-4\%$ in absolute mAP improvement for the models. For comparison, we can also attempt to quantify the effect of algorithmic innovation. The UvA team's 2013 framework achieved $26.3\%$ mAP on ILSVRC2014 data as mentioned above, and their improved method in 2014 obtained $32.0\%$ mAP (Table~\ref{table:sub14}). This is $5.8\%$ absolute increase in mAP over just one year from algorithmic innovation alone. 

In summary, we conclude that the absolute $21.3\%$ increase in mAP between winning entries of ILSVRC2013 ($22.6\%$ mAP) and of ILSVRC2014 ($43.9\%$ mAP) is the result of impressive algorithmic innovation and not just a consequence of increased training data. However, increasing the ISLVRC2014 object detection training dataset further \emph{is} likely to produce additional improvements in detection accuracy for current algorithms.

 % For comparison, 

%We can
%again see that even this additional increase in training data is extremely unlikely to account for the $21.3\%$ absolute increase between the winning entries of %ILSVRC2013 and ILSVRC2014. Further, 

%Finally, to the best of our knowledge the highest reported object detection performance on ILSVRC2013 data alone to date is $31.4\%$ mean AP with the RCNN %model~\citep{Girshick13}, significantly improved from state-of-the-art of $22.6\%$ mAP in ILSVRC2013. In conclusion, the rate of progress in object detection %and classification has been truly astounding since the inception of ILSVRC!

%The increase in absolute mAP is $3.4\%$. Thus we conclude that
%(1) even the addition of this extra data is extremely unlikely to account for the  %ILSVRC2014, and (2) 

%Finally, we can also loosely quantify the rate of algorithmic improvements. The UvA team's 2013 entry trained on ILSVRC2014 data achieved $26.2\%$ mAP,  %Further, 

%Additionally, the $22.6\%$ mAP in 2013 was obtained by the UvA team, which achieved up to $35.4\%$ by ILSVRC2014. 

%%%%%%%%%%%%%%%%%%%%%%%%%%%%%%%%%%%%%%%%%%%%%%%%%%%%%%%%%%%%%%%%%%%%%%%%%%%%%%%%%%%%%%%%%%
%%%%%%%%%%%                                                            %%%%%%%%%%%%%%%%%%%%%%%%%%%%%%%%%%%%%%%%%%%%%%%%%%%
%%%%%%%%%%%        Statistical significance?                %%%%%%%%%%%%%%%%%%%%%%%%%%%%%%%%%%%%%%%%%%%%%%%%%%%
%%%%%%%%%%%                                          	             %%%%%%%%%%%%%%%%%%%%%%%%%%%%%%%%%%%%%%%%%%%%%%%%%%%
%%%%%%%%%%%%%%%%%%%%%%%%%%%%%%%%%%%%%%%%%%%%%%%%%%%%%%%%%%%%%%%%%%%%%%%%%%%%%%%%%%%%%%%%%%

\subsection{Statistical significance}
\label{sec:ResultsStats}

\begin{table}
%\begin{center}Image Classification\end{center}
\centering
{\bf Image classification}
\begin{tabular}{llcc}
\hline
Year & Codename & Error (percent) & $99.9\%$ Conf Int\\
\hline
{\bf 2014} & {\bf GoogLeNet} & {\bf 6.66}  & {\bf 6.40 - 6.92}\\
2014 & VGG & 7.32  & 7.05 - 7.60\\
2014 & MSRA & 8.06  & 7.78 - 8.34\\
2014 & AHoward & 8.11  & 7.83 - 8.39\\
2014 & DeeperVision & 9.51  & 9.21 - 9.82\\  %%%%%%%%%%%%%%%%%%%%%%
2013 & Clarifai\oi & 11.20  & 10.87 - 11.53\\
2014 & CASIAWS\oi & 11.36  & 11.03 - 11.69\\
2014 & Trimps\oi & 11.46  & 11.13 - 11.80\\
2014 & Adobe\oi & 11.58  & 11.25 - 11.91\\
{\bf 2013} & {\bf Clarifai} & {\bf 11.74}  & {\bf 11.41 - 12.08}\\
2013 & NUS & 12.95  & 12.60 - 13.30\\
2013 & ZF & 13.51  & 13.14 - 13.87\\
2013 & AHoward & 13.55  & 13.20 - 13.91\\
2013 & OverFeat & 14.18  & 13.83 - 14.54\\  %%%%%%%%%%%%%%%%%%%%%%
2014 & Orange\oi & 14.80  & 14.43 - 15.17\\
2012 & SuperVision\oi & 15.32  & 14.94 - 15.69\\
{\bf 2012} & {\bf SuperVision} & {\bf 16.42}  & {\bf 16.04 - 16.80}\\
2012 & ISI & 26.17  & 25.71 - 26.65\\
2012 & VGG & 26.98  & 26.53 - 27.43\\
2012 & XRCE & 27.06  & 26.60 - 27.52\\
2012 & UvA & 29.58  & 29.09 - 30.04\\  %%%%%%%%%%%%%%%%%%%%%%
\hline
\end{tabular}

%\begin{center}Single-object localization\end{center}
{\bf Single-object localization}
\begin{tabular}{llcc}
\hline
Year & Codename & Error (percent) &  $99.9\%$ Conf Int\\
\hline
{\bf 2014} & {\bf VGG} & {\bf 25.32}  & {\bf 24.87 - 25.78}\\
2014 & GoogLeNet & 26.44  & 25.98 - 26.92\\
{\bf 2013} & {\bf OverFeat} & {\bf 29.88}  & {\bf 29.38 - 30.35}\\
2014 & Adobe\oi & 30.10  & 29.61 - 30.58\\
2014 & SYSU & 31.90  & 31.40 - 32.40\\
2012 & SuperVision\oi & 33.55  & 33.05 - 34.04\\
2014 & MIL & 33.74  & 33.24 - 34.25\\
{\bf 2012} & {\bf SuperVision} & {\bf 34.19}  & {\bf 33.67 - 34.69}\\
2014 & MSRA & 35.48  & 34.97 - 35.99\\ %%%%%%%%%%%%%%%%%%%%%%
2014 & Trimps\oi & 42.22  & 41.69 - 42.75\\
2014 & Orange\oi & 42.70  & 42.18 - 43.24\\
2013 & VGG & 46.42  & 45.90 - 46.95\\
2012 & VGG & 50.03  & 49.50 - 50.57\\
2012 & ISI & 53.65  & 53.10 - 54.17\\
2014 & CASIAWS\oi & 61.96  & 61.44 - 62.48\\
\hline
\end{tabular}

%\begin{center}Object detection\end{center}
{\bf Object detection}
\begin{tabular}{llcc}
\hline
Year & Codename & AP (percent) &  $99.9\%$ Conf Int\\
\hline
2014 & GoogLeNet\oi & 43.93  & 42.92 - 45.65\\
2014 & CUHK\oi & 40.67  & 39.68 - 42.30\\
2014 & DeepInsight\oi & 40.45  & 39.49 - 42.06\\
{\bf 2014} & {\bf NUS} & {\bf 37.21}  & {\bf 36.29 - 38.80}\\
2014 & UvA\oi & 35.42  & 34.63 - 36.92\\
2014 &MSRA & 35.11  & 34.36 - 36.70\\
2014 & Berkeley\oi & 34.52  & 33.67 - 36.12\\  %%%%%%%%%%%%%%%%%%%%%%
2014 & UvA & 32.03  & 31.28 - 33.49\\
2014 & Southeast & 30.48  & 29.70 - 31.93\\
2014 & HKUST & 28.87  & 28.03 - 30.20\\  %%%%%%%%%%%%%%%%%%%%%%
{\bf 2013} & {\bf UvA} & {\bf 22.58}  & {\bf 22.00 - 23.82}\\
2013 & NEC\oi & 20.90  & 20.40 - 22.15\\
2013 & NEC & 19.62  & 19.14 - 20.85\\
2013 & OverFeat\oi & 19.40  & 18.82 - 20.61\\
2013 & Toronto & 11.46  & 10.98 - 12.34\\
2013 & SYSU & 10.45  & 10.04 - 11.32\\
2013 & UCLA & 9.83  & 9.48 - 10.77\\  %%%%%%%%%%%%%%%%%%%%%%
\hline
\end{tabular}

\caption{We use bootstrapping to construct $99.9\%$ confidence intervals around the
result of up to top 5 submissions to each ILSVRC task in 2012-2014. \oi means the entry used external training data. The winners using the provided data for each track and each year are bolded. The difference between the winning method and the runner-up each year is significant even at the $99.9\%$ level. }
\label{table:bootstrap}
\end{table}

One important question to ask is whether results of different submissions to ILSVRC are statistically significantly different from each other.
Given the large scale, it is no surprise that even minor differences in accuracy are statistically significant; 
we seek to quantify exactly how much of a difference is enough.

Following the strategy employed by  PASCAL VOC \citep{PASCALIJCV2}, for each method we obtain a 
confidence interval of its score using bootstrap sampling. During each bootstrap round, we sample $N$ images with replacement
from all the available $N$ test images and evaluate the performance of the algorithm 
on those sampled images. This can be done very efficiently by precomputing the accuracy on each image. 
Given the results of all the bootstrapping rounds we discard the lower and the upper
 $\alpha$ fraction. The range of the remaining results represents the $1-2\alpha$ confidence interval.
We run a large number of bootstrapping rounds
(from 20,000 until convergence).   Table~\ref{table:bootstrap} shows
the results of the top entries to each task of ILSVRC2012-2014. The winning methods are statistically significantly different
from the other methods, even at the $99.9\%$ level.

%%%%%%%%%%%%%%%%%%%%%%%%%%%%%%%%%%%%%%%%%%%%%%%%%%%%%%%%%%%%%%%%%%%%%%%%%%%%%%%%%%%%%%%%%%
%%%%%%%%%%%                                                            %%%%%%%%%%%%%%%%%%%%%%%%%%%%%%%%%%%%%%%%%%%%%%%%%%%
%%%%%%%%%%%        Per-class evaluation                    %%%%%%%%%%%%%%%%%%%%%%%%%%%%%%%%%%%%%%%%%%%%%%%%%%%
%%%%%%%%%%%                                          	             %%%%%%%%%%%%%%%%%%%%%%%%%%%%%%%%%%%%%%%%%%%%%%%%%%%
%%%%%%%%%%%%%%%%%%%%%%%%%%%%%%%%%%%%%%%%%%%%%%%%%%%%%%%%%%%%%%%%%%%%%%%%%%%%%%%%%%%%%%%%%%

\subsection{Current state of categorical object recognition}
\label{sec:ResultsICCV}

Besides looking at just the average accuracy across hundreds of object categories and tens of thousands of images, we can also delve deeper to understand where mistakes are being made and where researchers' efforts should be focused to expedite progress. 

To do so, in this section we will be analyzing an ``optimistic'' measurement of state-of-the-art recognition performance instead of focusing on the differences in individual algorithms. For each task and each object class, we compute the best performance of \emph{any} entry submitted to \emph{any} ILSVRC2012-2014, including methods using additional training data. Since the test sets have remained the same, we can directly compare all the entries in the past three years to obtain the most ``optimistic'' measurement of state-of-the-art accuracy on each category. 

For consistency with the object detection metric (higher is better), in this section we will be using image classification and single-object localization \emph{accuracy} instead of error, where $accuracy = 1-error$. 

\begin{figure}
\includegraphics[width=\linewidth]{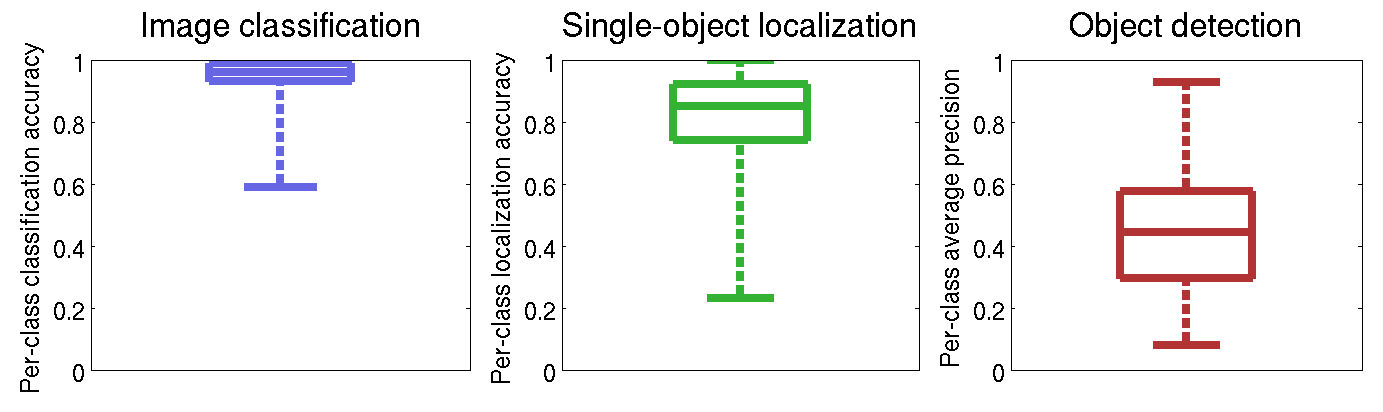}
\caption{For each object class, we consider the best performance of any entry submitted to ILSVRC2012-2014, including entries using additional training data. The plots show the distribution of these ``optimistic'' per-class results.  Performance is measured as accuracy for image classification (left) and for single-object localization (middle), and as average precision for object detection (right). While the results are very promising in image classification, the ILSVRC datasets are far from saturated: many object classes continue to be challenging for current algorithms. }
\label{fig:yearscats}
\end{figure}

\subsubsection{Range of accuracy across object classes}

Figure~\ref{fig:yearscats} shows the distribution of accuracy achieved by the ``optimistic'' models across the object categories. The image classification model achieves $94.6\%$ accuracy on average (or $5.4\%$ error), but there remains a $41.0\%$ absolute difference inaccuracy between the most and least accurate object class. The single-object localization model achieves $81.5\%$ accuracy on average (or $18.5\%$ error), with a $77.0\%$  range in accuracy across the object classes. The object detection model achieves $44.7\%$ average precision, with an $84.7\%$ range across the object classes. It is clear that the ILSVRC dataset is far from saturated: performance on many categories has remained poor despite the strong overall performance of the models.

\begin{figure*}
\centering \large
\begin{tabular}{c}
{\bf  Image classification} \\
Easiest classes \\
\includegraphics[height=1.6in]{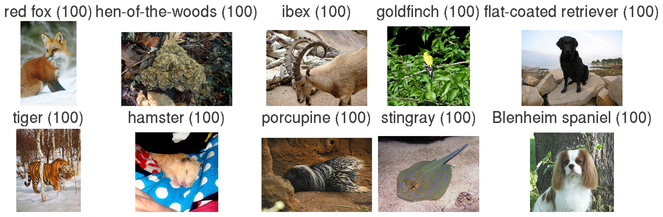} \\
Hardest classes \\
\includegraphics[height=1.9in]{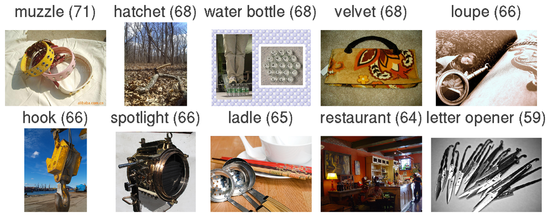} \\
\vspace{0.1in} \\
{\bf  Single-object localization} \\
Easiest classes \\
\includegraphics[height=1.5in]{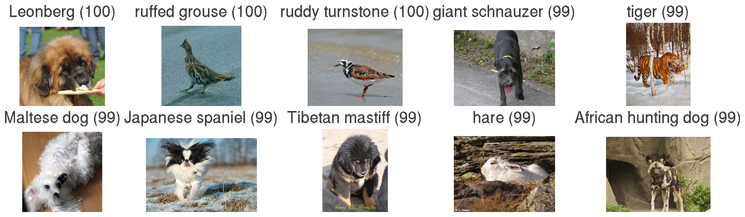} \\
Hardest classes \\
\includegraphics[height=1.9in]{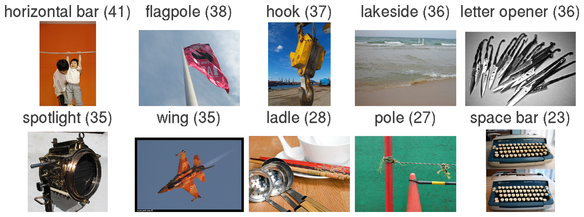} \\
\end{tabular}
\caption{For each object category, we take the best performance of any entry submitted to ILSVRC2012-2014 (including entries using additional training data). Given these ``optimistic'' results we show the easiest and harder classes for each task. The numbers in parentheses indicate classification and localization accuracy. For image classification the 10 easiest classes are randomly selected from among 121 object classes with $100\%$ accuracy. Object detection results are shown in Figure~\ref{fig:bestworstdetcats}.} 
\label{fig:bestworstclscats}
\end{figure*}

\begin{figure*}
\centering \large
\begin{tabular}{c}
{\bf Object detection} \\
Easiest classes \\
\includegraphics[width=0.7\linewidth]{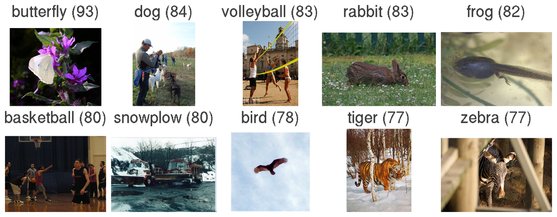} \\
Hardest classes \\
\includegraphics[width=0.7\linewidth]{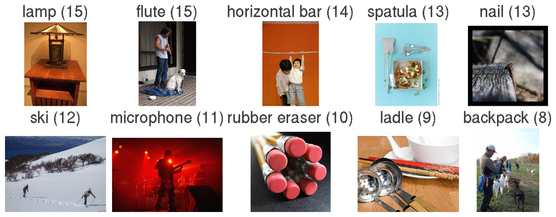} \\
\hline
\end{tabular}
\caption{For each object category, we take the best performance of any entry submitted to ILSVRC2012-2014 (including entries using additional training data). 
Given these ``optimistic'' results we show the easiest and harder classes for the object detection task, i.e., classes with best and worst results. The numbers in parentheses indicate average precision. Image classification and single-object localization results are shown in Figure~\ref{fig:bestworstclscats}.}
\label{fig:bestworstdetcats}
\vspace{0.5in}
\end{figure*}

\subsubsection{Qualitative examples of easy and hard  classes}

Figures~\ref{fig:bestworstclscats} and~\ref{fig:bestworstdetcats} show the easiest and hardest classes for each task, i.e., classes with the best and worst results obtained with the ``optimistic'' models.

For image classification, 121 out of 1000 object classes have $100\%$ image classification accuracy according to the optimistic estimate. Figure~\ref{fig:bestworstclscats} (top) shows a random set of 10 of them. They contain a variety of classes, such as mammals like ``red fox'' and animals with distinctive structures like ``stingray''.  The hardest classes in the image classification task, with accuracy as low as $59.0\%$, include metallic and see-through man-made objects, such as ``hook''  and ``water bottle,'' the material ``velvet''  and the highly varied scene class ``restaurant.''

For single-object localization, the 10 easiest classes with $99.0-100\%$ accuracy are all mammals and birds. The hardest classes include metallic man-made objects such as ``letter opener'' and ``ladle'', plus thin structures such as ``pole'' and ``spacebar'' and highly varied classes such as ``wing''. The most challenging class ``spacebar'' has a only $23.0\%$ localization accuracy. 

Object detection results are shown in Figure~\ref{fig:bestworstdetcats}. The easiest classes are living organisms such as ``dog'' and ``tiger'', plus ``basketball'' and ``volleyball''  with distinctive shape and color, and a somewhat surprising ``snowplow.'' The easiest class ``butterfly'' is not yet perfectly detected but is very close with $92.7\%$ AP. The hardest classes are as expected small thin objects such as  ``flute''  and ``nail'', and the highly varied ``lamp'' and ``backpack'' classes, with as low as $8.0\%$ AP.

\subsubsection{Per-class accuracy as a function of image properties}
\label{sec:imageproperties}

We now take a closer look at the image properties to try to understand why current algorithms perform well on some object classes but not others. One hypothesis is that variation in accuracy comes from the fact that instances of some classes tend to be much smaller in images than instances of other classes, and smaller objects may be harder for computers to recognize. In this section we argue that while accuracy is correlated with object scale in the image, not all variation in accuracy can be accounted for by scale alone.

For every object class, we compute its \emph{average scale}, or 
the average fraction of image area occupied by an instance of the object class on the ILSVRC2012-2014 validation set.
Since the images and object classes in the image classification and single-object localization tasks are the same,  we use the bounding box annotations of the single-object localization dataset for both tasks. In that dataset the object classes range from ``swimming trunks'' with scale of $1.5\%$ to ``spider web'' with scale of $85.6\%$. In the object detection validation dataset the object classes range from ``sunglasses'' with scale of $1.3\%$ to ``sofa'' with scale of $44.4\%$.

\begin{figure*}
\includegraphics[width=\linewidth]{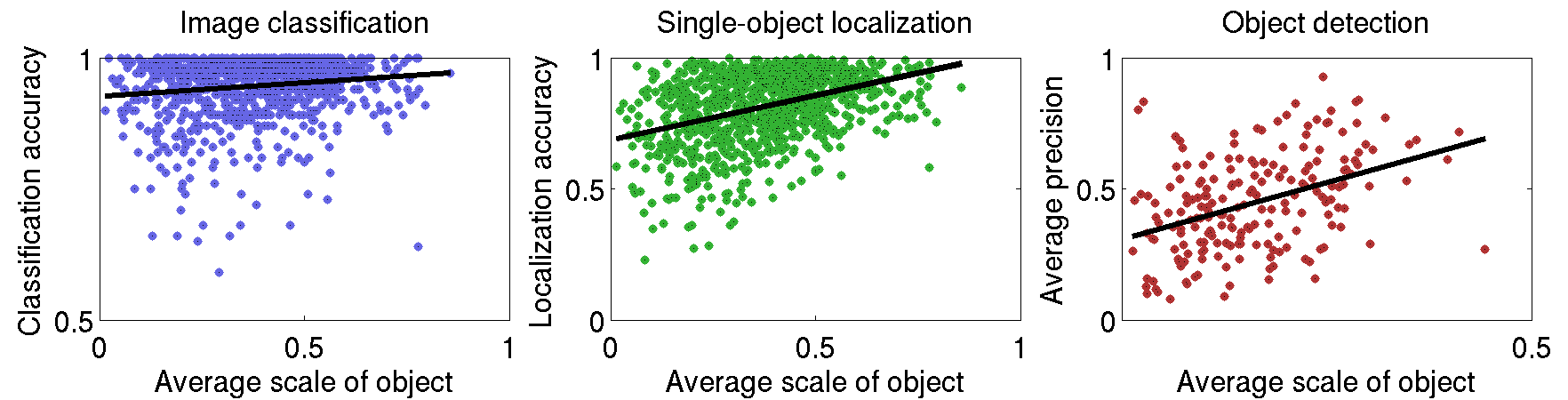}
\caption{Performance of the ``optimistic'' method  as a function of object scale in the image, on each task. Each dot corresponds to one object class. Average scale (x-axis) is computed as the average fraction of the image area occupied by an instance of that object class on the ILSVRC2014 validation set.  ``Optimistic'' performance (y-axis) corresponds to  
the best performance on the test set of any entry submitted to ILSVRC2012-2014 (including entries with additional training data). The test set has remained the same over these three years.
We see that accuracy tends to increase as the objects get bigger in the image. However, it is clear that
far from all the variation in accuracy on these classes can be accounted for by scale alone.
}
\label{fig:scalecats}
\end{figure*}

Figure~\ref{fig:scalecats} shows the performance of the ``optimistic'' method as a function of the average scale of the object in the image. Each dot corresponds to one object class. We observe a very weak positive correlation between object scale and image classification accuracy: $\rho = 0.14$. For single-object localization and object detection the correlation is stronger, at $\rho = 0.40$ and $\rho = 0.41$ respectively. It is clear that not all variation in accuracy can be accounted for by scale alone. Nevertheless, in the next section we will normalize for object scale to ensure that this factor is not affecting our conclusions.

\subsubsection{Per-class accuracy as a function of object properties.}
\label{sec:objectproperties}

\iffalse
\begin{figure}
\begin{tabular}{c|c}
\hline
\multirow{2}{*}{{\bf Real-world size}} \\
\hline
XS: \includegraphics[width=0.48\linewidth]{ex_cls_real_size_1.png} &
\includegraphics[width=0.32\linewidth]{ex_det_real_size_1.png} \\
\hline

\end{tabular}

\end{figure}
\fi

Besides considering image-level properties we can also observe  how accuracy changes as a function of intrinsic object properties. We define three properties inspired by human vision: the real-world size of the object, whether it's deformable within instance, and how textured it is. For each property, the object classes are assigned to one of a few bins (listed below). These properties are illustrated in Figure~\ref{fig:diversity}.

Human subjects annotated each of the 1000 image classification and single-object localization object classes from ILSVRC2012-2014 with these properties.~\citep{Russakovsky13}. By construction (see Section~\ref{sec:AnnotDetObjects}), each of the 200 object detection classes is either also one of 1000 object classes or is an ancestor of one or more of the 1000 classes in the ImageNet hierarchy. To compute the values of the properties for each object detection class, we simply average the annotated values of the descendant classes.

In this section we draw the following  conclusions about state-of-the-art recognition accuracy as a function of these object properties:
\begin{itemize}
\item {\bf Real-world size:}  \emph{XS for extra small (e.g. nail), small (e.g. fox), medium (e.g. bookcase), large (e.g. car) or XL for extra large (e.g. church)} \\
The image classification and single-object localization ``optimistic'' models performs better on large and extra large real-world objects than on smaller ones. The ``optimistic'' object detection model surprisingly performs better on extra small objects than on small or medium ones.
\item {\bf Deformability within instance:} \emph{Rigid (e.g., mug) or deformable (e.g., water snake)}\\
The ``optimistic'' model on each of the three tasks performs statistically significantly better on deformable objects compared to rigid ones. However, this effect disappears when analyzing natural objects separately from man-made objects.
\item {\bf Amount of texture:} \emph{none (e.g. punching bag),  low (e.g. horse), medium (e.g. sheep) or high (e.g. honeycomb)}\\
The ``optimistic'' model on each of the three tasks is significantly better on objects with at least low level of texture compared to  untextured objects.
\end{itemize} 
These and other findings are justified and discussed in detail below.

\paragraph{Experimental setup.}
We observed in Section~\ref{sec:imageproperties} that objects that occupy a larger area in the image tend to be somewhat easier to recognize. To make sure that differences in object scale are not influencing results in this section, we normalize each bin by object scale. We discard object classes with the largest scales from each bin as needed until the average object scale of  object classes in each bin across one property is the same (or as close as possible). For real-world size property for example, the resulting average object scale in each of the five bins is $31.6\%-31.7\%$ in the image classification and single-object localization tasks, and $12.9\%-13.4\%$ in the object detection task.\footnote{For rigid versus deformable objects, the average scale in each bin is $34.1\%-34.2\%$ for classification and localization, and $13.5\%-13.7\%$ for detection. For texture, the average scale in each of the four bins is $31.1\%-31.3\%$ for classification and localization, and $12.7\%-12.8\%$ for detection.}

\begin{figure*}
\centering \large
\begin{tabular}{c}
{\bf Real-world size} \\
\includegraphics[width=0.57\linewidth]{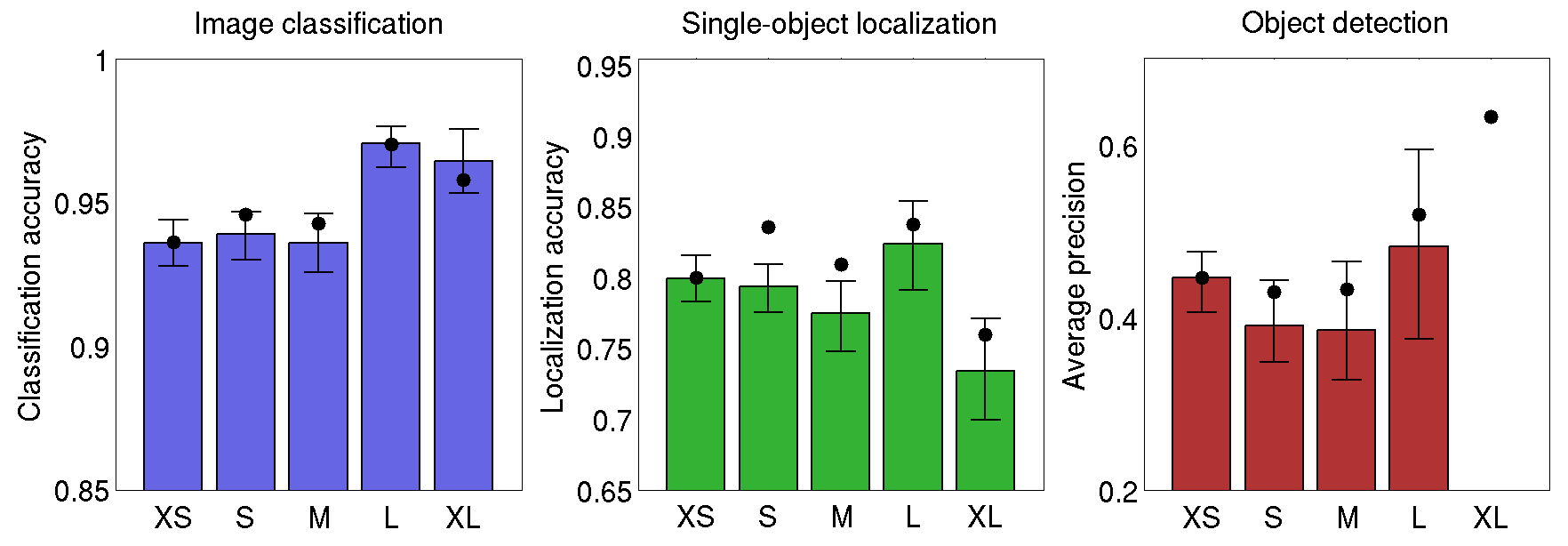} 
\vspace{0.02in} \\
\vspace{0.01in}\\
{\bf Deformability within instance}\\
\includegraphics[width=0.57\linewidth]{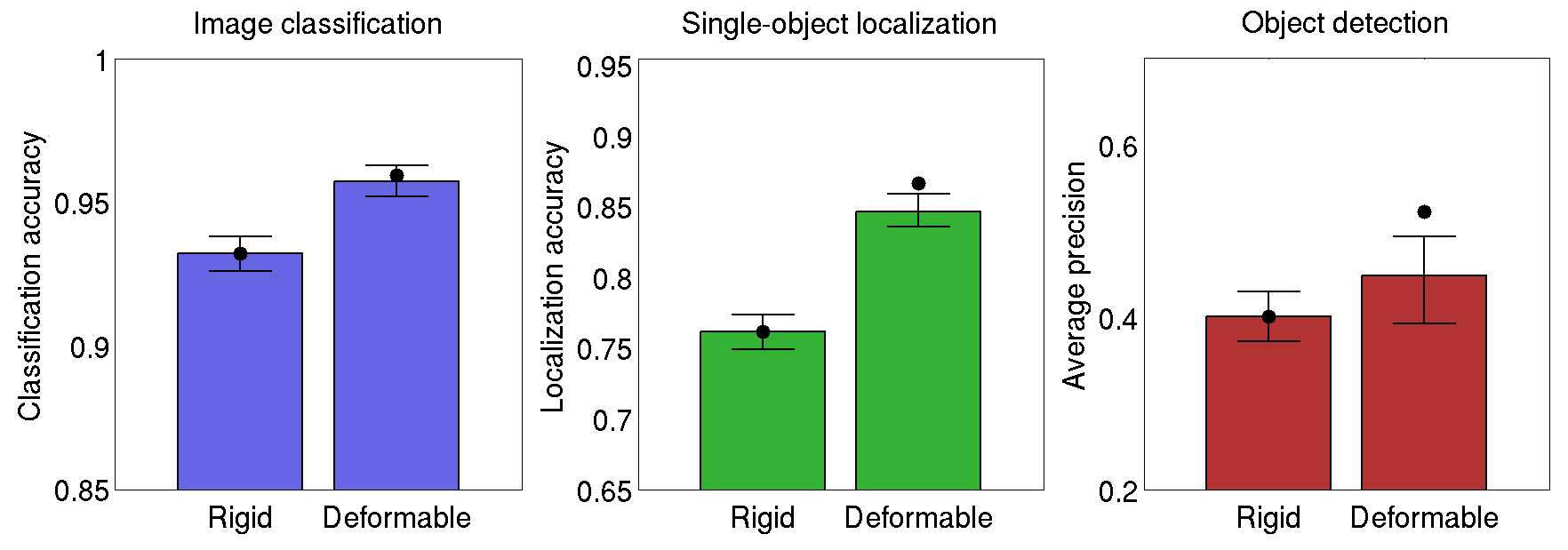} \\
\includegraphics[width=0.57\linewidth]{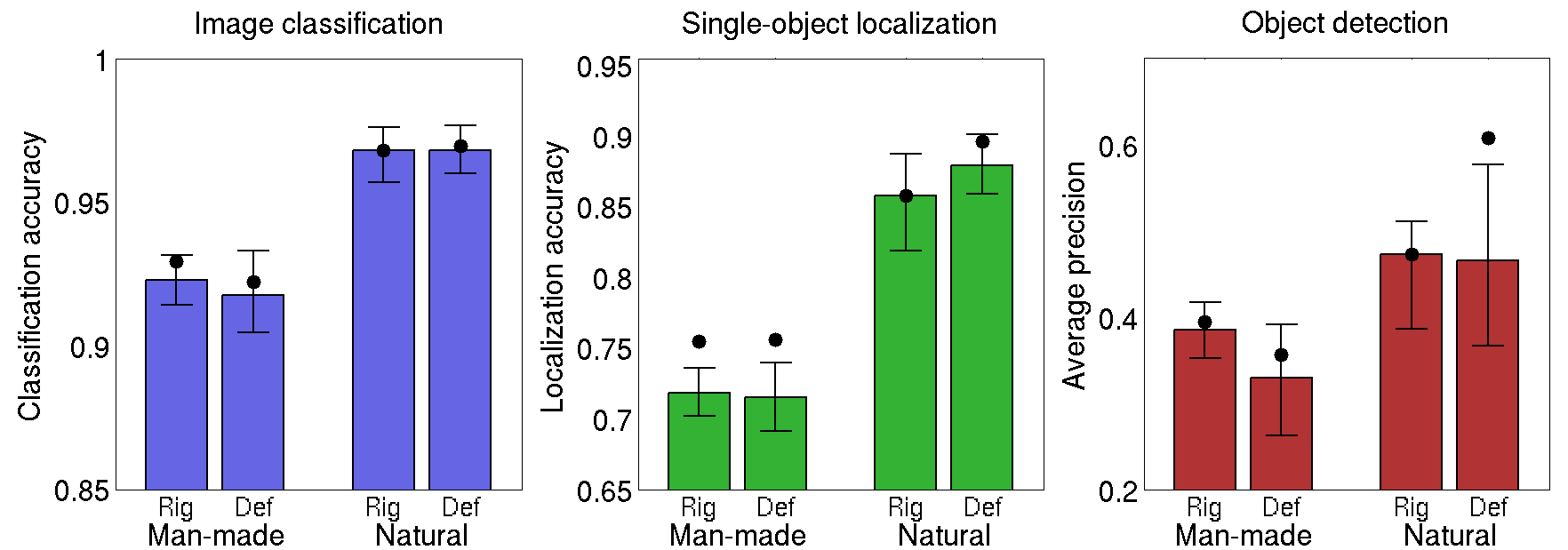} 
\vspace{0.02in}\\
\vspace{0.01in}\\
{\bf Amount of texture}\\
\includegraphics[width=0.57\linewidth]{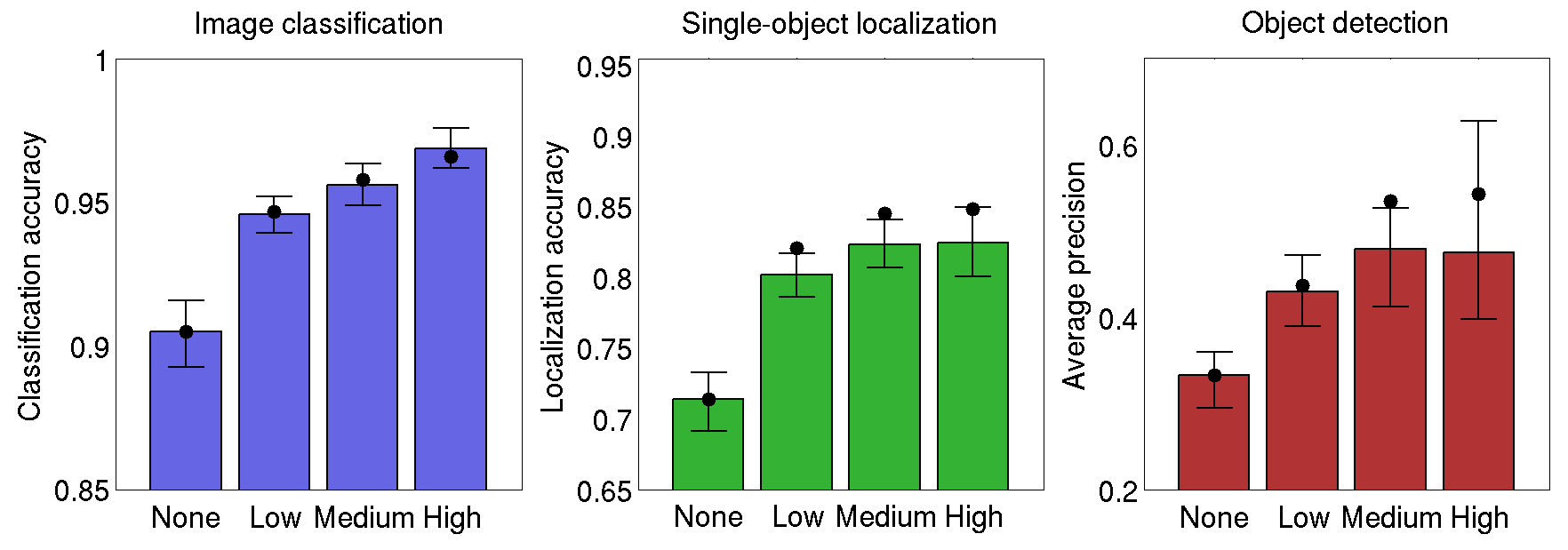} \\
\includegraphics[width=0.57\linewidth]{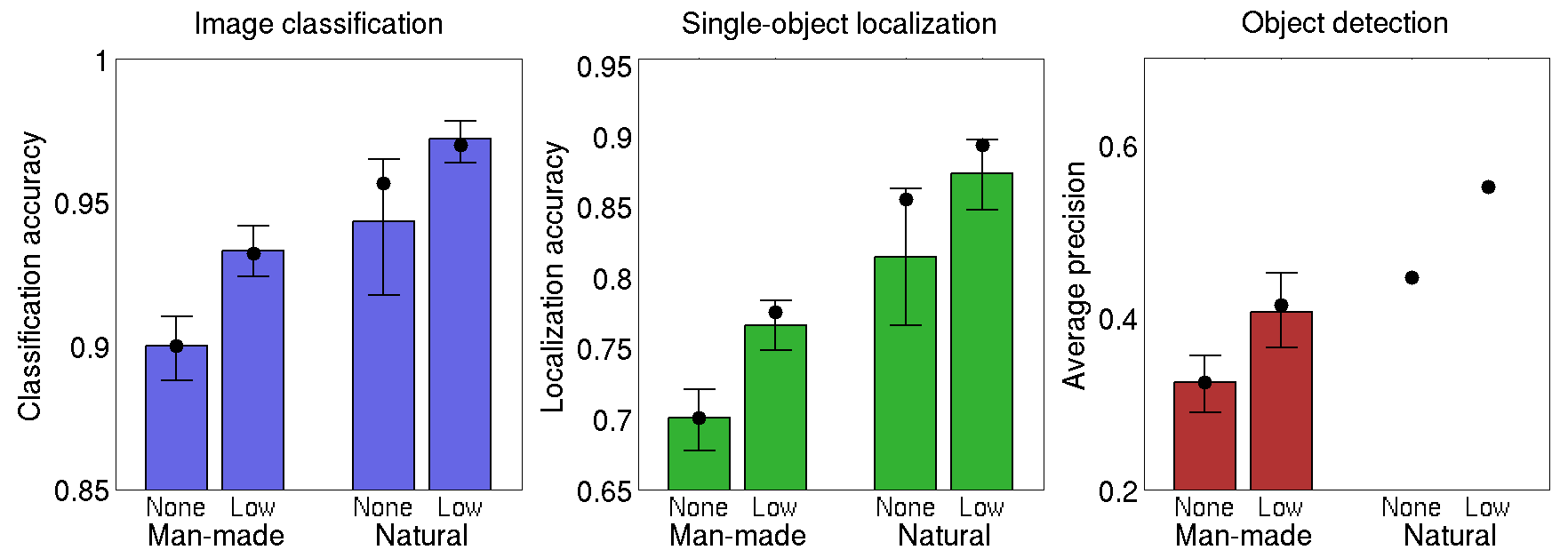} \\
\end{tabular}
\caption{Performance of the ``optimistic'' computer vision model as a function of object properties. The x-axis corresponds to object properties annotated by human labelers for each object class~\citep{Russakovsky13} and illustrated in Figure~\ref{fig:diversity}. The y-axis is the average accuracy of the ``optimistic'' model. Note that the range of the y-axis is different for each task to make the trends more visible. The black circle is the average accuracy of the model on all object classes that fall into each bin. We  control for the effects of object scale by normalizing the object scale within each bin (details in Section~\ref{sec:objectproperties}). The color bars show the model accuracy averaged across the remaining classes. Error bars show the $95\%$ confidence interval obtained with bootstrapping. Some bins are missing color bars because less than 5 object classes remained in the bin after scale normalization. For example, the bar for XL real-world object detection classes is missing because that bin has only 3 object classes (airplane, bus, train) and after normalizing by scale no classes remain. }
\label{fig:props}
\end{figure*}

Figure~\ref{fig:props} shows the average performance of the ``optimistic'' model on the object classes that fall into each bin for each property. We analyze the results in detail below. Unless otherwise specified, the reported accuracies below are after the scale normalization step. 

To evaluate statistical significance, we compute the $95\%$ confidence interval for accuracy using bootstrapping: we repeatedly sample the object classes within the bin with replacement, discard some as needed to normalize by scale, and compute the average accuracy of the ``optimistic'' model on the remaining classes. We report the $95\%$ confidence intervals (CI) in parentheses. 

\paragraph{Real-world size.} In Figure~\ref{fig:props}(top, left) we observe that in the image classification task the ``optimistic'' model tends to perform significantly better on objects which are larger in the real-world.  The classification accuracy is $93.6\%-93.9\%$ on XS, S and M objects compared to $97.0\%$ on L and $96.4\%$ on XL objects. Since this is after normalizing for scale and thus can't be explained by the objects' size in the image, we conclude that either (1) larger real-world objects are easier for the model to recognize, or (2) larger real-world objects usually occur in images with very distinctive backgrounds.

To distinguish between the two cases we look Figure~\ref{fig:props}(top, middle). We see that in the single-object localization task, the L objects are easy to localize at $82.4\%$ localization accuracy. XL objects, however, tend to be the hardest to localize with only $73.4\%$ localization accuracy. We conclude that the appearance of L objects must be easier for the model to learn, while XL objects tend to appear in distinctive backgrounds. The image background make these XL classes easier for the image-level classifier, but the individual instances are difficult to accurately localize. Some examples of L objects are ``killer whale,'' ``schooner,'' and ``lion,'' and some examples of XL objects are ``boathouse,'' ``mosque,'' ``toyshop'' and ``steel arch bridge.''

In Figure~\ref{fig:props}(top,right) corresponding to the object detection task, the influence of real-world object size is not as apparent. One of the key reasons is that many of the XL and L object classes of the image classification and single-object localization datasets were removed in constructing the detection dataset (Section~\ref{sec:AnnotDetObjects}) since they were not basic categories well-suited for detection. There were only 3 XL object classes remaining in the dataset (``train,'' ``airplane'' and ``bus''), and none after scale normalization.We omit them from the analysis. The average precision of XS, S, M objects ($44.5\%$, $39.0\%$, and $38.5\%$ mAP respectively) is statistically insignificant from average precision on L objects: $95\%$ confidence interval of L objects is $37.5\%-59.5\%$. This may be due to the fact that there are only 6 L object classes remaining after scale normalization; all other real-world size bins have at least 18 object classes. 

Finally, it is interesting that performance on XS objects of $44.5\%$ mAP (CI $40.5\%-47.6\%$) is statistically significantly better than performance on S or M objects with $39.0\%$ mAP and $38.5\%$ mAP respectively. Some examples of XS objects are ``strawberry,'' ``bow tie'' and ``rugby ball.''

\paragraph{Deformability within instance.} In Figure~\ref{fig:props}(second row) it is clear that the ``optimistic'' model performs statistically significantly worse on rigid objects than on deformable objects. Image classification accuracy is $93.2\%$ on rigid objects (CI $92.6\%-93.8\%$), much smaller than $95.7\%$ on deformable ones. Single-object localization accuracy is $76.2\%$ on rigid objects (CI $74.9\%-77.4\%$), much smaller than $84.7\%$ on deformable ones. 
Object detection mAP is $40.1\%$ on rigid objects (CI $37.2\%-42.9\%$), much smaller than $44.8\%$ on deformable ones.

%For object detection, before scale normalization deformable objects are much easier than rigid ($52.2\%$ AP for deformable, $40.1\%$ AP for rigid), but after normalizing by scale the two bins %become much more similar at 

We can further analyze the effects of deformability after separating object classes into ``natural'' and ``man-made'' bins based on the ImageNet hierarchy. Deformability is highly correlated with whether the object is natural or man-made: $0.72$ correlation for image classification and single-object localization classes, and $0.61$ for object detection classes. Figure~\ref{fig:props}(third row) shows the effect of deformability on performance of the model for man-made and natural objects separately.

Man-made classes are significantly harder than natural classes: classification accuracy $92.8\%$ (CI $92.3\%-93.3\%$) for man-made versus $97.0\%$ for natural, localization accuracy $75.5\%$ (CI $74.3\%-76.5\%$) for man-made versus $88.5\%$ for natural, and detection mAP $38.7\%$ (CI $35.6-41.3\%$) for man-made versus $50.9\%$ for natural. However, whether the classes are rigid or deformable within this subdivision is no longer significant in most cases. For example, the image classification accuracy is $92.3\%$ (CI $91.4\%-93.1\%$) on man-made rigid objects and $91.8\%$ on man-made deformable objects -- not statistically significantly different. 

There are two cases where the differences in performance \emph{are} statistically significant. First, for single-object localization, natural deformable objects are easier than natural rigid objects: localization accuracy of $87.9\%$ (CI $85.9\%- 90.1\%$) on natural deformable objects is higher than $85.8\%$ on natural rigid objects -- falling slightly outside the $95\%$ confidence interval. This difference in performance is likely because deformable natural animals tend to be easier to localize than rigid natural fruit.

Second, for object detection, man-made rigid objects are easier than man-made deformable objects: $38.5\%$ mAP (CI $35.2\%-41.7\%$) on man-made rigid objects is higher than $33.0\%$ mAP on man-made deformable objects. This is because man-made rigid objects include classes like ``traffic light'' or ``car'' whereas the man-made deformable objects contain challenging classes like ``plastic bag,'' ``swimming trunks'' or ``stethoscope.''

\paragraph{Amount of texture.} Finally, we analyze the effect that object texture has on the accuracy of the ``optimistic'' model. Figure~\ref{fig:props}(fourth row) demonstrates that the model performs better as the amount of texture on the object increases. The most significant difference is between
the performance on untextured objects and the performance on objects with low texture. Image classification accuracy is $90.5\%$ on untextured objects (CI $89.3\%-91.6\%$),  lower than $94.6\%$ on low-textured objects. Single-object localization accuracy is $71.4\%$  on untextured objects (CI $69.1\%-73.3\%$),  lower
than $80.2\%$ on low-textured objects. Object detection mAP is $33.2\%$  on untextured objects (CI $29.5\%-35.9\%$),  lower than $42.9\%$ on low-textured objects.

Texture is correlated with whether the object is natural or man-made, at $0.35$ correlation for image classification and single-object localization, and $0.46$ correlation for object detection. To determine if this is a contributing factor, in Figure~\ref{fig:props}(bottom row) we break up the object classes into natural and man-made and show the accuracy on objects with no texture versus objects with low texture. We observe that the model is still statistically significantly better on low-textured object classes than on untextured ones, both on man-made and natural object classes independently.\footnote{Natural object detection classes are removed from this analysis because there are only 3 and 13 natural untextured and low-textured classes respectively, and none remain after scale normalization. All other bins contain at least 9 object classes after scale normalization.}

\subsection{Human accuracy on large-scale image classification}
\label{sec:ResultsHuman}

Recent improvements in state-of-the-art accuracy on the ILSVRC dataset are easier to put in perspective when compared to human-level accuracy. In this section we compare the performance of the leading large-scale image classification method with the performance of humans on this task.  

To support this comparison, we developed an interface that allowed a human labeler to annotate images with up to five ILSVRC target classes. We compare human errors to those of the winning ILSRC2014 image classification model, GoogLeNet (Section~\ref{sec:MethodsEntries}). For this analysis we use a random sample of 1500 ILSVRC2012-2014 image classification test set images.

% \subsubsection{Annotation Interface.} 
%\subsubsection{Annotation interface}

\paragraph{Annotation interface.} Our web-based annotation interface consists of one test set image and a list of 1000 ILSVRC categories on the side. Each category is described by its title, such as ``cowboy boot.'' The categories are sorted in the topological order of the ImageNet hierarchy, which places semantically similar concepts nearby in the list. For example, all motor vehicle-related classes are arranged contiguously in the list. Every class category is additionally accompanied by a row of 13 examples images from the training set to allow for faster visual scanning. The user of the interface selects 5 categories from the list by clicking on the desired items. Since our interface is web-based, it allows for natural scrolling through the list, and also search by text.

%\subsubsection{Annotation protocol}

\paragraph{Annotation protocol.} We found the task of annotating images with one of 1000 categories to be an extremely challenging task for an untrained annotator. The most common error that an untrained annotator is susceptible to is a failure to consider a relevant class as a possible label because they are unaware of its existence. 

%This proved to be the case even in an {\it assisted} setting, in which we narrowed the list of categories to those predicted by GoogLeNet and categories nearby in the hierarchy.

Therefore, in evaluating the human accuracy we relied primarily on expert annotators who learned to recognize a large portion of the 1000 ILSVRC classes. During training, the annotators labeled a few hundred validation images for practice and later switched to the test set images.

\subsubsection{Quantitative comparison of human and computer accuracy on large-scale image classification}

We report results based on experiments with two expert annotators. The first annotator (A1) trained on 500 images and annotated 1500 test images. The second annotator (A2) trained on 100 images and then annotated 258 test images. The average pace of labeling was approximately 1 image per minute, but the distribution is strongly bimodal: some images are quickly recognized, while some images (such as those of fine-grained breeds of dogs, birds, or monkeys) may require multiple minutes of concentrated effort.

The results are reported in Table \ref{tab:humanresults}. 

\paragraph{Annotator 1.} Annotator A1 evaluated a total of 1500 test set images. The GoogLeNet classification error on this sample was estimated to be $6.8\%$ (recall that the error on full test set of 100,000 images is $6.7\%$, as shown in Table~\ref{table:sub14}). The human error was estimated to be $\textbf{5.1\%}$. Thus, annotator A1 achieves a performance superior to GoogLeNet, by approximately $1.7\%$. We can analyze the statistical significance of this result under the null hypothesis that they are from the same distribution. In particular, comparing the two proportions with a z-test yields a one-sided $p$-value of $p = 0.022$. Thus, we can conclude that this result is statistically significant at the $95\%$ confidence level. 

\paragraph{Annotator 2.} Our second annotator (A2) trained on a smaller sample of only 100 images and then labeled 258 test set images. As seen in Table \ref{tab:humanresults}, the final classification error is significantly worse, at approximately $12.0\%$ Top-5 error. The majority of these errors ($48.8\%$) can be attributed to the annotator failing to spot and consider the ground truth label as an option.

Thus, we conclude that a significant amount of training time is necessary for a human to achieve competitive performance on ILSVRC. However, with a sufficient amount of training, a human annotator is still able to outperform the GoogLeNet result ($p = 0.022$) by approximately $1.7\%$.

\paragraph{Annotator comparison.} We also compare the prediction accuracy of the two annotators. Of a total of 204 images that both A1 and A2 labeled, 174 ($85\%$) were correctly labeled by both A1 and A2, 19 ($9\%$) were correctly labeled by A1 but not A2, 6 ($3\%$) were correctly labeled by A2 but not A1, and 5 ($2\%$) were incorrectly labeled by both. These include 2 images that we consider to be incorrectly labeled in the ground truth.

In particular, our results suggest that the human annotators do not exhibit strong overlap in their predictions. We can approximate the performance of an ``optimistic'' human classifier by assuming an image to be correct if at least one of A1 or A2 correctly labeled the image. On this sample of 204 images, we approximate the error rate of an ``optimistic'' human annotator at $2.4\%$, compared to the GoogLeNet error rate of $4.9\%$.

\begin{table}[t]
\resizebox{1\linewidth}{!} {
\begin{tabular}{|l|r|r|}
  \hline
  \textbf{Relative Confusion} & \textbf{A1} & \textbf{A2}\\
  \hline
  \hline
  Human succeeds, GoogLeNet succeeds & 1352 & 219 \\
  Human succeeds, GoogLeNet fails & 72 & 8 \\
  Human fails, GoogLeNet succeeds & 46 & 24 \\
  Human fails, GoogLeNet fails & 30 & 7 \\
  \hline
  Total number of images & 1500 & 258 \\
  \hline
  Estimated GoogLeNet classification error & $6.8\%$ & $5.8\%$\\
  Estimated human classification error & $5.1\%$ & $12.0\%$\\
  \hline
\end{tabular}
}
\caption{Human classification results on the ILSVRC2012-2014 classification test set, for two expert annotators A1 and A2. We report top-5 classification error.}
\label{tab:humanresults}
\end{table}

\begin{figure*}[t]
    \includegraphics[width=1\textwidth]{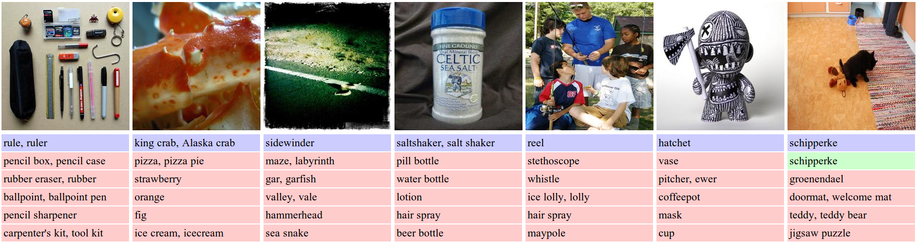}
    \caption{Representative validation images that highlight common sources of error. For each image, we display the ground truth in blue, and top 5 predictions from GoogLeNet follow (red = wrong, green = right). GoogLeNet predictions on the validation set images were graciously provided by members of the GoogLeNet team. From left to right: Images that contain multiple objects, images of extreme closeups and uncharacteristic views, images with filters, images that significantly benefit from the ability to read text, images that contain very small and thin objects, images with abstract representations, and example of a fine-grained image that GoogLeNet correctly identifies but a human would have significant difficulty with.}
    \label{fig:example_errors}
\end{figure*}

\subsubsection{Analysis of human and computer errors on large-scale image classification} 

We manually inspected both human and GoogLeNet errors to gain an understanding of common error types and how they compare. For purposes of this section, we only discuss results based on the larger sample of 1500 images that were labeled by annotator A1. Examples of representative mistakes are in Figure \ref{fig:example_errors}. The analysis and insights below were derived specifically from GoogLeNet predictions, but we suspect that many of the same errors may be present in other methods.

\paragraph{Types of errors in both computer and human annotations:}

\begin{enumerate}
\item \textbf{Multiple objects.} Both GoogLeNet and humans struggle with images that contain multiple ILSVRC classes (usually many more than five), with little indication of which object is the focus of the image. This error is only present in the Classification setting, since every image is constrained to have exactly one correct label. In total, we attribute 24 ($24\%$) of GoogLeNet errors and 12 ($16\%$) of human errors to this category. It is worth noting that humans can have a slight advantage in this error type, since it can sometimes be easy to identify the most salient object in the image.

\item \textbf{Incorrect annotations.} We found that approximately 5 out of 1500 images ($0.3\%$) were incorrectly annotated in the ground truth. This introduces an approximately equal number of errors for both humans and GoogLeNet.
\end{enumerate}

\paragraph{Types of errors that the computer is more susceptible to than the human:}

\begin{enumerate}
\item \textbf{Object small or thin.} GoogLeNet struggles with recognizing objects that are very small or thin in the image, even if that object is the only object present. Examples of this include an image of a standing person wearing sunglasses, a person holding a quill in their hand, or a small ant on a stem of a flower. We estimate that approximately 22 ($21\%$) of GoogLeNet errors fall into this category, while none of the human errors do. In other words, in our sample of images, no image was mislabeled by a human because they were unable to identify a very small or thin object. This discrepancy can be attributed to the fact that a human can very effectively leverage context and affordances to accurately infer the identity of small objects (for example, a few barely visible feathers near person's hand as very likely belonging to a mostly occluded quill).

\item \textbf{Image filters.} Many people enhance their photos with filters that distort the contrast and color distributions of the image. We found that 13 ($13\%$) of the images that GoogLeNet incorrectly classified contained a filter. Thus, we posit that GoogLeNet is not very robust to these distortions. In comparison, only one image among the human errors contained a filter, but we do not attribute the source of the error to the filter.

\item \textbf{Abstract representations}. GoogLeNet struggles with images that depict objects of interest in an abstract form, such as 3D-rendered images, paintings, sketches, plush toys, or statues. An example is the abstract shape of a bow drawn with a light source in night photography, a 3D-rendered robotic scorpion, or a shadow on the ground, of a child on a swing. We attribute approximately 6 ($6\%$) of GoogLeNet errors to this type of error and believe that humans are significantly more robust, with no such errors seen in our sample.

\item \textbf{Miscellaneous sources.} Additional sources of error that occur relatively infrequently include extreme closeups of parts of an object, unconventional viewpoints such as a rotated image, images that can significantly benefit from the ability to read text (e.g. a featureless container identifying itself as ``face powder''), objects with heavy occlusions, and images that depict a collage of multiple images. In general, we found that humans are more robust to all of these types of error.
\end{enumerate}

\paragraph{Types of errors that the human is more susceptible to than the computer:}

\begin{enumerate}

\item \textbf{Fine-grained recognition.} We found that humans are noticeably worse at fine-grained recognition (e.g. dogs, monkeys, snakes, birds), even when they are in clear view. To understand the difficulty, consider that there are more than 120 species of dogs in the dataset. We estimate that 28 ($37\%$) of the human errors fall into this category, while only 7 ($7\%$) of GoogLeNet errors do.

\item \textbf{Class unawareness.} The annotator may sometimes be unaware of the ground truth class present as a label option. When pointed out as an ILSVRC class, it is usually clear that the label applies to the image. These errors get progressively less frequent as the annotator becomes more familiar with ILSVRC classes. Approximately 18 ($24\%$) of the human errors fall into this category.

\item \textbf{Insufficient training data.} Recall that the annotator is only presented with 13 examples of a class under every category name. However, 13 images are not always enough to adequately convey the allowed class variations. For example, a brown dog can be incorrectly dismissed as a ``Kelpie'' if all examples of a ``Kelpie'' feature a dog with black coat. However, if more than 13 images were listed it would have become clear that a ``Kelpie'' may have brown coat. Approximately 4 ($5\%$) of human errors fall into this category.

\end{enumerate}

\subsubsection{Conclusions from human image classification experiments} 

We investigated the performance of trained human annotators on a sample of 1500 ILSVRC test set images. Our results indicate that a trained human annotator is capable of outperforming the best model (GoogLeNet) by approximately $1.7\%$ ($p = 0.022$). 

We expect that some sources of error may be relatively easily eliminated (e.g. robustness to filters, rotations, collages, effectively reasoning over multiple scales), while others may prove more elusive (e.g. identifying abstract representations of objects). On the other hand, a large majority of human errors come from fine-grained categories and class unawareness. We expect that the former can be significantly reduced with fine-grained expert annotators, while the latter could be reduced with more practice and greater familiarity with ILSVRC classes. Our results also hint that human errors are not strongly correlated and that human ensembles may further reduce human error rate.

It is clear that humans will soon outperform state-of-the-art ILSVRC image classification models only by use of significant effort, expertise, and time. One interesting follow-up question for future investigation is how computer-level accuracy compares with human-level accuracy on more complex image understanding tasks.

\section{Conclusions}
\label{sec:Conclusion}

In this paper we described the large-scale data collection process of ILSVRC, provided a summary of the most successful algorithms on this data, and analyzed the success and failure modes of these algorithms. In this section we discuss some of the key lessons we learned over the years of ILSVRC, strive to address the key criticisms of the datasets and the challenges we encountered over the years, and conclude by looking forward into the future.

\subsection{Lessons learned}

The key lesson of collecting the datasets and running the challenges for five years is this: {\bf All human intelligence tasks need to be exceptionally well-designed.} We learned this lesson both when annotating the dataset using Amazon Mechanical Turk workers (Section~\ref{sec:Annot}) and even
when trying to evaluate human-level image classification accuracy using expert labelers (Section~\ref{sec:ResultsHuman}).  The first iteration of the labeling interface was always bad -- generally meaning \emph{completely unusable}. If there was any inherent ambiguity in the questions posed (and there almost always was), workers found it and accuracy suffered. If there is one piece of advice we can offer to future research, it is to very carefully design, continuously monitor, and extensively sanity-check all crowdsourcing tasks. 

The other lesson, already well-known to large-scale researchers, is this: {\bf Scaling up the dataset always reveals unexpected challenges}. From designing complicated multi-step annotation strategies (Section~\ref{sec:AnnotLocBbox}) to having to modify the evaluation procedure (Section~\ref{sec:Evaluation}), we had to continuously adjust to the large-scale setting. On the plus side, of course, the major breakthroughs in object recognition accuracy (Section~\ref{sec:Methods}) and the analysis of the strength and weaknesses of current algorithms as a function of object class properties ( Section~\ref{sec:ResultsICCV}) would never have been possible on a smaller scale.

\subsection{Criticism}

In the past five years, we encountered three major criticisms of the ILSVRC dataset and the corresponding challenge: (1) the ILSVRC dataset is insufficiently challenging, (2) the ILSVRC dataset contains annotation errors, and (3) the rules of ILSVRC competition are too restrictive. We discuss these in order.

The first criticism is that the objects in the dataset tend to be large and centered in the images, making the dataset insufficiently challenging. In Sections~\ref{sec:LocStats} and \ref{sec:AnnotDetStats}  we tried to put those concerns to rest by analyzing the statistics of the ILSVRC dataset and concluding that it is comparable with, and in many cases much more challenging than, the long-standing PASCAL VOC benchmark~\citep{PASCALIJCV}.

The second is regarding the errors in ground truth labeling. We went through several rounds of in-house post-processing of the annotations obtained using crowdsourcing, and corrected many common sources of errors (e.g., Appendix~\ref{sec:AppDetBbox}). The major remaining source of annotation errors stem from fine-grained object classes, e.g., labelers failing to distinguish different species of birds. This is a tradeoff that had to be made: in order to annotate data at this scale on a reasonable budget, we had to rely on non-expert crowd labelers. However, overall the dataset is encouragingly clean. By our estimates, $99.7\%$ precision is achieved in the image classification dataset (Sections~\ref{sec:AnnotClsAnnot} and~\ref{sec:ResultsHuman}) and $97.9\%$ of images that went through the bounding box annotation system have all instances of the target object class labeled with bounding boxes (Section~\ref{sec:AnnotLocBbox}).

 The third criticism we encountered is over the rules of the competition regarding using external training data. In ILSVRC2010-2013, algorithms had to only use the provided training and validation set images and annotations for training their models. With the growth of the field of large-scale unsupervised feature learning, however, questions began to arise about what exactly constitutes ``outside'' data: for example, are  image features trained on a large pool of ``outside'' images in an unsupervised fashion allowed in the competition? After much discussion, in ILSVRC2014 we took the first step towards addressing this problem. We followed the PASCAL VOC strategy and created two tracks in the competition: entries using only ``provided'' data and entries using ``outside'' data, meaning \emph{any} images or annotations not provided as part of ILSVRC training or validation sets. However, in the future this strategy will likely need to be further revised as the computer vision field evolves. For example, competitions can consider allowing the use of any image features which are publically available, even if these features were learned on an external source of data.

%Many traditional computer vision features, like SIFT~\citep{Lowe04}, were tuned to perform well on other image classification datasets. So by the same reasoning, one can %argue that. 

\subsection{The future}

Given the massive algorithmic breakthroughs over the past five years, we are very eager to see what will happen in the next five years. There are many potential directions of improvement and growth for ILSVRC and other large-scale image datasets.

First, continuing the trend of moving towards richer image understanding (from image classification to single-object localization to object detection), the next challenge would be to tackle pixel-level object segmentation. The recently released large-scale COCO dataset~\citep{COCO} is already taking a step in that direction. 

Second, as datasets grow  even larger in scale, it may become impossible to fully annotate them manually. The 
scale of ILSVRC is already imposing limits on the manual annotations that are feasible to obtain: for example, we had to restrict the number of objects labeled per image in the image classification and single-object localization datasets. In the future, with billions of images, it will become impossible to obtain even one clean label for every image.  Datasets such as Yahoo's Flickr Creative Commons 100M,\footnote{\url{http://webscope.sandbox.yahoo.com/catalog.php?datatype=i&did=67}} released with weak human tags but no centralized annotation, will become more common.

The growth of unlabeled or only partially labeled large-scale datasets implies two things. First, algorithms will have to rely more on weakly supervised training data. Second, even evaluation might have to be done \emph{after} the algorithms make predictions, not before. This means that rather than evaluating \emph{accuracy} (how many of the test images or objects did the algorithm get right) or \emph{recall} (how many of the desired images or objects did the algorithm manage to find), both of which require a fully annotated test set, we will be focusing more on \emph{precision}: of the predictions that the algorithm made, how many were deemed correct by humans. 

We are eagerly awaiting the future development of object recognition datasets and algorithms, and are grateful that ILSVRC served as a stepping stone along this path.

\begin{acknowledgements}
We thank Stanford University, UNC Chapel Hill, Google and Facebook for sponsoring the challenges, and NVIDIA for providing computational resources to participants of ILSVRC2014. 
We thank our advisors over the years: Lubomir Bourdev, Alexei Efros, Derek Hoiem, Jitendra Malik, Chuck Rosenberg and Andrew Zisserman. We thank the PASCAL VOC organizers for partnering with us in running ILSVRC2010-2012.
We thank all members of the Stanford vision lab for supporting the challenges and putting up with us along the way. Finally, and most importantly, we thank all researchers that have made the ILSVRC effort a success by competing in the challenges and by using the datasets to advance  computer vision. 
\end{acknowledgements}

\begin{appendices}

%%%%%%%%%%%%%%%%%%%%%%%%%%%%%%%%%%%%%%%%%%%%%%%%%%%%%%%%%%%%%%%%%%%%%%%%%%%%%%%%%%

%\section{Selecting object categories}
%\label{sec:AppClasses}

\section{ILSVRC2012-2014 image classification and single-object localization object categories}
\label{app:classes}

{\tiny
%{\bf Image classification and single-object localization (1000 categories)}:
abacus, abaya, academic gown, accordion, acorn, acorn squash, acoustic guitar, admiral, affenpinscher, Afghan hound, African chameleon, African crocodile, African elephant, African grey, African hunting dog, agama, agaric, aircraft carrier, Airedale, airliner, airship, albatross, alligator lizard, alp, altar, ambulance, American alligator, American black bear, American chameleon, American coot, American egret, American lobster, American Staffordshire terrier, amphibian, analog clock, anemone fish, Angora, ant, apiary, Appenzeller, apron, Arabian camel, Arctic fox, armadillo, artichoke, ashcan, assault rifle, Australian terrier, axolotl, baboon, backpack, badger, bagel, bakery, balance beam, bald eagle, balloon, ballplayer, ballpoint, banana, Band Aid, banded gecko, banjo, bannister, barbell, barber chair, barbershop, barn, barn spider, barometer, barracouta, barrel, barrow, baseball, basenji, basketball, basset, bassinet, bassoon, bath towel, bathing cap, bathtub, beach wagon, beacon, beagle, beaker, bearskin, beaver, Bedlington terrier, bee, bee eater, beer bottle, beer glass, bell cote, bell pepper, Bernese mountain dog, bib, bicycle-built-for-two, bighorn, bikini, binder, binoculars, birdhouse, bison, bittern, black and gold garden spider, black grouse, black stork, black swan, black widow, black-and-tan coonhound, black-footed ferret, Blenheim spaniel, bloodhound, bluetick, boa constrictor, boathouse, bobsled, bolete, bolo tie, bonnet, book jacket, bookcase, bookshop, Border collie, Border terrier, borzoi, Boston bull, bottlecap, Bouvier des Flandres, bow, bow tie, box turtle, boxer, Brabancon griffon, brain coral, brambling, brass, brassiere, breakwater, breastplate, briard, Brittany spaniel, broccoli, broom, brown bear, bubble, bucket, buckeye, buckle, bulbul, bull mastiff, bullet train, bulletproof vest, bullfrog, burrito, bustard, butcher shop, butternut squash, cab, cabbage butterfly, cairn, caldron, can opener, candle, cannon, canoe, capuchin, car mirror, car wheel, carbonara, Cardigan, cardigan, cardoon, carousel, carpenter's kit, carton, cash machine, cassette, cassette player, castle, catamaran, cauliflower, CD player, cello, cellular telephone, centipede, chain, chain mail, chain saw, chainlink fence, chambered nautilus, cheeseburger, cheetah, Chesapeake Bay retriever, chest, chickadee, chiffonier, Chihuahua, chime, chimpanzee, china cabinet, chiton, chocolate sauce, chow, Christmas stocking, church, cicada, cinema, cleaver, cliff, cliff dwelling, cloak, clog, clumber, cock, cocker spaniel, cockroach, cocktail shaker, coffee mug, coffeepot, coho, coil, collie, colobus, combination lock, comic book, common iguana, common newt, computer keyboard, conch, confectionery, consomme, container ship, convertible, coral fungus, coral reef, corkscrew, corn, cornet, coucal, cougar, cowboy boot, cowboy hat, coyote, cradle, crane, crane, crash helmet, crate, crayfish, crib, cricket, Crock Pot, croquet ball, crossword puzzle, crutch, cucumber, cuirass, cup, curly-coated retriever, custard apple, daisy, dalmatian, dam, damselfly, Dandie Dinmont, desk, desktop computer, dhole, dial telephone, diamondback, diaper, digital clock, digital watch, dingo, dining table, dishrag, dishwasher, disk brake, Doberman, dock, dogsled, dome, doormat, dough, dowitcher, dragonfly, drake, drilling platform, drum, drumstick, dugong, dumbbell, dung beetle, Dungeness crab, Dutch oven, ear, earthstar, echidna, eel, eft, eggnog, Egyptian cat, electric fan, electric guitar, electric locomotive, electric ray, English foxhound, English setter, English springer, entertainment center, EntleBucher, envelope, Eskimo dog, espresso, espresso maker, European fire salamander, European gallinule, face powder, feather boa, fiddler crab, fig, file, fire engine, fire screen, fireboat, flagpole, flamingo, flat-coated retriever, flatworm, flute, fly, folding chair, football helmet, forklift, fountain, fountain pen, four-poster, fox squirrel, freight car, French bulldog, French horn, French loaf, frilled lizard, frying pan, fur coat, gar, garbage truck, garden spider, garter snake, gas pump, gasmask, gazelle, German shepherd, German short-haired pointer, geyser, giant panda, giant schnauzer, gibbon, Gila monster, go-kart, goblet, golden retriever, goldfinch, goldfish, golf ball, golfcart, gondola, gong, goose, Gordon setter, gorilla, gown, grand piano, Granny Smith, grasshopper, Great Dane, great grey owl, Great Pyrenees, great white shark, Greater Swiss Mountain dog, green lizard, green mamba, green snake, greenhouse, grey fox, grey whale, grille, grocery store, groenendael, groom, ground beetle, guacamole, guenon, guillotine, guinea pig, gyromitra, hair slide, hair spray, half track, hammer, hammerhead, hamper, hamster, hand blower, hand-held computer, handkerchief, hard disc, hare, harmonica, harp, hartebeest, harvester, harvestman, hatchet, hay, head cabbage, hen, hen-of-the-woods, hermit crab, hip, hippopotamus, hog, hognose snake, holster, home theater, honeycomb, hook, hoopskirt, horizontal bar, hornbill, horned viper, horse cart, hot pot, hotdog, hourglass, house finch, howler monkey, hummingbird, hyena, ibex, Ibizan hound, ice bear, ice cream, ice lolly, impala, Indian cobra, Indian elephant, indigo bunting, indri, iPod, Irish setter, Irish terrier, Irish water spaniel, Irish wolfhound, iron, isopod, Italian greyhound, jacamar, jack-o'-lantern, jackfruit, jaguar, Japanese spaniel, jay, jean, jeep, jellyfish, jersey, jigsaw puzzle, jinrikisha, joystick, junco, keeshond, kelpie, Kerry blue terrier, killer whale, kimono, king crab, king penguin, king snake, kit fox, kite, knee pad, knot, koala, Komodo dragon, komondor, kuvasz, lab coat, Labrador retriever, lacewing, ladle, ladybug, Lakeland terrier, lakeside, lampshade, langur, laptop, lawn mower, leaf beetle, leafhopper, leatherback turtle, lemon, lens cap, Leonberg, leopard, lesser panda, letter opener, Lhasa, library, lifeboat, lighter, limousine, limpkin, liner, lion, lionfish, lipstick, little blue heron, llama, Loafer, loggerhead, long-horned beetle, lorikeet, lotion, loudspeaker, loupe, lumbermill, lycaenid, lynx, macaque, macaw, Madagascar cat, magnetic compass, magpie, mailbag, mailbox, maillot, maillot, malamute, malinois, Maltese dog, manhole cover, mantis, maraca, marimba, marmoset, marmot, mashed potato, mask, matchstick, maypole, maze, measuring cup, meat loaf, medicine chest, meerkat, megalith, menu, Mexican hairless, microphone, microwave, military uniform, milk can, miniature pinscher, miniature poodle, miniature schnauzer, minibus, miniskirt, minivan, mink, missile, mitten, mixing bowl, mobile home, Model T, modem, monarch, monastery, mongoose, monitor, moped, mortar, mortarboard, mosque, mosquito net, motor scooter, mountain bike, mountain tent, mouse, mousetrap, moving van, mud turtle, mushroom, muzzle, nail, neck brace, necklace, nematode, Newfoundland, night snake, nipple, Norfolk terrier, Norwegian elkhound, Norwich terrier, notebook, obelisk, oboe, ocarina, odometer, oil filter, Old English sheepdog, orange, orangutan, organ, oscilloscope, ostrich, otter, otterhound, overskirt, ox, oxcart, oxygen mask, oystercatcher, packet, paddle, paddlewheel, padlock, paintbrush, pajama, palace, panpipe, paper towel, papillon, parachute, parallel bars, park bench, parking meter, partridge, passenger car, patas, patio, pay-phone, peacock, pedestal, Pekinese, pelican, Pembroke, pencil box, pencil sharpener, perfume, Persian cat, Petri dish, photocopier, pick, pickelhaube, picket fence, pickup, pier, piggy bank, pill bottle, pillow, pineapple, ping-pong ball, pinwheel, pirate, pitcher, pizza, plane, planetarium, plastic bag, plate, plate rack, platypus, plow, plunger, Polaroid camera, pole, polecat, police van, pomegranate, Pomeranian, poncho, pool table, pop bottle, porcupine, pot, potpie, potter's wheel, power drill, prairie chicken, prayer rug, pretzel, printer, prison, proboscis monkey, projectile, projector, promontory, ptarmigan, puck, puffer, pug, punching bag, purse, quail, quill, quilt, racer, racket, radiator, radio, radio telescope, rain barrel, ram, rapeseed, recreational vehicle, red fox, red wine, red wolf, red-backed sandpiper, red-breasted merganser, redbone, redshank, reel, reflex camera, refrigerator, remote control, restaurant, revolver, rhinoceros beetle, Rhodesian ridgeback, rifle, ringlet, ringneck snake, robin, rock beauty, rock crab, rock python, rocking chair, rotisserie, Rottweiler, rubber eraser, ruddy turnstone, ruffed grouse, rugby ball, rule, running shoe, safe, safety pin, Saint Bernard, saltshaker, Saluki, Samoyed, sandal, sandbar, sarong, sax, scabbard, scale, schipperke, school bus, schooner, scoreboard, scorpion, Scotch terrier, Scottish deerhound, screen, screw, screwdriver, scuba diver, sea anemone, sea cucumber, sea lion, sea slug, sea snake, sea urchin, Sealyham terrier, seashore, seat belt, sewing machine, Shetland sheepdog, shield, Shih-Tzu, shoe shop, shoji, shopping basket, shopping cart, shovel, shower cap, shower curtain, siamang, Siamese cat, Siberian husky, sidewinder, silky terrier, ski, ski mask, skunk, sleeping bag, slide rule, sliding door, slot, sloth bear, slug, snail, snorkel, snow leopard, snowmobile, snowplow, soap dispenser, soccer ball, sock, soft-coated wheaten terrier, solar dish, sombrero, sorrel, soup bowl, space bar, space heater, space shuttle, spaghetti squash, spatula, speedboat, spider monkey, spider web, spindle, spiny lobster, spoonbill, sports car, spotlight, spotted salamander, squirrel monkey, Staffordshire bullterrier, stage, standard poodle, standard schnauzer, starfish, steam locomotive, steel arch bridge, steel drum, stethoscope, stingray, stinkhorn, stole, stone wall, stopwatch, stove, strainer, strawberry, street sign, streetcar, stretcher, studio couch, stupa, sturgeon, submarine, suit, sulphur butterfly, sulphur-crested cockatoo, sundial, sunglass, sunglasses, sunscreen, suspension bridge, Sussex spaniel, swab, sweatshirt, swimming trunks, swing, switch, syringe, tabby, table lamp, tailed frog, tank, tape player, tarantula, teapot, teddy, television, tench, tennis ball, terrapin, thatch, theater curtain, thimble, three-toed sloth, thresher, throne, thunder snake, Tibetan mastiff, Tibetan terrier, tick, tiger, tiger beetle, tiger cat, tiger shark, tile roof, timber wolf, titi, toaster, tobacco shop, toilet seat, toilet tissue, torch, totem pole, toucan, tow truck, toy poodle, toy terrier, toyshop, tractor, traffic light, trailer truck, tray, tree frog, trench coat, triceratops, tricycle, trifle, trilobite, trimaran, tripod, triumphal arch, trolleybus, trombone, tub, turnstile, tusker, typewriter keyboard, umbrella, unicycle, upright, vacuum, valley, vase, vault, velvet, vending machine, vestment, viaduct, vine snake, violin, vizsla, volcano, volleyball, vulture, waffle iron, Walker hound, walking stick, wall clock, wallaby, wallet, wardrobe, warplane, warthog, washbasin, washer, water bottle, water buffalo, water jug, water ouzel, water snake, water tower, weasel, web site, weevil, Weimaraner, Welsh springer spaniel, West Highland white terrier, whippet, whiptail, whiskey jug, whistle, white stork, white wolf, wig, wild boar, window screen, window shade, Windsor tie, wine bottle, wing, wire-haired fox terrier, wok, wolf spider, wombat, wood rabbit, wooden spoon, wool, worm fence, wreck, yawl, yellow lady's slipper, Yorkshire terrier, yurt, zebra, zucchini

}

%\noindent
%{\bf Object detection (200 categories)}:
%\input{det_hierarchy.tex}
%}

%%%%%%%%%%%%%%%%%%%%%%%%%%%%%%%%%%%%%%%%%%%%%%%%%%%%%%%%%%%%%%%%%%%%%%%%%%%%%%%%%%

\section{Additional single-object localization dataset statistics}
\label{app:ICCV}

\begin{figure*}
\center{{\bf 1000 classes of ILSVRC2012-2014 single-object localization (dark green) \\ versus 20 classes of PASCAL 2012 (light blue)}}
\begin{tabular}{p{0.22\textwidth} p{0.22\textwidth} p{0.22\textwidth} p{0.22\textwidth}}
\center{$\rightarrow$ more diffiult} & 
\center{more diffiult $\leftarrow$}  &
\center{more diffiult $\leftarrow$}  &
\center{ $\rightarrow$ more diffiult}
\end{tabular}
\\
\includegraphics[width=\textwidth]{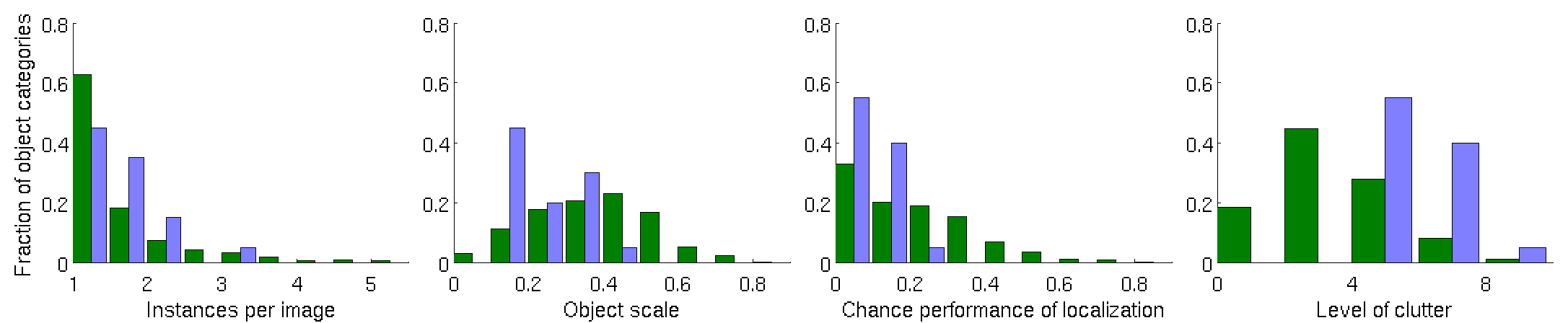} \\
\center{{\bf 200 hardest classes of ILSVRC2012-2014 single-object localization (dark green) \\ versus 20 classes of PASCAL 2012  (light blue)}}
\begin{tabular}{p{0.22\textwidth} p{0.22\textwidth} p{0.22\textwidth} p{0.22\textwidth}}
\center{$\rightarrow$ more diffiult} & 
\center{more diffiult $\leftarrow$}  &
\center{more diffiult $\leftarrow$}  &
\center{ $\rightarrow$ more diffiult}
\end{tabular}
\\
\includegraphics[width=\textwidth]{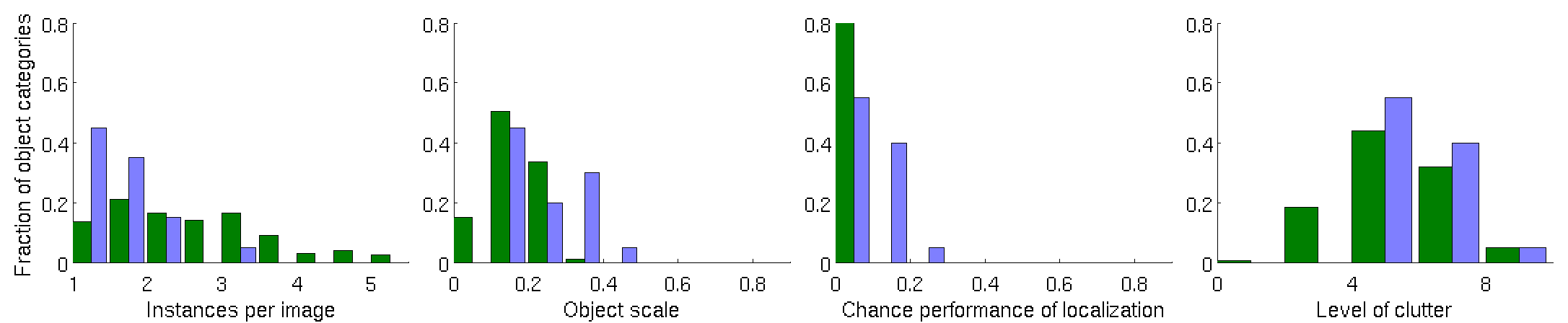} \\
\caption{Distribution of various measures of localization difficulty on the ILSVRC2012-2014 single-object localization (dark green) and PASCAL VOC 2012 (light blue) validation sets. Object scale is fraction of image area occupied by an average object instance. Chance performance of localization and level of clutter are defined in Appendix~\ref{app:ICCV}. The plots on top contain the full ILSVRC validation set with 1000 classes; the plots on the bottom contain 200 ILSVRC classes with the lowest chance performance of localization. All plots contain all 20 classes of PASCAL VOC. 
}
\label{fig:locdiff}
\end{figure*}

We consider two additional metrics of object localization difficulty: chance performance of localization and the level of clutter. We use these metrics to compare ILSVRC2012-2014
single-object localization dataset to the PASCAL VOC 2012 object detection benchmark. The measures of localization difficulty are computed on the validation set of both datasets. According to both of these measures of difficulty there is a subset of ILSVRC which is as challenging as PASCAL but more than an order of magnitude greater in size. Figure~\ref{fig:locdiff} shows the distributions of different properties (object scale, chance performance of localization and level of clutter) across the different classes in the two datasets.

\iffalse
\begin{table}
{\small
\begin{tabular}{|c|c|c|c|}
\hline
Property & PASCAL & ILSVRC subset & ILSVRC \\
%\hline
%SCALE & 24.08\%  & 24.06\% (537) & 35.83\% \\
\hline
Instances & 1.69 (20) & 1.69 (843)  & 1.61 (1000)\\
\hline
%Neighbors & 0.52 (20) & 0.52 (913) & 0.47 (1000)\\
%\hline
CPL & 8.76\%  (20) & 8.74\% (562) & 20.83\% (1000) \\
\hline
Clutter & 5.90  (20) & 5.90 (250) & 3.59 (1000)\\
\hline
\end{tabular}
}
\caption{Comparison of the PASCAL VOC 2012 object detection dataset and the ILSVRC 2012 single-object localization dataset on several key properties of object localization difficulty (please see Section~\ref{sec:LocStats} for definitions). For each property, the ILSVRC subset is chosen by selecting the largest set of ILSVRC classes with the same average difficulty as that of the PASCAL classes. The number of object classes is indicated in parentheses.  Note that according to any of these measures of difficulty there is a subset of ILSVRC which is as challenging as PASCAL but more than an order of magnitude greater in size.}
\label{table:impascal}
\end{table}
\fi

\paragraph{Chance performance of localization (CPL).} Chance performance on a dataset is a common metric to consider. We define the CPL measure as the expected accuracy of a detector which first randomly samples an object instance of that class and then uses its bounding box directly as the proposed localization window on all other images (after rescaling the images to the same size). Concretely, let $B_1,B_2,\dots,B_N$ be all the bounding boxes of the object instances within a class, then
\begin{equation}
\label{cpleq}
\mbox{\sc CPL} = \frac{\sum_i \sum_{j \neq i} IOU(B_i,B_j)\geq 0.5}{N(N-1)}
\end{equation}
Some of the most difficult ILSVRC categories to localize according to this metric are basketball, swimming trunks, ping pong ball and rubber eraser, all with less than $0.2\%$ CPL. This measure correlates strongly ($\rho = 0.9$) with the average scale of the object (fraction of image occupied by object). The average CPL across the $1000$ ILSVRC categories is $20.8\%$. The 20 PASCAL categories have an average CPL of $8.7\%$, which is the same as the CPL of the $562$ most difficult categories of ILSVRC.

\paragraph{Clutter.} Intuitively, even small objects are easy to localize on a plain background. To quantify clutter we employ the objectness measure of~\citep{Alexe12}, which is a class-generic object detector evaluating how likely a window in the image contains a coherent object (of any class) as opposed to background (sky, water, grass). For every image $m$ containing target object instances at positions $B_1^m,B_2^m,\dots$, we use the publicly available objectness software to sample 1000 windows $W_1^m,W_2^m,\dots W_{1000}^m$, in order of decreasing probability of the window containing any generic object. Let {\sc obj}(m) be the number of generic object-looking windows sampled before localizing an instance of the target category, i.e., $\mbox{\sc obj}(m) = \min \{k: \max_i \mbox{\sc iou}(W_k^m,B_i^m) \geq 0.5\}$. For a category containing M images, we compute the average number of such windows per image and define
\begin{equation}
\label{cluttereq}
\textstyle
\mbox{\sc Clutter} = \log_{2} \large ( \frac{1}{M}\sum_m \mbox{\sc obj}(m) \large )
\end{equation}
The higher the clutter of a category, the harder the objects are to localize according to generic cues. If an object can't be localized with the first 1000 windows (as is the case for $1\%$ of images on average per category in ILSVRC and $5\%$ in PASCAL), we set {\sc obj}$(m)=1001$. The fact that more than $95\%$ of objects can be localized with these windows imply that the objectness cue is already quite strong, so objects that require many windows on average will be extremely difficult to detect: e.g., ping pong ball (clutter of 9.57, or 758 windows on average), basketball (clutter of 9.21), puck (clutter of 9.17) in ILSVRC. The most difficult object in PASCAL is bottle with clutter score of $8.47$.  On average, ILSVRC has clutter score of $3.59$. The most difficult subset of ILSVRC with 250 object categories has an order of magnitude more categories and the same average amount of clutter (of $5.90$) as the PASCAL dataset. 

%%%%%%%%%%%%%%%%%%%%%%%%%%%%%%%%%%%%%%%%%%%%%%%%%%%%%%%%%%%%%%%%%%%%%%%%%%%%%%%%%%

\iffalse

%\section{Complete image-object annotation}
%\label{sec:AppChi}

With this algorithm in mind, the hierarchy of questions was constructed following the principle that false positives only add extra cost whereas false
negatives can significantly affect the quality of the labeling. 
%Figure~\ref{fig:qhierarchy} shows the constructed hierarchy. 
This was a surprisingly time-consuming process, involving multiple iterations, to ensure high accuracy of labeling and avoid false negatives.
\fi
\iffalse
\begin{figure}

\caption{The constructed hierarchy of questions for complete multi-class annotation.}
\label{fig:qhierarchy}
\end{figure}
\fi

\section{Manually curated queries for obtaining object detection scene images}
\label{sec:AppQueries}

In Section~\ref{sec:AnnotDetImages} we discussed three types of queries we used for collecting the object detection images:
(1) single object category name or a synonym; (2) a pair of object category names; (3) a manual query, typically targetting one or more object categories with insufficient data.
Here we provide a list of the 129 manually curated queries:

\vspace{0.1in}

{\tiny

\noindent 
afternoon tea, ant bridge building, armadillo race, armadillo yard, artist studio, auscultation, baby room, banjo orchestra, banjo rehersal, banjo show, califone headphones \& media player sets, camel dessert, camel tourist, carpenter drilling, carpentry, centipede wild, coffee shop, continental breakfast toaster, continental breakfast waffles, crutch walking, desert scorpion, diner, dining room, dining table, dinner, dragonfly friendly, dragonfly kid, dragonfly pond, dragonfly wild, drying hair, dumbbell curl, fan blow wind, fast food, fast food restaurant, firewood chopping, flu shot, goldfish aquarium, goldfish tank, golf cart on golf course, gym dumbbell, hamster drinking water, harmonica orchestra, harmonica rehersal, harmonica show, harp ensemble, harp orchestra, harp rehersal, harp show, hedgehog cute, hedgehog floor, hedgehog hidden, hippo bird, hippo friendly, home improvement diy drill, horseback riding, hotel coffee machine, hotel coffee maker, hotel waffle maker, jellyfish scuba, jellyfish snorkling, kitchen, kitchen counter coffee maker, kitchen counter toaster, kitchenette, koala feed, koala tree, ladybug flower, ladybug yard, laundromat, lion zebra friendly, lunch, mailman, making breakfast, making waffles, mexican food, motorcycle racing, office, office fan, opossum on tree branch, orchestra, panda play, panda tree, pizzeria, pomegranate tree, porcupine climbing trees, power drill carpenter, purse shop, red panda tree, riding competition, riding motor scooters, school supplies, scuba starfish, sea lion beach, sea otter, sea urchin habitat, shopping for school supplies, sitting in front of a fan, skunk and cat, skunk park, skunk wild, skunk yard, snail flower, snorkling starfish, snowplow cleanup, snowplow pile, snowplow winter, soccer game, south american zoo, starfish sea world, starts shopping, steamed artichoke, stethoscope doctor, strainer pasta, strainer tea, syringe doctor, table with food, tape player, tiger circus, tiger pet, using a can opener, using power drill, waffle iron breakfast, wild lion savana, wildlife preserve animals, wiping dishes, wombat petting zoo, zebra savana, zoo feeding, zoo in australia

}

\section{Hierarchy of questions for full image annotation}
\label{app:hierarchy}

The following is a hierarchy of questions manually constructed for crowdsourcing full annotation of images with the presence or absence of 200 object detection categories in ILSVRC2013 and ILSVRC2014. 
All questions are of the form ``is there a ... in the image?'' Questions marked with $\bullet$ are asked on every image.  If the answer to a question is determined to be ``no'' then the answer to all descendant questions is assumed to be ``no''. The 200 numbered leaf nodes correspond to the 200 object detection categories. 

The goal in the hierarchy construction is to save cost (by asking as few questions as possible on every image) while avoiding any ambiguity
in questions which would lead to false negatives during annotation. This hierarchy is not tree-structured; some questions have multiple parents. 

\vspace{0.1in}

{\tiny
\noindent {\bf Hierarchy of questions:} \\
\noindent \leftskip=0.0cm $\bullet$ first aid/ medical items

\noindent \leftskip=0.25cm $\circ$ (1) stethoscope

\noindent \leftskip=0.25cm $\circ$ (2) syringe

\noindent \leftskip=0.25cm $\circ$ (3) neck brace

\noindent \leftskip=0.25cm $\circ$ (4) crutch

\noindent \leftskip=0.25cm $\circ$ (5) stretcher

\noindent \leftskip=0.25cm $\circ$ (6) band aid: an adhesive bandage to cover small cuts or blisters

\noindent \leftskip=0.0cm $\bullet$ musical instruments

\noindent \leftskip=0.25cm $\circ$ (7) accordion (a portable box-shaped free-reed instrument; the reeds are made to vibrate by air from the bellows controlled by the player)

\noindent \leftskip=0.25cm $\circ$ (8) piano, pianoforte, forte-piano

\noindent \leftskip=0.25cm $\circ$ percussion instruments: chimes, maraccas, drums, etc

\noindent \leftskip=0.5cm $\circ$ (9) chime: a percussion instrument consisting of a set of tuned bells that are struck with a hammer; used as an orchestral instrument

\noindent \leftskip=0.5cm $\circ$ (10) maraca

\noindent \leftskip=0.5cm $\circ$ (11) drum

\noindent \leftskip=0.25cm $\circ$ stringed instrument

\noindent \leftskip=0.5cm $\circ$ (12) banjo, the body of a banjo is round.  please do not confuse with guitar

\noindent \leftskip=0.5cm $\circ$ (13) cello: a large stringed instrument; seated player holds it upright while playing

\noindent \leftskip=0.5cm $\circ$ (14) violin: bowed stringed instrument that has four strings, a hollow body, an unfretted fingerboard and is played with a bow. please do not confuse with cello, which is held upright while playing

\noindent \leftskip=0.5cm $\circ$ (15) harp

\noindent \leftskip=0.5cm $\circ$ (16) guitar, please do not confuse with banjo.  the body of a banjo is round

\noindent \leftskip=0.25cm $\circ$ wind instrument: a musical instrument in which the sound is produced by an enclosed column of air that is moved by the breath (such as trumpet, french horn, harmonica, flute, etc)

\noindent \leftskip=0.5cm $\circ$ (17) trumpet: a brass musical instrument with a narrow tube and a flared bell, which is played by means of valves. often has 3 keys on top

\noindent \leftskip=0.5cm $\circ$ (18) french horn: a brass musical instrument consisting of a conical tube that is coiled into a spiral, with a flared bell at the end

\noindent \leftskip=0.5cm $\circ$ (19) trombone: a brass instrument consisting of a long tube whose length can be varied by a u-shaped slide

\noindent \leftskip=0.5cm $\circ$ (20) harmonica

\noindent \leftskip=0.5cm $\circ$ (21) flute: a high-pitched musical instrument that looks like a straight tube and is usually played sideways (please do not confuse with oboes, which have a distinctive straw-like mouth piece and a slightly flared end)

\noindent \leftskip=0.5cm $\circ$ (22) oboe: a slender musical instrument roughly 65cm long with metal keys, a distinctive straw-like mouthpiece and often a slightly flared end (please do not confuse with flutes)

\noindent \leftskip=0.5cm $\circ$ (23) saxophone: a musical instrument consisting of a brass conical tube, often with a u-bend at the end

\noindent \leftskip=0.0cm $\bullet$ food: something you can eat or drink (includes growing fruit, vegetables and mushrooms, but does not include living animals)

\noindent \leftskip=0.25cm $\circ$ food with bread or crust: pretzel, bagel, pizza, hotdog, hamburgers, etc

\noindent \leftskip=0.5cm $\circ$ (24) pretzel

\noindent \leftskip=0.5cm $\circ$ (25) bagel, beigel

\noindent \leftskip=0.5cm $\circ$ (26) pizza, pizza pie

\noindent \leftskip=0.5cm $\circ$ (27) hotdog, hot dog, red hot

\noindent \leftskip=0.5cm $\circ$ (28) hamburger, beefburger, burger

\noindent \leftskip=0.25cm $\circ$ (29) guacamole

\noindent \leftskip=0.25cm $\circ$ (30) burrito

\noindent \leftskip=0.25cm $\circ$ (31) popsicle (ice cream or water ice on a small wooden stick)

\noindent \leftskip=0.25cm $\circ$ fruit

\noindent \leftskip=0.5cm $\circ$ (32) fig

\noindent \leftskip=0.5cm $\circ$ (33) pineapple, ananas

\noindent \leftskip=0.5cm $\circ$ (34) banana

\noindent \leftskip=0.5cm $\circ$ (35) pomegranate

\noindent \leftskip=0.5cm $\circ$ (36) apple

\noindent \leftskip=0.5cm $\circ$ (37) strawberry

\noindent \leftskip=0.5cm $\circ$ (38) orange

\noindent \leftskip=0.5cm $\circ$ (39) lemon

\noindent \leftskip=0.25cm $\circ$ vegetables

\noindent \leftskip=0.5cm $\circ$ (40) cucumber, cuke

\noindent \leftskip=0.5cm $\circ$ (41) artichoke, globe artichoke

\noindent \leftskip=0.5cm $\circ$ (42) bell pepper

\noindent \leftskip=0.5cm $\circ$ (43) head cabbage

\noindent \leftskip=0.25cm $\circ$ (44) mushroom

\noindent \leftskip=0.0cm $\bullet$ items that run on electricity (plugged in or using batteries); including clocks, microphones, traffic lights, computers, etc

\noindent \leftskip=0.25cm $\circ$ (45) remote control, remote

\noindent \leftskip=0.25cm $\circ$ electronics that blow air

\noindent \leftskip=0.5cm $\circ$ (46) hair dryer, blow dryer

\noindent \leftskip=0.5cm $\circ$ (47) electric fan: a device for creating a current of air by movement of a surface or surfaces (please do not consider hair dryers)

\noindent \leftskip=0.25cm $\circ$ electronics that can play music or amplify sound

\noindent \leftskip=0.5cm $\circ$ (48) tape player

\noindent \leftskip=0.5cm $\circ$ (49) iPod

\noindent \leftskip=0.25cm $\circ$ (50) microphone, mike

\noindent \leftskip=0.25cm $\circ$ computer and computer peripherals: mouse, laptop, printer, keyboard, etc

\noindent \leftskip=0.5cm $\circ$ (51) computer mouse

\noindent \leftskip=0.5cm $\circ$ (52) laptop, laptop computer

\noindent \leftskip=0.5cm $\circ$ (53) printer (please do not consider typewriters to be printers)

\noindent \leftskip=0.5cm $\circ$ (54) computer keyboard

\noindent \leftskip=0.25cm $\circ$ (55) lamp

\noindent \leftskip=0.25cm $\circ$ electric cooking appliance (an appliance which generates heat to cook food or boil water)

\noindent \leftskip=0.5cm $\circ$ (56) microwave, microwave oven

\noindent \leftskip=0.5cm $\circ$ (57) toaster

\noindent \leftskip=0.5cm $\circ$ (58) waffle iron

\noindent \leftskip=0.5cm $\circ$ (59) coffee maker: a kitchen appliance used for brewing coffee automatically

\noindent \leftskip=0.25cm $\circ$ (60) vacuum, vacuum cleaner

\noindent \leftskip=0.25cm $\circ$ (61) dishwasher, dish washer, dishwashing machine

\noindent \leftskip=0.25cm $\circ$ (62) washer, washing machine: an electric appliance for washing clothes

\noindent \leftskip=0.25cm $\circ$ (63) traffic light, traffic signal, stoplight

\noindent \leftskip=0.25cm $\circ$ (64) tv or monitor: an electronic device that represents information in visual form

\noindent \leftskip=0.25cm $\circ$ (65) digital clock: a clock that displays the time of day digitally

\noindent \leftskip=0.0cm $\bullet$ kitchen items: tools,utensils and appliances usually found in the kitchen

\noindent \leftskip=0.25cm $\circ$ electric cooking appliance (an appliance which generates heat to cook food or boil water)

\noindent \leftskip=0.5cm $\circ$ (56) microwave, microwave oven

\noindent \leftskip=0.5cm $\circ$ (57) toaster

\noindent \leftskip=0.5cm $\circ$ (58) waffle iron

\noindent \leftskip=0.5cm $\circ$ (59) coffee maker: a kitchen appliance used for brewing coffee automatically

\noindent \leftskip=0.25cm $\circ$ (61) dishwasher, dish washer, dishwashing machine

\noindent \leftskip=0.25cm $\circ$ (66) stove

\noindent \leftskip=0.25cm $\circ$ things used to open cans/bottles: can opener or corkscrew

\noindent \leftskip=0.5cm $\circ$ (67) can opener (tin opener)

\noindent \leftskip=0.5cm $\circ$ (68) corkscrew

\noindent \leftskip=0.25cm $\circ$ (69) cocktail shaker

\noindent \leftskip=0.25cm $\circ$ non-electric item commonly found in the kitchen: pot, pan, utensil, bowl, etc

\noindent \leftskip=0.5cm $\circ$ (70) strainer

\noindent \leftskip=0.5cm $\circ$ (71) frying pan (skillet)

\noindent \leftskip=0.5cm $\circ$ (72) bowl: a dish for serving food that is round, open at the top, and has no handles (please do not confuse with a cup, which usually has a handle and is used for serving drinks)

\noindent \leftskip=0.5cm $\circ$ (73) salt or pepper shaker: a shaker with a perforated top for sprinkling salt or pepper

\noindent \leftskip=0.5cm $\circ$ (74) plate rack

\noindent \leftskip=0.5cm $\circ$ (75) spatula: a turner with a narrow flexible blade

\noindent \leftskip=0.5cm $\circ$ (76) ladle: a spoon-shaped vessel with a long handle; frequently used to transfer liquids from one container to another

\noindent \leftskip=0.25cm $\circ$ (77) refrigerator, icebox

\noindent \leftskip=0.0cm $\bullet$ furniture (including benches)

\noindent \leftskip=0.25cm $\circ$ (78) bookshelf:  a shelf on which to keep books

\noindent \leftskip=0.25cm $\circ$ (79) baby bed: small bed for babies, enclosed by sides to prevent baby from falling

\noindent \leftskip=0.25cm $\circ$ (80) filing cabinet: office furniture consisting of a container for keeping papers in order

\noindent \leftskip=0.25cm $\circ$ (81) bench (a long seat for several people, typically made of wood or stone)

\noindent \leftskip=0.25cm $\circ$ (82) chair: a raised piece of furniture for one person to sit on; please do not confuse with benches or sofas, which are made for more people

\noindent \leftskip=0.25cm $\circ$ (83) sofa, couch: upholstered seat for more than one person; please do not confuse with benches (which are made of wood or stone) or with chairs (which are for just one person)

\noindent \leftskip=0.25cm $\circ$ (84) table

\noindent \leftskip=0.0cm $\bullet$ clothing, article of clothing: a covering designed to be worn on a person's body

\noindent \leftskip=0.25cm $\circ$ (85) diaper: Garment consisting of a folded cloth drawn up between the legs and fastened at the waist; worn by infants to catch excrement

\noindent \leftskip=0.25cm $\circ$ swimming attire: clothes used for swimming or bathing (swim suits, swim trunks, bathing caps)

\noindent \leftskip=0.5cm $\circ$ (86) swimming trunks: swimsuit worn by men while swimming

\noindent \leftskip=0.5cm $\circ$ (87) bathing cap, swimming cap: a cap worn to keep hair dry while swimming or showering

\noindent \leftskip=0.5cm $\circ$ (88) maillot: a woman's one-piece bathing suit

\noindent \leftskip=0.25cm $\circ$ necktie: a man's formal article of clothing worn around the neck (including bow ties)

\noindent \leftskip=0.5cm $\circ$ (89) bow tie: a man's tie that ties in a bow

\noindent \leftskip=0.5cm $\circ$ (90) tie: a long piece of cloth worn for decorative purposes around the neck or shoulders, resting under the shirt collar and knotted at the throat (NOT a bow tie)

\noindent \leftskip=0.25cm $\circ$ headdress, headgear: clothing for the head (hats, helmets, bathing caps, etc)

\noindent \leftskip=0.5cm $\circ$ (87) bathing cap, swimming cap: a cap worn to keep hair dry while swimming or showering

\noindent \leftskip=0.5cm $\circ$ (91) hat with a wide brim

\noindent \leftskip=0.5cm $\circ$ (92) helmet: protective headgear made of hard material to resist blows

\noindent \leftskip=0.25cm $\circ$ (93) miniskirt, mini: a very short skirt

\noindent \leftskip=0.25cm $\circ$ (94) brassiere, bra: an undergarment worn by women to support their breasts

\noindent \leftskip=0.25cm $\circ$ (95) sunglasses

\noindent \leftskip=0.0cm $\bullet$ living organism (other than people): dogs, snakes, fish, insects, sea urchins, starfish, etc.

\noindent \leftskip=0.25cm $\circ$ living organism which can fly

\noindent \leftskip=0.5cm $\circ$ (96) bee

\noindent \leftskip=0.5cm $\circ$ (97) dragonfly

\noindent \leftskip=0.5cm $\circ$ (98) ladybug

\noindent \leftskip=0.5cm $\circ$ (99) butterfly

\noindent \leftskip=0.5cm $\circ$ (100) bird

\noindent \leftskip=0.25cm $\circ$ living organism which cannot fly (please don't include humans)

\noindent \leftskip=0.5cm $\circ$ living organism with 2 or 4 legs (please don't include humans):

\noindent \leftskip=0.75cm $\circ$ mammals (but please do not include humans)

\noindent \leftskip=1.0cm $\circ$ feline (cat-like) animal: cat, tiger or lion

\noindent \leftskip=1.25cm $\circ$ (101) domestic cat

\noindent \leftskip=1.25cm $\circ$ (102) tiger

\noindent \leftskip=1.25cm $\circ$ (103) lion

\noindent \leftskip=1.0cm $\circ$ canine (dog-like animal): dog, hyena, fox or wolf

\noindent \leftskip=1.25cm $\circ$ (104) dog, domestic dog, canis familiaris

\noindent \leftskip=1.25cm $\circ$ (105) fox: wild carnivorous mammal with pointed muzzle and ears and a bushy tail (please do not confuse with dogs)

\noindent \leftskip=1.0cm $\circ$ animals with hooves: camels, elephants, hippos, pigs, sheep, etc

\noindent \leftskip=1.25cm $\circ$ (106) elephant

\noindent \leftskip=1.25cm $\circ$ (107) hippopotamus, hippo

\noindent \leftskip=1.25cm $\circ$ (108) camel

\noindent \leftskip=1.25cm $\circ$ (109) swine: pig or boar

\noindent \leftskip=1.25cm $\circ$ (110) sheep: woolly animal, males have large spiraling horns (please do not confuse with antelope which have long legs)

\noindent \leftskip=1.25cm $\circ$ (111) cattle: cows or oxen (domestic bovine animals)

\noindent \leftskip=1.25cm $\circ$ (112) zebra

\noindent \leftskip=1.25cm $\circ$ (113) horse

\noindent \leftskip=1.25cm $\circ$ (114) antelope: a graceful animal with long legs and horns directed upward and backward

\noindent \leftskip=1.0cm $\circ$ (115) squirrel

\noindent \leftskip=1.0cm $\circ$ (116) hamster: short-tailed burrowing rodent with large cheek pouches

\noindent \leftskip=1.0cm $\circ$ (117) otter

\noindent \leftskip=1.0cm $\circ$ (118) monkey

\noindent \leftskip=1.0cm $\circ$ (119) koala bear

\noindent \leftskip=1.0cm $\circ$ (120) bear (other than pandas)

\noindent \leftskip=1.0cm $\circ$ (121) skunk (mammal known for its ability fo spray a liquid with a strong odor; they may have a single thick stripe across back and tail, two thinner stripes, or a series of white spots and broken stripes

\noindent \leftskip=1.0cm $\circ$ (122) rabbit

\noindent \leftskip=1.0cm $\circ$ (123) giant panda: an animal characterized by its distinct black and white markings

\noindent \leftskip=1.0cm $\circ$ (124) red panda: Reddish-brown Old World raccoon-like carnivore

\noindent \leftskip=0.75cm $\circ$ (125) frog, toad

\noindent \leftskip=0.75cm $\circ$ (126) lizard: please do not confuse with snake (lizards have legs)

\noindent \leftskip=0.75cm $\circ$ (127) turtle

\noindent \leftskip=0.75cm $\circ$ (128) armadillo

\noindent \leftskip=0.75cm $\circ$ (129) porcupine, hedgehog

\noindent \leftskip=0.5cm $\circ$ living organism with 6 or more legs: lobster, scorpion, insects, etc.

\noindent \leftskip=0.75cm $\circ$ (130) lobster: large marine crustaceans with long bodies and muscular tails; three of their five pairs of legs have claws

\noindent \leftskip=0.75cm $\circ$ (131) scorpion

\noindent \leftskip=0.75cm $\circ$ (132) centipede: an arthropod having a flattened body of 15 to 173 segments each with a pair of legs, the foremost pair being modified as prehensors

\noindent \leftskip=0.75cm $\circ$ (133) tick (a small creature with 4 pairs of legs which lives on the blood of mammals and birds)

\noindent \leftskip=0.75cm $\circ$ (134) isopod: a small crustacean with seven pairs of legs adapted for crawling

\noindent \leftskip=0.75cm $\circ$ (135) ant

\noindent \leftskip=0.5cm $\circ$ living organism without legs: fish, snake, seal, etc. (please don't include plants)

\noindent \leftskip=0.75cm $\circ$ living organism that lives in water: seal, whale, fish, sea cucumber, etc.

\noindent \leftskip=1.0cm $\circ$ (136) jellyfish

\noindent \leftskip=1.0cm $\circ$ (137) starfish, sea star

\noindent \leftskip=1.0cm $\circ$ (138) seal

\noindent \leftskip=1.0cm $\circ$ (139) whale

\noindent \leftskip=1.0cm $\circ$ (140) ray: a marine animal with a horizontally flattened body and enlarged winglike pectoral fins with gills on the underside

\noindent \leftskip=1.0cm $\circ$ (141) goldfish: small golden or orange-red fishes

\noindent \leftskip=0.75cm $\circ$ living organism that slides on land: worm, snail, snake

\noindent \leftskip=1.0cm $\circ$ (142) snail

\noindent \leftskip=1.0cm $\circ$ (143) snake: please do not confuse with lizard (snakes do not have legs)

\noindent \leftskip=0.0cm $\bullet$ vehicle: any object used to move people or objects from place to place

\noindent \leftskip=0.25cm $\circ$ a vehicle with wheels

\noindent \leftskip=0.5cm $\circ$ (144) golfcart, golf cart

\noindent \leftskip=0.5cm $\circ$ (145) snowplow: a vehicle used to push snow from roads

\noindent \leftskip=0.5cm $\circ$ (146) motorcycle (or moped)

\noindent \leftskip=0.5cm $\circ$ (147) car, automobile (not a golf cart or a bus)

\noindent \leftskip=0.5cm $\circ$ (148) bus: a vehicle carrying many passengers; used for public transport

\noindent \leftskip=0.5cm $\circ$ (149) train

\noindent \leftskip=0.5cm $\circ$ (150) cart: a heavy open wagon usually having two wheels and drawn by an animal

\noindent \leftskip=0.5cm $\circ$ (151) bicycle, bike: a two wheeled vehicle moved by foot pedals

\noindent \leftskip=0.5cm $\circ$ (152) unicycle, monocycle

\noindent \leftskip=0.25cm $\circ$ a vehicle without wheels (snowmobile, sleighs)

\noindent \leftskip=0.5cm $\circ$ (153) snowmobile: tracked vehicle for travel on snow

\noindent \leftskip=0.5cm $\circ$ (154) watercraft (such as ship or boat): a craft designed for water transportation

\noindent \leftskip=0.25cm $\circ$ (155) airplane: an aircraft powered by propellers or jets

\noindent \leftskip=0.0cm $\bullet$ cosmetics: toiletry designed to beautify the body

\noindent \leftskip=0.25cm $\circ$ (156) face powder

\noindent \leftskip=0.25cm $\circ$ (157) perfume, essence (usually comes in a smaller bottle than hair spray

\noindent \leftskip=0.25cm $\circ$ (158) hair spray

\noindent \leftskip=0.25cm $\circ$ (159) cream, ointment, lotion

\noindent \leftskip=0.25cm $\circ$ (160) lipstick, lip rouge

\noindent \leftskip=0.0cm $\bullet$ carpentry items: items used in carpentry, including nails, hammers, axes, screwdrivers, drills, chain saws, etc

\noindent \leftskip=0.25cm $\circ$ (161) chain saw, chainsaw

\noindent \leftskip=0.25cm $\circ$ (162) nail: pin-shaped with a head on one end and a point on the other

\noindent \leftskip=0.25cm $\circ$ (163) axe: a sharp tool often used to cut trees/ logs

\noindent \leftskip=0.25cm $\circ$ (164) hammer: a blunt hand tool used to drive nails in or break things apart (please do not confuse with axe, which is sharp)

\noindent \leftskip=0.25cm $\circ$ (165) screwdriver

\noindent \leftskip=0.25cm $\circ$ (166) power drill: a  power tool for drilling holes into hard materials

\noindent \leftskip=0.0cm $\bullet$ school supplies: rulers, erasers, pencil sharpeners, pencil boxes, binders

\noindent \leftskip=0.25cm $\circ$ (167) ruler,rule: measuring stick consisting of a strip of wood or metal or plastic with a straight edge that is used for drawing straight lines and measuring lengths

\noindent \leftskip=0.25cm $\circ$ (168) rubber eraser, rubber, pencil eraser

\noindent \leftskip=0.25cm $\circ$ (169) pencil sharpener

\noindent \leftskip=0.25cm $\circ$ (170) pencil box, pencil case

\noindent \leftskip=0.25cm $\circ$ (171) binder, ring-binder

\noindent \leftskip=0.0cm $\bullet$ sports items: items used to play sports or in the gym (such as skis, raquets, gymnastics bars, bows, punching bags, balls)

\noindent \leftskip=0.25cm $\circ$ (172) bow: weapon for shooting arrows, composed of a curved piece of resilient wood with a taut cord to propel the arrow

\noindent \leftskip=0.25cm $\circ$ (173) puck, hockey puck: vulcanized rubber disk 3 inches in diameter that is used instead of a ball in ice hockey

\noindent \leftskip=0.25cm $\circ$ (174) ski

\noindent \leftskip=0.25cm $\circ$ (175) racket, racquet

\noindent \leftskip=0.25cm $\circ$ gymnastic equipment: parallel bars, high beam, etc

\noindent \leftskip=0.5cm $\circ$ (176) balance beam: a horizontal bar used for gymnastics which is raised from the floor and wide enough to walk on

\noindent \leftskip=0.5cm $\circ$ (177) horizontal bar, high bar: used for gymnastics; gymnasts grip it with their hands (please do not confuse with balance beam, which is wide enough to walk on)

\noindent \leftskip=0.25cm $\circ$ ball

\noindent \leftskip=0.5cm $\circ$ (178) golf ball

\noindent \leftskip=0.5cm $\circ$ (179) baseball

\noindent \leftskip=0.5cm $\circ$ (180) basketball

\noindent \leftskip=0.5cm $\circ$ (181) croquet ball

\noindent \leftskip=0.5cm $\circ$ (182) soccer ball

\noindent \leftskip=0.5cm $\circ$ (183) ping-pong ball

\noindent \leftskip=0.5cm $\circ$ (184) rugby ball

\noindent \leftskip=0.5cm $\circ$ (185) volleyball

\noindent \leftskip=0.5cm $\circ$ (186) tennis ball

\noindent \leftskip=0.25cm $\circ$ (187) punching bag, punch bag, punching ball, punchball

\noindent \leftskip=0.25cm $\circ$ (188) dumbbell: An exercising weight; two spheres connected by a short bar that serves as a handle

\noindent \leftskip=0.0cm $\bullet$ liquid container: vessels which commonly contain liquids such as bottles, cans, etc.

\noindent \leftskip=0.25cm $\circ$ (189) pitcher: a vessel with a handle and a spout for pouring

\noindent \leftskip=0.25cm $\circ$ (190) beaker: a flatbottomed jar made of glass or plastic; used for chemistry

\noindent \leftskip=0.25cm $\circ$ (191) milk can

\noindent \leftskip=0.25cm $\circ$ (192) soap dispenser

\noindent \leftskip=0.25cm $\circ$ (193) wine bottle

\noindent \leftskip=0.25cm $\circ$ (194) water bottle

\noindent \leftskip=0.25cm $\circ$ (195) cup or mug (usually with a handle and usually cylindrical)

\noindent \leftskip=0.0cm $\bullet$ bag

\noindent \leftskip=0.25cm $\circ$ (196) backpack: a bag carried by a strap on your back or shoulder

\noindent \leftskip=0.25cm $\circ$ (197) purse: a small bag for carrying money

\noindent \leftskip=0.25cm $\circ$ (198) plastic bag

\noindent \leftskip=0.0cm $\bullet$ (199) person

\noindent \leftskip=0.0cm $\bullet$ (200) flower pot: a container in which plants are cultivated

}

%%%%%%%%%%%%%%%%%%%%%%%%%%%%%%%%%%%%%%%%%%%%%%%%%%%%%%%%%%%%%%%%%%%%%%%%%%%%%%%%%%

\section{Modification to bounding box system for object detection}
\label{sec:AppDetBbox}

The bounding box annotation system described in Section~\ref{sec:AnnotLocBbox} is used for
annotating images for both the single-object localization dataset and the object detection dataset.
However, two additional manual  post-processing are needed to ensure accuracy in the object
detection scenario:

\paragraph{Ambiguous objects.} The first common source of error was that
workers were not able to accurately differentiate some object classes during annotation. 
Some commonly confused
labels were seal and sea otter, backpack and purse, banjo and guitar, violin and cello,
brass instruments (trumpet, trombone, french horn and brass), flute and oboe, ladle and spatula. 
Despite our best efforts (providing positive and negative example images in the annotation task,
adding text explanations to alert the user to the distinction between these categories) these errors persisted.

In the single-object localization
setting, this problem was not as prominent for two reasons. First, the way the data was collected imposed a strong prior on the object class which was present.
Second, since only one object category needed to be annotated per image, ambiguous images could be discarded: for example, if workers couldn't agree on whether or not a trumpet was in fact present, this image could simply
be removed. In contrast, for the object detection setting
consensus had to be reached for all target categories on all images. 

To fix this problem, once bounding box annotations were collected we manually looked through all cases where the bounding
boxes for two different object classes had significant overlap with each other (about $3\%$ of the collected boxes). 
About a quarter of these boxes were found to correspond to incorrect objects and were removed.
Crowdsourcing this post-processing step (with very stringent accuracy constraints) would
be possible but it occurred in few enough cases that it was faster (and more accurate) to do this in-house.

\paragraph{Duplicate annotations.} The second common source of error were duplicate bounding boxes drawn
on the same object instance. Despite instructions not to draw more than one bounding box around the same object instance
and constraints in the annotation UI enforcing at least a 5 pixel difference between different bounding boxes,
these errors persisted. One reason was that sometimes  the initial bounding box was not
perfect and subsequent labelers drew a slightly improved alternative.

This type of error was also present in the single-object localization scenario but was not a major cause for concern. 
A duplicate bounding box is a slightly perturbed but still correct positive example, and single-object localization is only
concerned with correctly localizing one object instance. For the detection task algorithms are evaluated on the ability to localize
\emph{every} object instance, and penalized for duplicate detections, so it is imperative that these labeling errors
are corrected (even if they only appear in about $0.6\%$ of cases).

Approximately $1\%$ of bounding boxes were found to have significant overlap of more than $50\%$ with another
bounding box of the same object class.We again manually verified all of these cases in-house.  In approximately $40\%$ of the cases the two bounding boxes correctly
corresponded to different people in a crowd, to stacked plates, or to musical instruments nearby in an orchestra.
In the other $60\%$ of cases one of the boxes was randomly removed. 

These verification steps complete the annotation procedure of bounding boxes around every instance of every object class in 
validation, test and a subset of training images for the detection task.

\paragraph{Training set annotation.}
With the optimized algorithm of Section~\ref{sec:AnnotDetList} we fully annotated the validation and test sets. However, annotating \emph{all} training images 
with all target object classes was still a budget challenge.
Positive training images taken from the single-object localization dataset already had bounding box annotations of all instances
of one object class on each image. We extended the existing annotations to the detection dataset by making two modification. First, we corrected any bounding box omissions resulting from merging fine-grained categories: 
i.e., if an image belonged to the "dalmatian" category and all instances of "dalmatian" were annotated with bounding boxes for single-object localization, we ensured that
all remaining "dog" instances are also annotated for the object detection task. 
Second, we collected significantly more training data for the person class because the existing annotation set was not diverse enough to be
representative (the only people categories in the single-object localization task are scuba diver, groom, and ballplayer). To compensate,
we additionally annotated people in a large fraction of the existing training set images. 

%%%%%%%%%%%%%%%%%%%%%%%%%%%%%%%%%%%%%%%%%%%%%%%%%%%%%%%%%%%%%%%%%%%%%%%%%%%%%%%%%%

\iffalse
\section{Training set annotation}
\label{sec:AppDetLocExt}

First, we addressed the cases where a subcategory
of the target class was annotated (such as a specific dog breed for single-object localization instead of the target dog class for detection task).
We continued running the bounding box annotation system to ensure that all instances of the target class are annotated.

Second, we collected significantly more training data for the person class because the existing annotation set was not nearly diverse enough to be
representative (the only people categories in the single-object localization task were scuba diver, groom, and ballplayer). 
Based on the statistics on the fully annotated validation images we selected
 \todo{100?} of the 200 object categories most likely to contain a person and annotated all instances of people on a subset of the images \todo{how many}.
This provided \todo{XXX} additional person annotations in a wide variety of settings.
\fi 

%%%%%%%%%%%%%%%%%%%%%%%%%%%%%%%%%%%%%%%%%%%%%%%%%%%%%%%%%%%%%%%%%%%%%%%%%%%%%%%%%%

\section{Competition protocol}
\label{sec:AppCompetition}

%\todo{discuss external data and other competition rules here if needed; discuss requiring localization submission}

\paragraph{Competition format.} 
At the beginning of the competition period each year we release the new training/validation/test images, training/validation annotations, and competition
specification for the year. We then specify a deadline for submission, usually approximately 4 months after the release of data. Teams are asked to upload
a text file of their predicted annotations on test images by this deadline to a provided server. We then evaluate all submissions and release the results.

For every task we released code that takes a text file of automatically generated image annotations and compares it with
the ground truth annotations to return a quantitative measure of algorithm accuracy. Teams can use this code to evaluate
their performance on the validation data.

As described in~\citep{PASCALIJCV2}, 
there are three options for measuring performance on test data: (i) Release test images and annotations, and allow participants to assess
performance themselves; (ii) Release test images but not test annotations -- participants submit results and organizers assess performance; (iii) Neither
test images nor annotations are released -- participants submit software and organizers run it on new data and assess performance. In line with the 
PASCAL VOC choice, we opted for option (ii). Option (i) allows too much leeway in overfitting to the test data; option (iii) is infeasible, especially given
the scale of our test set (40K-100K images).

We released ILSVRC2010 test annotations for the image classification task, but all other test annotations have remained hidden to discourage
fine-tuning results on the test data.

\paragraph{Evaluation protocol after the challenge.} 

After the challenge period we set up an automatic evaluation server that researchers can use throughout the year to continue evaluating their algorithms
against the ground truth test annotations. We limit teams to 2 submissions per week to discourage parameter tuning on the test data, and in practice
we have never had a problem with researchers abusing the system.

\end{appendices}

\bibliographystyle{apalike}
\bibliography{template}   % name your BibTeX data base

%\begin{thebibliography}{}
% etc
%\end{thebibliography}

\end{document}